\documentclass[sigconf,10pt]{acmart}
\renewcommand\footnotetextcopyrightpermission[1]{} 
\usepackage{booktabs} 
\usepackage{balance}
\usepackage{textcomp}

\usepackage{graphics, epstopdf, graphicx, epsfig, psfrag, epsf, subfigure, wrapfig}
\usepackage{amssymb}
\usepackage{amsfonts, amsmath}
\usepackage{mathtools}
\usepackage{multicol, multirow, verbatim, caption,float}
\usepackage{enumerate}

\usepackage[listings,skins]{tcolorbox}
\usepackage{algorithm, algcompatible}

\usepackage{xspace}
\usepackage{xcolor, color}
\usepackage{hyphenat}
\usepackage{array, calc}

\newcommand{\ie}{{\em i.e., }}
\newcommand{\eg}{{\em e.g., }}
\newcommand{\etc}{{\em etc. }}
\newcommand{\etcfree}{{\em etc.}}
\newcommand{\etal}{{\em et al. }}
\newcommand{\aka}{{\em a.k.a. }}
\newcommand{\vs}{{\it vs.}\xspace}

\definecolor{heraldRed}{rgb}{1,0,0}
\definecolor{heraldBlue}{rgb}{0.0,0.0,0.8}
\definecolor{heraldGreen}{rgb}{0.0,0.8,0.8}
\definecolor{heraldPurple}{rgb}{0.9,0.1,0.9}

\newcommand{\manote}[1]{\textcolor{heraldBlue}{\small \bf [Manos: #1]}}

\newcommand{\redX}{{{\color{red}{\text{\sffamily X}}}}}

\newcommand{\RFs}{\textvtt{RFs}}
\newcommand{\RFsWeights}{\textvtt{RFs}$_w$}
\newcommand{\RFsSpatial}{\textvtt{RFs}$_{x,y}$}
\newcommand{\RFsSpatiotemporal}{\textvtt{RFs}$_{x,y,t}$}
\newcommand{\RFsAll}{\textvtt{RFs}$_{all}$}
\newcommand{\LDPL}{\textvtt{LDPL}}
\newcommand{\LDPLknn}{\textvtt{LDPL-knn}}
\newcommand{\LDPLhom}{\textvtt{LDPL-hom}}

\newcommand{\rsrp}{$y^P$}
\newcommand{\rsrq}{$y^I$}
\newcommand{\cqi}{$y^C$}

\newcommand{\Qobj}[1]{${Q}($#1$)$}
\newcommand{\Qhat}[1]{$\widehat{Q}($#1$)$}

\newcommand{\cdp}[1]{$Q_{cdp}($#1$)$}
\newcommand{\cdpPred}[1]{$\widehat{Q}_{cdp}($#1$)$}
\newcommand{\coverage}{$Q_{c}(y)$}
\newcommand{\signalbars}{$Q_{B}(y)$}

\newcommand{\targetdistr}{$p(\mathbf{x})$}
\newcommand{\population}{$d(\mathbf{x})$}
\newcommand{\pop}[1]{${d}($#1$)$}

\newcommand{\wu}{$w_u$}
\newcommand{\wP}{$w_d$}

\newcommand{\erroruniform}{$\varepsilon_{u}$}
\newcommand{\errorpopulation}{$\varepsilon_{d}$}

\newcommand{\problemspace}{$P=(Q,W)$}
\newcommand{\baseproblem}{$P_B=(I,k)$}
\newcommand{\problem}[1]{$P=($#1$)$}

\newcommand{\features}{$\mathbf{x}$}
\newcommand{\cid}{$cID$}
\newcommand{\lteta}{LTE TA}
\newcommand{\dev}{$dev$}

\newcommand{\freq}{$freq_{dl}$}

\newcommand{\mno}{operators}

\newcommand{\maps}{coverage maps}

\newcommand\textvtt[1]{{\normalfont\fontfamily{cmvtt}\selectfont #1}}

\newcommand{\inhouse}{\textvtt{Campus}\xspace}
\newcommand{\inhousedataset}{\textvtt{Campus data\-set}\xspace}

\newcommand{\externaldataset}{\textvtt{NYC dataset}\xspace}
\newcommand{\external}{\textvtt{NYC}\xspace}

\newcommand{\externalsuburban}{\textvtt{LA}\xspace}

\newcommand{\externalcities}{\textvtt{NYC and LA}\xspace}
\newcommand{\externalcitiesdataset}{\textvtt{NYC and LA datasets}\xspace}

\DeclareMathOperator{\E}{\mathbb{E}}

 \newcommand{\gCheck}{{{$\color{green}{\Huge \checkmark}$}}}

\usepackage[skipbelow=\topskip,skipabove=\topskip]{mdframed}

\usepackage{url}
\usepackage{breakurl}

\makeatletter
\def\url@leostyle{%
	\@ifundefined{selectfont}{\def\UrlFont{\sf}}{\def\UrlFont{\small\ttfamily}}}
\makeatother
\AtBeginDocument{%
  \providecommand\BibTeX{{%
    \normalfont B\kern-0.5em{\scshape i\kern-0.25em b}\kern-0.8em\TeX}}}

\acmConference[]{}{}{}

\begin{document}

\title{A Unified Prediction Framework for Signal  Maps}
\subtitle{Not all Measurements are Created Equal}
\settopmatter{authorsperrow=3}
\author{Emmanouil Alimpertis}
\affiliation{%
	\institution{University of California Irvine}
	\streetaddress{University of California Irvine}
	\country{USA}
	\postcode{92697}
}
\email{ealimper@uci.edu}
\author{Athina Markopoulou}
\affiliation{%
	\institution{University of California Irvine}
	\streetaddress{University of California Irvine}
	\country{USA}
	\postcode{92697}
}
\email{athina@uci.edu}
\author{Carter T. Butts}
\affiliation{%
	\institution{University of California Irvine}
	\streetaddress{University of California Irvine}
	\postcode{92697}
	\country{USA}
}
\email{buttsc@uci.edu}
\author{Evita Bakopoulou}
\affiliation{%
	\institution{University of California Irvine}
	\streetaddress{University of California Irvine}
	\postcode{92697}
	\country{USA}
}
\email{ebakopou@uci.edu}
\author{Konstantinos Psounis}
\affiliation{%
	\institution{University of Southern California}
	\streetaddress{University of Southern California}
	\postcode{90089}
	\country{USA}
}
\email{kpsounis@usc.edu}

\renewcommand{\shortauthors}{E. Alimpertis, et al.}

\newcommand\blfootnote[1]{%
  \begingroup
  \renewcommand\thefootnote{}\footnote{#1}%
  \addtocounter{footnote}{-1}%
  \endgroup
}

\begin{abstract}
Signal maps are essential for the planning and operation of cellular networks. However, the measurements needed to create such maps are expensive, often biased, not always reflecting the performance metrics of interest, and posing privacy risks. In this paper, we develop a unified framework for predicting cellular performance maps from limited available  measurements.
Our framework builds on a state-of-the-art random-forest predictor, or any other base predictor. We propose and combine three mechanisms that deal with the fact that not all measurements are equally important for a particular prediction task.
First, we  design \emph{quality-of-service functions ($Q$)},  including signal strength (RSRP) but also other metrics of interest to operators, such as number of bars, coverage (improving recall by 76\%-92\%) and call drop probability (reducing error by as much as 32\%). By implicitly altering the loss function employed in learning, quality functions can also improve prediction for RSRP itself where it matters (\eg MSE reduction up to 27\% 
 in the low signal strength regime, where high accuracy is critical).  
Second, we introduce \emph{weight functions} ($W$) to specify the relative importance of prediction at different locations and other parts of the feature space. We propose  re-weighting based on importance sampling to obtain unbiased estimators when the sampling and target distributions are different. This yields improvements up to 20\% for targets based on spatially uniform loss or losses based on user population density.  
Third, we apply the {\em Data Shapley} framework for the first time in this context: to assign values ($\phi$) to individual measurement points, which capture the importance of their contribution to the prediction task. This can improve prediction (\eg from 64\% to 94\% in recall for coverage loss) by removing points with negative values, and can also enable data minimization (\ie we show that we can remove 70\% of data w/o loss in performance).
We evaluate our methods and demonstrate significant improvement in prediction performance, using several real-world datasets.

\end{abstract}

\maketitle

\blfootnote{Preprint. Under review.}%

\section{Introduction}\label{sec:introduction}
\begin{figure}[t!]
  \begin{center}
    \includegraphics[scale = 0.15]{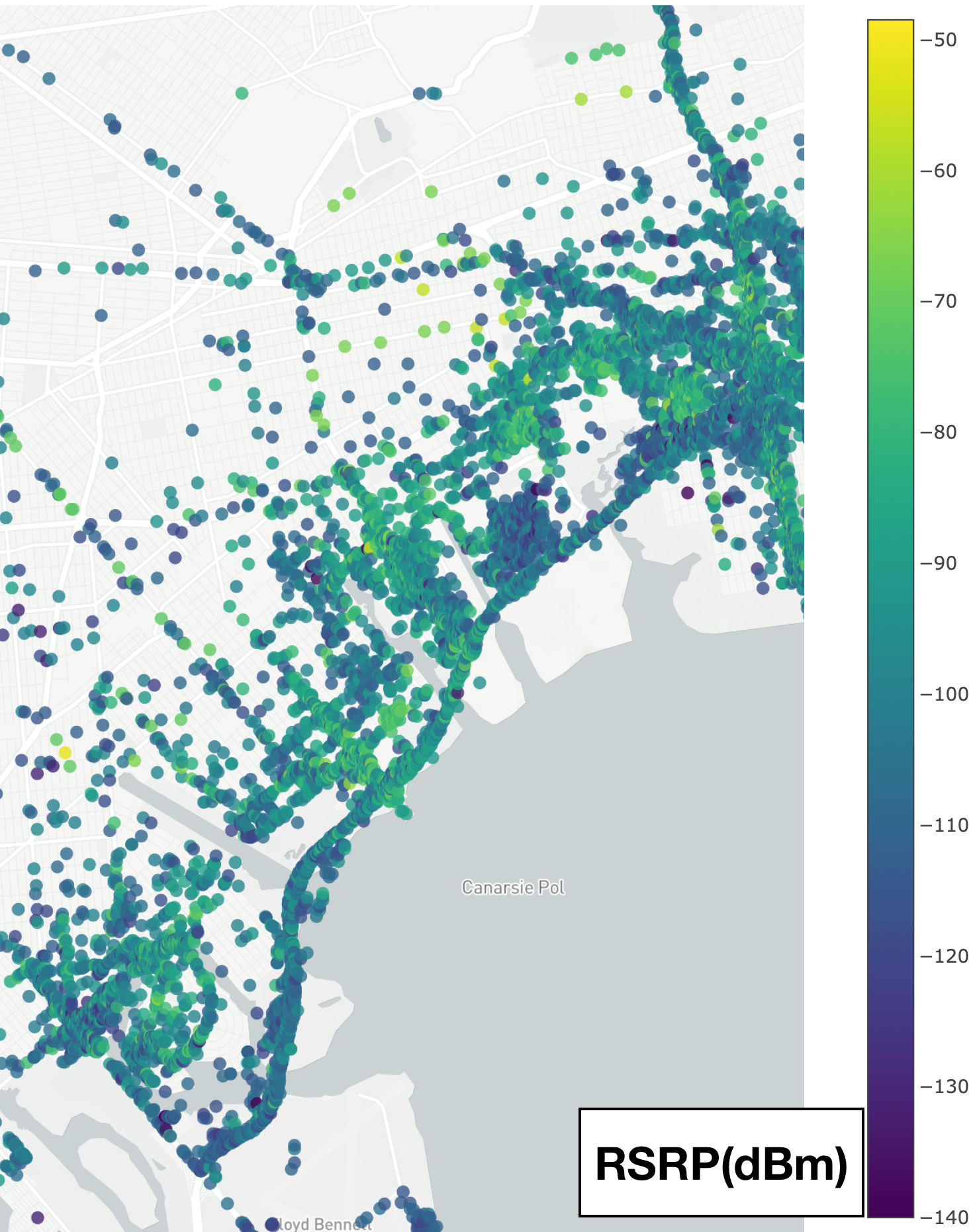}
  \end{center}
  \caption{Example of a typical Signal Strength (RSRP)  Map from the East \external dataset, near the JFK airport.}
  \label{fig:rsrpmap_eastnyc_x540}
   \vspace{-10pt}
\end{figure}

Cellular operators rely on key performance indicators (\aka KPIs) to understand the performance and coverage of their network, as well as that of their competitors, in their effort to provide the best user experience. KPIs usually include wireless signal strength measurements (\eg LTE reference signal received power, \aka RSRP), other performance metrics (\eg throughput, delay) and information associated with the measurement (\eg frequency band,  location, time, call drop probability \etcfree). Cellular performance maps (\aka signal maps) 
 consist of a large number of KPIs in several locations; an example is depicted on Fig. \ref{fig:rsrpmap_eastnyc_x540}. They are of immense importance to operators for planning, managing and upgrading their networks. 

Traditionally, cellular operators collected such measurements by hiring dedicated vans (\aka wardriving~\cite{yang:10}) with special equipment, to drive through, measure and map the received signal strength (RSS) in a particular area of interest. However, in recent years they increasingly outsource the collection of signal maps to third parties~\cite{alimpertis:19}. 
Mobile analytics companies, such as OpenSignal \cite{opensignal:11} and Tutela \cite{tutela}, crowdsource measurements directly from end-user devices, via standalone  mobile apps, or measurement SDKs integrated into other apps, such as games, utilities or streaming apps.
Either way, signal strength maps are expensive for both \mno{}~
 and crowdsourcing companies to obtain, 
and may not be available for all locations, times, frequencies, and other parameters of interest. The upcoming dense deployment of small cells at metropolitan scales will only increase the need for accurate and comprehensive signal maps to enable 5G network  management~\cite{5gonapkpis,imran:14}.

\begin{figure*}[t!]
	\centering
	{\includegraphics[scale=0.3]{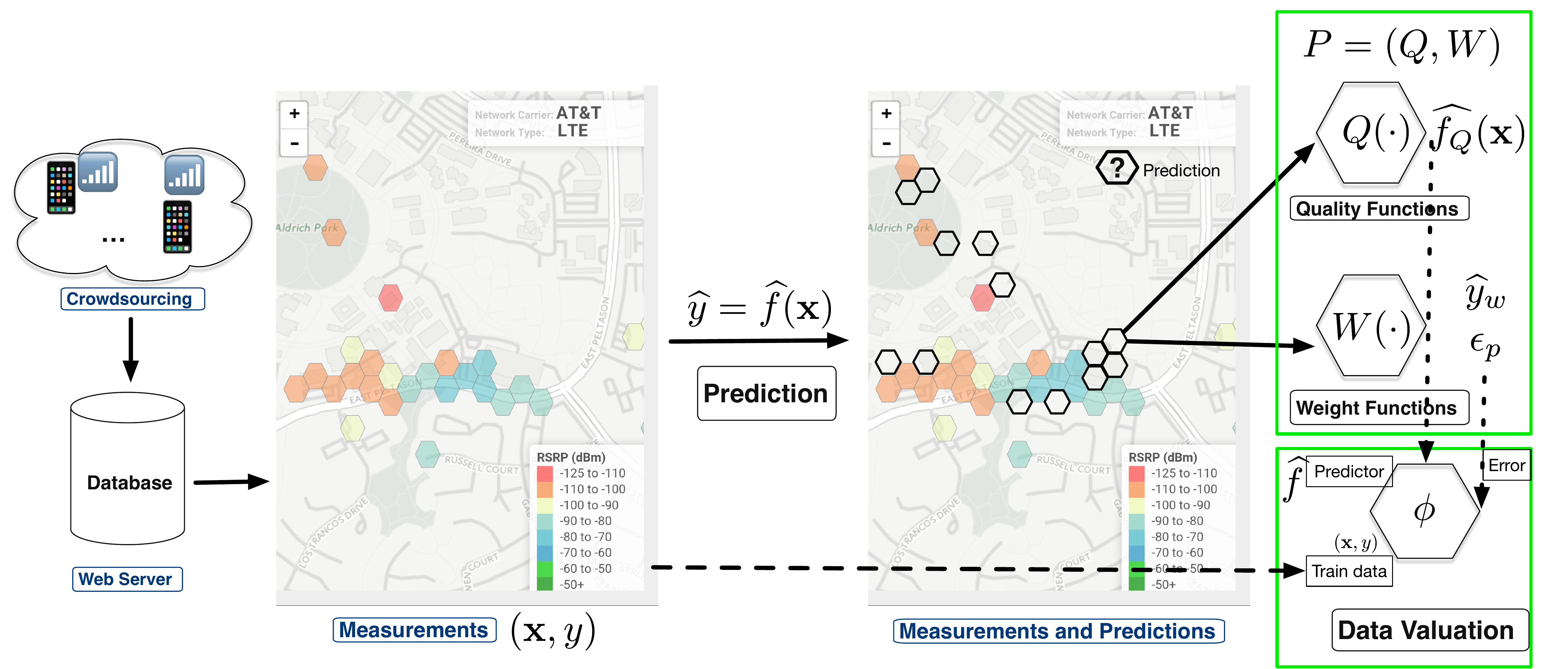}}
	\caption{{\bf Overview of our framework:} the goal is to predict cellular values $\hat{y}$, based on features $\mathbf{x}$, while training on available (and unequally important) measurements. As base workhorse ML model, we use our state-of-the-art RF-predictor (Sec.~\ref{sec:rfsprediction}), but any other predictor would do as well. Our framework builds on top of the base predictor and deals with the different importance of data points in three distinct ways.  \textbf{(1}) We use quality-of-service functions $Q$ to predict directly KPIs of interest; these depend on but are different from raw signal strength RSRP (Sec.~\ref{sec:qualityfunctions}). \textbf{(2)} We use weight functions $W$ to target the mismatch between  sampling and target distributions (Sec.~\ref{sec:importancesampling}).  \textbf{(3)} We  apply the Data Shapley framework to assess the importance (\ie predictive value) $\phi$ of the measurements w.r.t. the prediction task  at hand $(P,W)$  (Sec.~\ref{sec:shapley-theory}). 
	\label{fig:paper_oveview}
	}
\end{figure*}

For these reasons, there has been significant interest in signal map prediction techniques based on a limited number of spatiotemporal cellular measurements. These include  propagation models~\cite{raytracing:15, winnerreport}, data-driven approaches~\cite{fidaZipWeave:17,specsense:17,heBCS:18} and combinations thereof~\cite{phillips:12}. Increasingly sophisticated machine learning models are being developed to capture various spatial, temporal and other characteristics of signal strength~\cite{ray:16,raik:18,alimpertis:19} and throughput~\cite{infocomLSTM:17,milanMobihocDeepnets:18}.  Prior work has focused exclusively on minimizing the mean squared error (MSE) in predictions of raw signal strength (RSRP). In our prior work ~\cite{alimpertis:19}, we developed a Random Forests (\RFs)-based  predictor, and we showed that it outperforms state-of-the-art RSRP predictors. In this paper, we use this RFs-predictor as our base "workhorse" ML model and we develop a framework on top of it, to deal with the fact that not all measurements are equally important.

We observe that three different factors affect the importance of  measurement data points when those are used for training ML predictors. First, {\em what} KPI we predict: operators are typically interested in performance metrics such as coverage, call drop probability, number of bars; these depend on but go beyond raw signal strength (RSRP). Second, {\em where} we make the prediction: the operators may be interested in predicting performance better in some  locations (\eg those with weak coverage or at important  sites), while they may have no control on how crowdsourced mobile measurements are distributed. Third, since measurements are expensive and may pose privacy risks, we may want to identify those  data points with the highest predictive {\em value} and discard outliers or redundant measurements.

 In order to  address the aforementioned challenges, we develop a unified framework for predicting cellular performance maps\footnote{We refer to all cellular performance maps collectively and for simplicity as ``cellular'' or ``signal'' maps. Examples include maps of: RSRP or other KPIs, coverage indicator, number of bars, all of which are ultimately functions of cellular signal strength. The terms mobile coverage, signal strength and performance maps are used interchangeably in prior art (\eg~\cite{alimpertis:19, fidaZipWeave:17, raik:18}) and throughout this paper.},
 which provides cellular operators and mobile analytics companies with knobs to express and deal with the unequal importance of available cellular measurements.
We define two classes of functions $Q$ and $W$, that jointly define the performance of the signal maps prediction problem \problemspace. 
 These functions tackle the mismatch between: \textbf{(1)} the operators’ quality (QoS) metrics  of interest and the raw signal strength (RSRP) and \textbf{(2)} the sampling and target distributions, respectively. In addition,  \textbf{(3)} we compute the data Shapley values ($\phi$) of measurement data points that capture their importance for training a predictor for the particular cellular map prediction problem \problemspace ~at hand. Our three contributions are summarized on Fig.~\ref{fig:paper_oveview} and are further elaborated upon next.

\textbf{(1) Quality Functions $Q$.} 
We  consider \emph{quality-of-service functions ($Q$)}, based on signal strength, which specify {\em what metric} operators and users care about, such as mobile coverage indicators ($Q_c$), call drop probability ($Q_{CDP}$), and number of bars ($Q_B$).  Prior work exclusively minimizes the MSE for signal strength (RSRP)  ~\cite{raik:18, fidaZipWeave:17, heBCS:18,phillips:12}, which does not directly optimize prediction for the aforementioned metrics. 
In contrast, we show that  {\em learning directly on the $Q$ domain} can significantly improve performance  vs. state-of-the art, including: (i) recall gains from 76\% to 92\% and balanced accuracy gains from 87\% to 94\% for predictions of coverage loss (where false negatives are costly to operators); and (ii) reductions in relative error in predicted CDP by up to 32\% (in the high CDP regime of greatest concern to cellular \mno). Even for predicting {\em signal strength (RSRP) itself},  accuracy is often more critical in the low 
 than in the high signal-strength regime.\footnote{For example,  errors of several dB  have little impact when RSRP is in the range of -50 to -60 dBm (Fig.~\ref{fig:empirical_cdp_vs_kpis}) but significantly affect accuracy in the -120 to -125 dBm range (\ie very weak coverage).}
Specifying objectives via quality functions implicitly tunes the loss function, and allows operators to put more emphasis on values and use cases of signal maps that matter most. For example, we show improvement of prediction of RSRP itself in the low signal strength regime, of up to 27\% ($3$dB in RMSE).

\textbf{(2) Weight functions $W$.}
Sampling bias is inherent in crowdsourced data due to the non-uniform population density, as well as the commute and usage patterns.
An operator may be interested in knowing KPIs at particular locations (as well as times, frequencies, and other features). Examples of target locations of interest for prediction include: locations where there is poor coverage, locations with dense user population, origins of calls to 911 dispatchers, client sites during working hours.  However, these target locations may differ from the sampling locations where measurements are available\footnote{For example, in Fig.~\ref{fig:rsrpmap_eastnyc_x540},  measurements obtained through crowdsourcing lead to over-sampling highways compared to nearby  residential blocks  (as shown on Fig.~\ref{fig:sampling_vs_reweighting_external}).}, thus optimization to available data can lead to biased inference.  This mismatch of the sampling distribution with the target distribution is also known as the dataset shift problem~\cite{datasetshiftNips:19} in ML. To tackle this mismatch, we propose a re-weighting method, rooted in the framework of importance sampling, which leads to unbiased error. We introduce \emph{weight functions ($W$)} that allow operators to express {\em where} (\eg in which  particular locations, times, \etcfree) they are interested in predicting performance most accurately. We demonstrate improvement up to 20\%  
 for two intuitive target distributions: uniform loss across a spatial area and loss proportional to population density. Combining weight targeting and quality functions improves further, \eg up to 5\% more for estimating CDP on targeted spatial losses.

\textbf{(3) Data Shapley $\phi$.} 
We recognize and exploit the fact that not all measurements are equally important for predicting the metric of interest at the location of interest. We apply, for the first time in the context of signal maps, the {\em Data Shapley}  framework, originally defined in economics and recently adapted to assign value to training data in ML~\cite{dshapIcml:2019}. Our Data Shapley framework takes as input the available cellular measurements, the ML prediction algorithm, and the error metric, re-weighted for the particular prediction problem $P=(Q,W)$; it then computes the Shapley values ($\phi$) of individual measurements used for training the ML model. This enables us to then remove measurements with negative or low data Shapely values (\eg outliers or irrelevant for the specific task) that improves prediction, and can also guide data minimization (thus improve privacy).
 For example, we show that we can remove up to 65-70\% of data points, while simultaneously improving the recall of cellular coverage indicator from 64\% to 99\%.

Throughout the paper, we leverage two types of {\em large, real-world LTE  datasets} to gain insights and evaluate prediction performance: (i) a dense Campus dataset, we collected on our own university campus; and (ii) several sparser city-wide (in~\externalcities) datasets, provided by a mobile data analytics company. 

The rest of the paper is organized as follows. Section \ref{sec:relatedwork} reviews related work. Section \ref{sec:objectivesprediction} presents our prediction framework, including the baseline predictor (Sec.~\ref{sec:rfsprediction}) and the methodologies that deal with unequal importance of measurements, \ie the quality functions (Sec.~\ref{sec:qualityfunctions}), the weight functions (Sec.~\ref{sec:importancesampling}), and the data valuation (Sec.~\ref{sec:shapley-theory}). Section \ref{sec:results} presents evaluation results, based on the aforementioned datasets. Section \ref{sec:conclusion} concludes the paper. The Appendix (in Supplemental Materials) provides details on the Data Shapley formulation (A1) and on the choice of the base predictor (A2).

\section{Our Work in Perspective}\label{sec:relatedwork}
Broadly speaking, signal strength can be predicted by propagation models or data-driven approaches, including geospatial interpolation, (\eg ~\cite{fidaZipWeave:17, phillips:12}) and  machine learning (\eg~\cite{raik:18,alimpertis:19}); or combinations thereof:~\cite{phillips:12}.

{\em Propagation models.} State-of-the-art propagation and path loss (equa\-tion-based) models include WINNER I/II~\cite{winnerreport}, Ray tracing~\cite{raytracing:15} and others. However, this family of models requires a detailed map of the environment and fine grained tuning of model's parameters~\cite{alimpertis:19}. A simple yet widely used propagation model is LDPL~\cite{alimpertis:14} and its indoor variant~\cite{applegate:18}.

{\em Geospatial interpolation} \cite{specsense:17,phillips:12,fidaZipWeave:17}.  Methods such as Ordinary Kriging (OK) and OK Detrending~\cite{phillips:12}, which are used by the ZipWeave~\cite{fidaZipWeave:17} and SpecSense~\cite{specsense:17} frameworks, cannot naturally incorporate additional features beyond location (such as time, frequency, hardware \etc), as our ML models do (both in this paper and in our prior work~\cite{alimpertis:19}).

 {\em Machine learning: DNNs.} Examples of RSRP/RSS prediction include~\cite{raik:18} and~\cite{heBCS:18}. Work in~\cite{raik:18} uses DNNs along with detailed 3D maps from LiDAR and work in~\cite{heBCS:18} uses Bayesian Compressive Sensing (BCS).  ML models for the signal strength likelihood for user localization have also been developed by prior art~\cite{ray:16, margolies:17, GPsRss:2006}. Krijestorac \etal \cite{cnnUCLA} propose state-of-the-art CNNs for estimating signal strength values and utilize a 3D map of the environment as features. In particular, they treat signal strength as a Gaussian random variable and predict its mean and variance. The evaluation is based on simulated data via ray-tracing software, 
which allows continuous infill of regions. However, when trained and evaluated on real-world datasets, this approach faces the limitation of sparse data.  \cite{autoencodersSignalMaps} uses autoencoders for predicting signal strength, but similarly their method is limited to simulated data.

{\em Machine learning: Random Forests.} In our prior work~\cite{alimpertis:19}, we proposed random forests (\RFs) for predicting signal strength (RSRP) based on a number of features, including but not limited to location and time (see Section~\ref{sec:rfsprediction}). We demonstrated that \RFs~can outperforms prior art (including model-based and geospatial methods) because they can inherently capture spatial and temporal correlations, while also naturally extending 
to features beyond location (geospatial predictors can only handle location features)~\cite{alimpertis:19}. For instance, for MSE minimization, the \RFs~ requires 80\% less measurements for the same prediction accuracy, or reduces the relative error by 17\% for the same number of measurements compared to geospatial interpolation methods used by ZipWeave~\cite{fidaZipWeave:17} and SpecSense~\cite{specsense:17}. In Section \ref{sec:numerical_results_baseproblem} and Appendix A2, we also compare \RFs~ to recent DNN and CNN-based predictors, and we show that \RFs~still outperform alternatives.

 {\em Other Metrics - Dealing with unequal importance.} Recent work in QoS has considered how RSRP measurements could be used as  proxy to predict Quality of Experience (QoE)~\cite{adarshtoo} or reinforcement-learning prediction techniques for video playback~\cite{pensieveQoE:2017}. Apart from signal strength, prior work considered mobile traffic volume maps (\ie KPI throughput)~\cite{milanMobihocDeepnets:18, infocomLSTM:17, Lumos5G}, but solely focusing on the underlying ML model and MSE minimization.  Importance sampling for deep neural network training is considered in~\cite{deepnetsimportancesampling}. The mismatch between sampling and target distributions is also related to the dataset shift problem~\cite{datasetshiftNips:19} in machine learning.

{\em Contributions.} In this paper, we use our own implementation of \RFs-predictor as a state-of-the-art ``work horse'' signal map predictor. For completeness, we present the method and its evaluation in Sec.~\ref{sec:rfsprediction} and Appendix~2. However, our focus here is orthogonal: we propose a framework that combines three methodologies ($Q, W, \phi$) to  deal with the unequal importance of available measurements for a particular prediction problem $P=(Q,W)$. Our framework builds on top of \RFs,  but could also utilize any other underlying ML algorithm minimizing a squared-error loss. To the best of our knowledge, this is the first paper to leverage QoS  to improve prediction of those metrics but also of RSRP itself in regimes where it matters; and to develop a ($Q,W$)-specific data Shapley valuation  for data minimization and  prediction improvement.

A summary of prior approaches for signal map prediction, and a quantitative comparison to our framework, is presented in Table~\ref{tab:table-compare-approaches}. There is an increasing amount of work on mobile crowdsourced data privacy recently, with applications including signal strength, radiation maps, \etc ~\cite{boukorosLack:2019}. In other work from our group~\cite{bakopouloulocation:2021}, we have also considered distributed settings and privacy-utility tradeoffs. Location privacy is a large research area on its own and out-of-scope for this paper. Here we focus on optimizing prediction, in a centralized setting, given measurements of unequal importance. This being said, the data minimization enabled by the Shapley framework is a useful tool for improving privacy.

\begin{table}[]
  \centering
 
  \setlength\tabcolsep{1pt} 
  \scriptsize
\begin{tabular}{|l|c|c|c|c|c|c|c|c|}
\hline
                                                                                         & \multicolumn{4}{c|}{\textbf{Features}}                                                                                                                                                & \multicolumn{3}{c|}{\textbf{\tiny{Environment, Scale, Data}}}                                                                                                                 & \multicolumn{1}{l|}{\textbf{\tiny{Evaluation}}}                                                     \\ \hline
                                                                                         & Spatial    & Time       & \begin{tabular}[c]{@{}c@{}}Device,\\ Network\end{tabular} & \multicolumn{1}{l|}{\begin{tabular}[c]{@{}l@{}}LiDar,\\  3D Maps\\ Agnostic\end{tabular}} & \begin{tabular}[c]{@{}c@{}}{\tiny Environment}\\ Agnostic\end{tabular} & \begin{tabular}[c]{@{}c@{}}City\\ Wide\end{tabular} & \begin{tabular}[c]{@{}c@{}}Real\\ Data\end{tabular} & \multicolumn{1}{l|}{\begin{tabular}[c]{@{}l@{}}Quality \\ Beyond\\ RSRP\end{tabular}} \\ \hline
\begin{tabular}[c]{@{}l@{}}Log-Distance \\ Path-Loss\\ ($LDPL$)\end{tabular}             & \gCheck & \redX      & \redX                                                     & \gCheck                                                                                      & \gCheck                                                     & \redX                                               & \gCheck                                                & \redX     \\ \hline
\begin{tabular}[c]{@{}l@{}}COST-231/\\ WINNER I-II/\\ Ray Tracing\end{tabular}           & \gCheck & \redX      & \gCheck                                                & \redX                                                                                           & \redX                                                          & \redX                                               & \gCheck                                                & \redX                                                                                        \\ \hline
\begin{tabular}[c]{@{}l@{}}Geostatistics\\\tiny{Specsense\cite{specsense:17}}\end{tabular} & \gCheck & \redX      & \redX                                                     & \gCheck                                                                                      & \redX                         & \redX                                               & \gCheck                                                & \redX                                                                                        \\ \hline
BCS \cite{heBCS:18}                                                                      & \gCheck & \gCheck & \redX                                                     & \gCheck                                                                                      & \redX                                                          & \redX                                               & \gCheck                                                & \redX                                                                                        \\ \hline
\begin{tabular}[c]{@{}l@{}}DNNs\\ (RAIK \cite{raik:18})\end{tabular}                     & \gCheck & \redX      & \redX                                                     & \redX                                                                                           & \redX                                                          & \gCheck                                          & \gCheck                                                & \redX                                                                                        \\ \hline
CNNs \cite{cnnUCLA}                                                                      & \gCheck & \redX      & \redX                                                     & \redX                                                                                           & \gCheck                                                     & \redX                                               & \redX                                                     & \redX                                                                                        \\ \hline
\begin{tabular}[c]{@{}l@{}}Auto-\\ Encoders \cite{autoencodersSignalMaps}\end{tabular}    & \gCheck & \redX      & \redX                                                     & \redX                                                                                           & \gCheck                                                     & \redX                                               & \redX                                                     & \redX                                                                                        \\ \hline
\textbf{Our Framework}                                                                        & \gCheck & \gCheck & \gCheck                                                & \gCheck                                                                                      & \gCheck                                                     & \gCheck                                          & \gCheck                            & \gCheck                                                             \\ \hline
\end{tabular}
 	\caption{Comparison of our framework to other approaches for signal map prediction. 
 	\label{tab:table-compare-approaches}} 
\end{table}

\section{Prediction Framework}\label{sec:objectivesprediction}

\subsection{Signal Maps Prediction}\label{sec:problemStatementSystemModel}

\subsubsection{Definitions and Problem Space}

An observed \textbf{signal  map} is a collection of $N$ measurements $(\mathbf{x_i},y_i)$, $i=1,2...N $, where $y_i = \{$\rsrp, \rsrq, \cqi $\}$ denotes the KPI of interest given the feature vector $\mathbf{x}_i$ (\eg location \etcfree) w.r.t. which the signal is to be mapped.  In general, an operator's interest is not only in the observed signal strength map, but in an underlying ``true'' signal strength map, defined by the conditional distribution $Y|\mathbf{x}$ for an arbitrary $\mathbf{x} \in \mathbb{X}$, where $Y$ is the (generally unobserved) KPI at $\mathbf{x}$ and $\mathbb{X}$ specifies a region of interest (e.g., an areal unit, time period, \etcfree).  This suggests approximation of the true signal strength map by machine learning (ML), where our goal is to answer queries regarding $Y|\mathbf{x}$, or functions thereof, by training a predictor on the observed data.

\textbf{$y$: Key Performance Indicators (KPIs).} 
There are many KPIs for LTE defined by 3GPP: 
\begin{itemize}
\item 
{\em RSRP (\rsrp):}  The reference signal received power is the average over multiple reference and control channels, reported in dBm. It is of great importance for LTE and utilized for cell selection, handover decisions, mobility measurements \etc

\item
{\em RSRQ (\rsrq):} The reference signal received quality is a proxy to measure channel's interference. 
\item
{\em CQI (\cqi):} The channel quality indicator is a unit-less metric (\cqi = $\{0, \cdots, 15\}$)  of the overall performance. 
\end{itemize}
It is worth noting, that {\em all} prior work focused exclusively on predicting RSRP. Therefore, we use $y=\text{\rsrp}$, to refer to prediction of RSRP, unless otherwise noted.

\textbf{$\mathbf{x}$: Measurement Features.} For each measurement $i$ in our datasets, we use several features available via Android APIs \cite{alimpertis:19}.
$ \mathbf{x_i^{full}} = (l_i^x, l_i^y , d, h, \text{\cid}, \text{\dev}, out, ||\mathbf{l}_{BS} - \mathbf{l}_i ||_2, \text{\freq})$.
We consider the following features:
\begin{itemize}
    \item 
{\em Location:}  $\mathbf{l}_j= (l_i^x, l_i^y)$: GPS's latitude and longitude.
\item
{\em Time:}  $\mathbf{t_i} = (d, h)$, where $d$ is the weekday and $h$ the hour the measurement was collected. 
\item
{\em  Cell ID and \lteta:} KPIs are defined per serving LTE cell, 
which is uniquely identified by the CGI (cell global identifier, cell ID or \cid) which is the concatenation of the following identifiers: the MCC (mobile country code), MNC (mobile network code), TAC (tracking area code) and the cell ID.  LTE also defines Tracking Areas, \lteta, by the concatenation of  MCC, MNC and TAC, to describe a group of neighboring cells, under common LTE management.
\item
{\em Device hardware} type (\dev): This refers to the device model (\eg Galaxy s21 or iPhone 13) and \emph{not} to device identifiers.
\end{itemize}
In~\cite{alimpertis:19}, we considered all features and showed the most important ones to be location, time, cell \cid~and device hardware information~\dev. In this paper, we consider $\textbf{x}$ to be the full set of features, unless otherwise noted.

\begin{table}[t!]
		\centering
	\setlength\tabcolsep{1.5pt} 

    
	\footnotesize
	
	\begin{tabular}{l|l|l|}
	    
		\cline{2-3}
		& Notations                 & Definitions-Description                                                  \\ \hline
		\multicolumn{1}{|l|}{\multirow{5}{*}{\textbf{Data}}}                                                                       & $\mathbf{x}$              & Measurement's Features                                                   \\ \cline{2-3} 
		\multicolumn{1}{|l|}{}                                                                                            & $y$   & Label - KPI (Key Performance Indicator)                                  \\ \cline{2-3} 
				\multicolumn{1}{|l|}{}                                                                                            & $\{y^P, y^{I}, y^{C}\}$   & KPIs in this Paper                                \\ \cline{2-3} 

		\multicolumn{1}{|l|}{}                                                                                            & $y^P$                     & RSRP: Received Signal Reference Power                                    \\ \cline{2-3} 
		\multicolumn{1}{|l|}{}                                                                                            & $y^{I}$                   & RSRQ: Received Signal Reference Quality              \\ \cline{2-3} 
		\multicolumn{1}{|l|}{}                                                                                            & $y^{C}$                   & CQI: Channel Quality Indicator                                           \\ \hline
		\multicolumn{1}{|l|}{\multirow{3}{*}{\begin{tabular}[c]{@{}l@{}}Network\\ Quality\\ Functions\end{tabular}}}      & $Q(y)$                    & Network Quality Function               \\ \cline{2-3} 
		\multicolumn{1}{|l|}{}                                                                                            & $Q_c(y^P)$                & Mobile Coverage Indicator                                                \\ \cline{2-3} 
		\multicolumn{1}{|l|}{}                                                                                            & $Q_{cdp}(y)$              & Call Drop Probability                                                    \\ \hline
		\multicolumn{1}{|l|}{\multirow{2}{*}{\begin{tabular}[c]{@{}l@{}}Error / Loss\\ Scores\end{tabular}}}              & $L(\widehat{y},y)$        & Loss function; squared loss in this work            \\ \cline{2-3} 
		\multicolumn{1}{|l|}{}                                                                                            & $\varepsilon_p$           & Reweighted Error Metric for Target \targetdistr \\ \hline
		\multicolumn{1}{|l|}{\multirow{7}{*}{\begin{tabular}[c]{@{}l@{}}Importance\\ Sampling \\ Framework\end{tabular}}} & \targetdistr           & Target distribution                                                      \\ \cline{2-3} 
		\multicolumn{1}{|l|}{}                                                                                            & $s(\mathbf{x})$           & Sampling Distribution                                                    \\ \cline{2-3} 
		\multicolumn{1}{|l|}{}                                                                                            & \population          & Population Distribution                                                  \\ \cline{2-3} 
		\multicolumn{1}{|l|}{}                                                                                            & $u(\mathbf{x})$           & Unifom Distribution                                                      \\ \cline{2-3} 
		\multicolumn{1}{|l|}{}                                                                                            & $W(\mathbf{x})$           & Weighting Function                                                       \\ \cline{2-3} 
		\multicolumn{1}{|l|}{}                                                                                            & \wu                & Importance Ratio for  Uniform \targetdistr                 \\ \cline{2-3} 
		\multicolumn{1}{|l|}{}                                                                                            &\wP             & Importance Ratio  for Population  \targetdistr         \\ \hline
	\end{tabular}
	\caption{{\footnotesize Terminology and Notation.} 	\label{tab:notations_summary}}
\end{table}

\textbf{Signal Map Prediction.}
Our goal is to predict an \emph{unknown} signal map value $\hat{y}$ at a given location and other features ($\mathbf{x} \in \mathbb{X}$), based on available spatiotemporal measurements with labeled data $(\mathbf{x}_i, y_i)$, $i=1,...N$, either in the same \cid~or in the same \lteta. 
The real world underlying phenomenon 
 is a complex process $y=f(\textbf{x})$ that depends on $\mathbf{x}$, and characteristics of the wireless environment.

It is important to consider the \emph{loss} to be minimized by the choice of predictor: certain loss functions improve performance for certain objectives, while degrading it in others.  
We consider two factors relating to the choice of loss. 
First, an operators' interest is not always in KPI $y$ itself, but in some \emph{quality of service function}, $Q(y)$; see Section \ref{sec:qualityfunctions} for concrete examples.  While conventional training schemes focus on predicting $y$ (\eg w.r.t. mean squared error, or MSE), we consider signal map prediction that minimizes error in the predicted value of $Q(y)$ itself.  We demonstrate that the nonlinear dependence of quality-of-service on raw signal strength makes this direct approach superior for many practical applications.
Second, the operator may wish to weight accuracy for some values of $\mathbf{x}$ more heavily than others.  While conventional training schemes implicitly assume that importance corresponds to data sampling frequency, we instead consider optimization w.r.t. an application-specific weight function $W(\mathbf{x})$ that may or may not closely correspond to the distribution of sampled observations; see Table 3 and Section \ref{sec:importancesampling} for details.

\begin{table}
	\centering
	\setlength\tabcolsep{6pt} 
	\begin{tabular}{|l||l|l|}
		\hline
		\textbf{Training Options}                                                                           & \textbf{$y$ Domain}                             & \textbf{$Q$ Domain}                 \\ \hline \hline
		\multirow{2}{*}{\textbf{\begin{tabular}[c]{@{}l@{}}same $w = 1$ \\ for all points\end{tabular}}}    & \baseproblem                                   & \problem{$Q,k$}                     \\ \cline{2-3} 
		& $\widehat{y} \rightarrow$ $Q(\widehat{y})$      & $ Q(y) \rightarrow \widehat{Q}(y)$  \\ \hline
		\multirow{2}{*}{\textbf{\begin{tabular}[c]{@{}l@{}}training weights $w$ \\ (Table 3)\end{tabular}}} & \problem{$I,W$}                                 & \problem{$Q,W$}                     \\ \cline{2-3} 
		& $\widehat{y_w} \rightarrow$  $Q(\widehat{y_w})$ & $Q(y) \rightarrow \widehat{Q}^w(y)$ \\ \hline
	\end{tabular}
	\caption{{Overview of prediction problems $P=(Q,W)$. One can perform prediction on the y (signal)  or on the Q (quality) domain. One can assign the same or different weights to different points. 
	\label{tab:customfunctions_methodology_v2}}}
\end{table}

{\bf Prediction Problem Space: $P=(Q,W)$.}  
With the above motivation, we may formalize the prediction problem as follows.
 Let $\mathbb{Y}$ be the space of potential KPI values. $Q \in \mathbb{Q}: \mathbb{Y} \mapsto \mathbb{R}$ is a quality function, as described above.   Similarly, $W \in \mathbb{W}: \mathbb{X} \mapsto \mathbb{R}^+_0$ is the above described weight function. We define the \emph{prediction problem space} as $\mathbb{P} = \mathbb{Q} \times \mathbb{W}$, whose elements $P=(Q,W) \in \mathbb{P}$ are \emph{prediction problems}. 
This representation provides a simple unifying formalism for a range of different problems.  For instance, note the \emph{base problem} $P_B=(I,k)$ where $I$ is the identity function and $k$ is a constant function, which amounts to the conventional learning problem assumed in the prior literature.  Here, we develop not only predictors which minimize loss under $P_B$, but also or other arbitrary $P \in \mathbb{B}$.  In practice, this amounts to finding predictors $\widehat{y} = \widehat{f}_y(\mathbf{x})$ for signal strength (\eg LTE RSRP) as well as  $\widehat{Q}(y) = \widehat{f}_Q(\mathbf{x})$ for quality functions $Q$, where $\widehat{f}_y(\mathbf{x})$ and $\widehat{f}_Q(\mathbf{x})$ are optimized w.r.t. an appropriate  weight function $W(\mathbf{x})$. Table \ref{tab:customfunctions_methodology_v2} provides a taxonomy of all the prediction problems our framework can address.

{\bf Transformation between problems.} 
Any method for solving the base problem $P_B=(I,k)$ can be transformed to solve an alternative prediction problem, $P=(Q,k)$, by training on $Q(y)$ instead of $y$;  we pursue this in Sec.~\ref{sec:qualityfunctions}. 
Likewise, we can transform a procedure for solving $P_B$ to a procedure for solving $P=(I,W)$ by applying importance sampling, as described in Section \ref{sec:importancesampling}.  
In addition, given a  problem $P=(Q,W)$, we may transform any procedure for solving $P_B$ to a procedure for solving $P$ by (i) training on $Q(y)$ via the methods of section~\ref{sec:qualityfunctions} and (ii) applying the importance reweighting of section~\ref{sec:importancesampling} using $W$.

\subsection{Base Predictor: Random Forests (\RFs)}\label{sec:rfsprediction}

There is a rich body of work on predictors of signal maps, reviewed in Sec.~\ref{sec:relatedwork}. For the purposes of this paper, we consider one such state-of-the art signal map predictor: Random Forest (\RFs) regression and classification.  This predictor is an ensemble of multiple decision trees~\cite{breiman:2001} and provides good trade-off between bias and variance by exploiting the idea of bagging~\cite{breiman:2001}.

{\textbf{Why \RFs~for data-driven prediction:} In our previous work~\cite{alimpertis:19}, we demonstrated the \RFs~advantages with city-scale signal strength map (RSRP) prediction. \RFs~naturally incorporate multiple features \vs just location in geostatistics, automatically produce correlated areas in the feature space with similar wireless propagation properties, scalable, minimal hyper-parameter tuning, they do not overfit and they require minimum amount of data. Examples of decision boundaries produced by \RFsSpatial~(for LTE RSRP data) is depicted in Fig.~\ref{fig:RFs_xy_examplesplits}. One can see the splits according to the spatial coordinates (lat, lng) and the produced areas agree with our knowledge of the placement and direction of antennas on UCI campus (\eg notice the backlobe of the antenna). Essentially, these axis-parallel splits assume that measurements close in space, time most likely should be in the same tree node, which is a reasonable assumption for signal strength statistics. Automatically identifying these disjoint regions with spatiotemporally correlated RSRP comes for free to \RFs, and is particularly important in RSRP (and other KPI metrics) prediction due to wireless propagation properties. In contrast, prior art (\eg OKP,~\cite{fidaZipWeave:17, specsense:17}) requires additional pre-processing for addressing this spatial heterogeneity.

 \begin{figure}[t!]
 \centering
	\subfigure{\includegraphics[scale = 0.12]{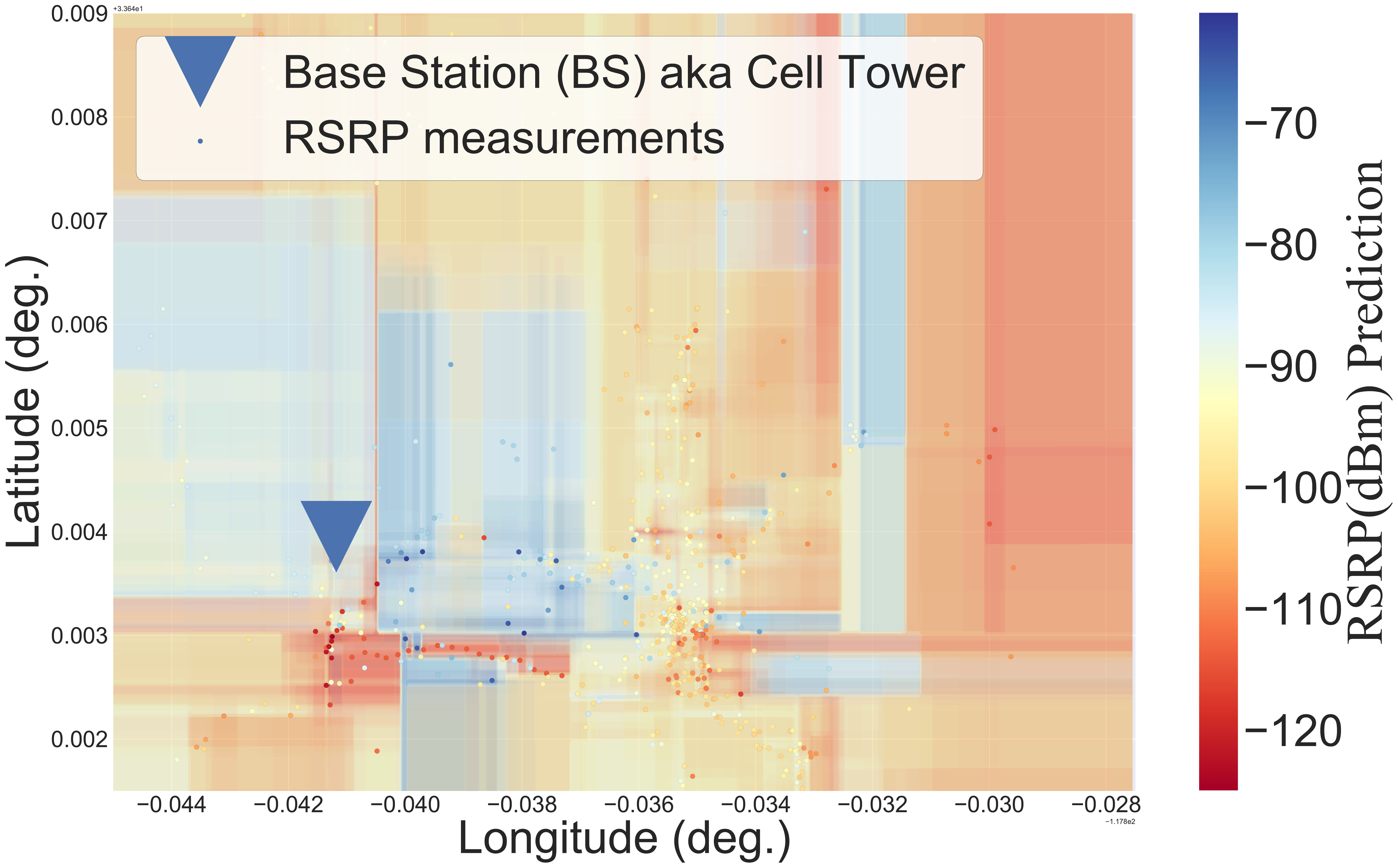} \label{fig:RFs_xy_examplesplits_x306}}
	\caption{Example of decision boundaries chosen by \RFsSpatial~for \inhouse cell x306. We can see that \RFs~ can naturally identify spatially correlated measurements,  \ie regions with similar wireless propagation characteristics. For example, note how the model identifies the antenna's backlobe and directionality.}
	\label{fig:RFs_xy_examplesplits} 
\end{figure}

 In~\cite{alimpertis:19}, we introduced this predictor for the first time for signal maps, and we showed its superiority compared to state-of-the-art including model-based, spatiotemporal and DNN predictors. Section \ref{sec:numerical_results_baseproblem}, including Table \ref{tab:campus_allcells}, shows that \RFs~outperforms alternatives on the \inhousedataset. Due to lack of space, we defer the detailed evaluation of the \RFs~ to Appendix~\ref{sec:appendeix_baseproblem_results}2 and  \cite{alimpertis:19,alimpertis:2020}. In this paper, we use \RFs~ as the underlying ML model for our framework. However, we emphasize that the ideas of our framework build on top and can be combined with any other learning strategy that can be applied on square-integrable real functions.

{\bf RFs predictor:} Under the \emph{base problem}, $P_B$, signal map value $y\in \mathbb{Y}$  to be estimated can be modeled as follows, given a set of feature vectors $\mathbf{x}\in\mathbb{X}$.
$ y | \mathbf{x} \sim \mathcal{N} (RFs_{\mu}(\mathbf{x}) , \sigma^2_{\mathbf{x}})\label{eq:RFs_kpimodel}$, 
where $RFs_{\mu}(\mathbf{x})$, $RFs_{\sigma}(\mathbf{x})$ are the mean and standard deviation respectively of the \RFs~predictor ($\equiv \widehat{f}_y(\mathbf{x})$). The total variance of the prediction is $\sigma^2_{\mathbf{x}} =  RFs_{\sigma}(\mathbf{x}) + \sigma^2_{RFs} \label{eq:RFs_var_model}$, where $\sigma^2_{RFs} $ is the  error from the construction of the \RFs~ itself. 
The final prediction $\widehat{y} = \widehat{f}_y(\mathbf{x}) = RFs_{\mu}(\mathbf{x})$  is the maximum likelihood estimate.


\subsection{Quality of Service  Functions ($Q(y)$)}\label{sec:qualityfunctions}
As described above, \emph{QoS function}, $Q$, is a function of KPI $y$ that reflects  an outcome that depends on $y$. Examples of QoS of interest to cellular operators  include the following.

{\em Call Drop Probability (CDP)}. One of the most important cellular network quality metrics is the call drop probability.  We  model CDP with the exponential function, $Q_{cdp}(y)  = a e^{-b y} + c$, with parameters $a,b,c$ estimated using  empirical data from the literature~\cite{jia:15},~\cite{Iyer:17}. An example of CDP vs. KPIs (RSRP and CQI) is shown on Fig.~\ref{fig:empirical_cdp_vs_kpis}.  It is immediately apparent that nearly all of the variation in CDP occurs at signal strengths below $-100$ dBm, implying that signal strength errors at high dBm will have far less impact on predictions of $Q_{cdp}$ than errors of equal size at low dBm.  As a continuous outcome, call drop probability estimation \cdpPred{$y$} can be treated as a ML regression problem.

{\em LTE Signal Bars.} Absolute RSRP values $y$ are translated to the widely used signal bars $Q_B(y)$ on mobiles' screens. 
Mobile analytics companies usually produce 5-colors map to visualize signal bars~\cite{opensignal:11}. 
Despite variation across devices, typical values of \signalbars~for iOS and Android devices are: 
\begin{equation}
Q_B(y) =
\begin{cases}
0 & \text{if $y^P <= -124$ dBm}\\
1 & \text{if $ y^P \in [-123, -115])$ dBm}\\
2 & \text{if $ y^P \in [-114, -105])$ dBm}\\
3 & \text{if $y^P \in [-104, -85])$ dBm}\\
4 & \text{otherwise (\ie excellent reception)}
\end{cases}       
\end{equation}

\begin{figure}[t!]
	\subfigure[ \cdp{\rsrp}  \vs~ RSRP.]{\includegraphics[scale = 0.13]{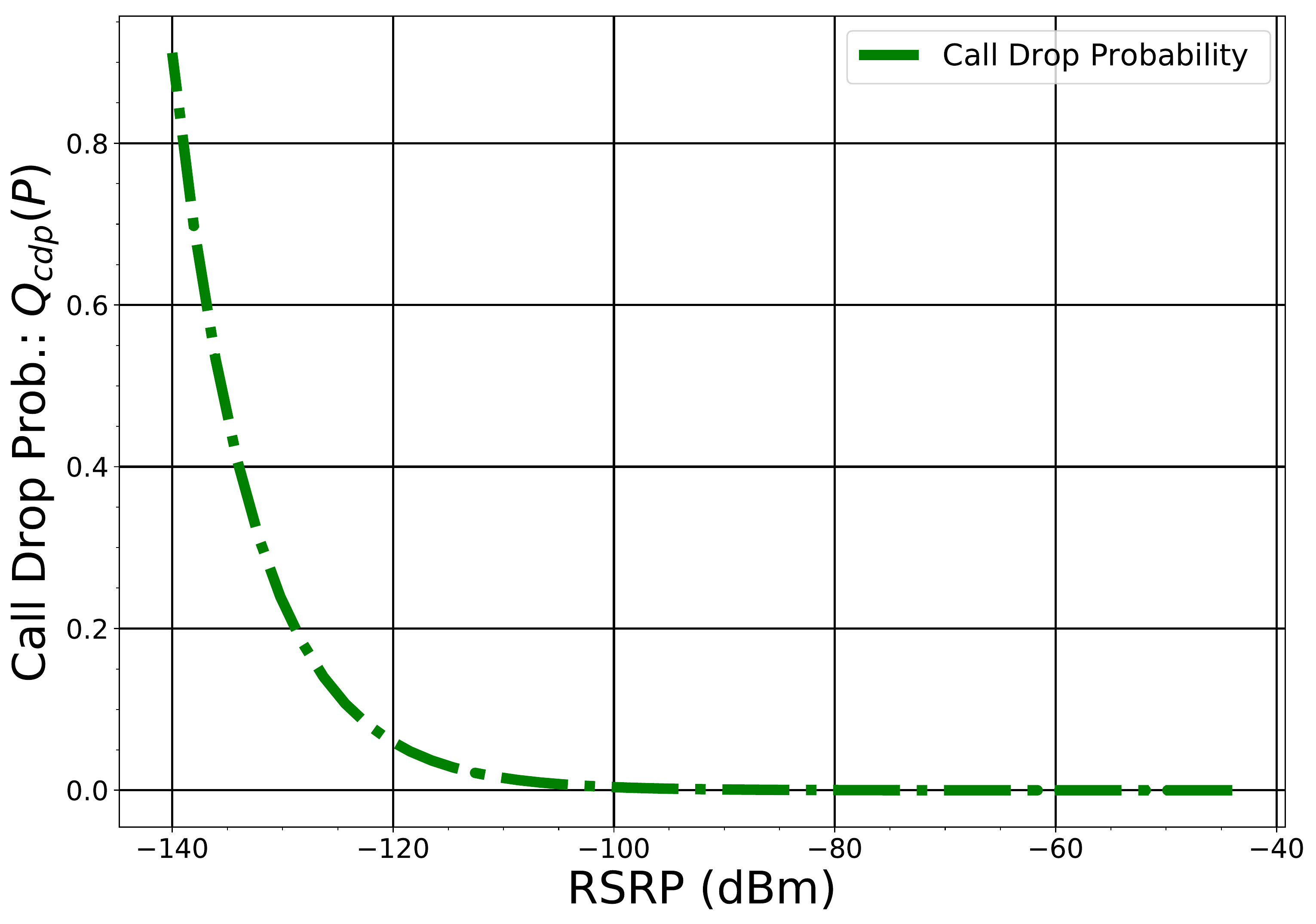} \label{fig:empirical_cdp_vs_rsrp}}
	\subfigure[ \cdp{\cqi}  \vs~ CQI.]{\includegraphics[scale = 0.13]{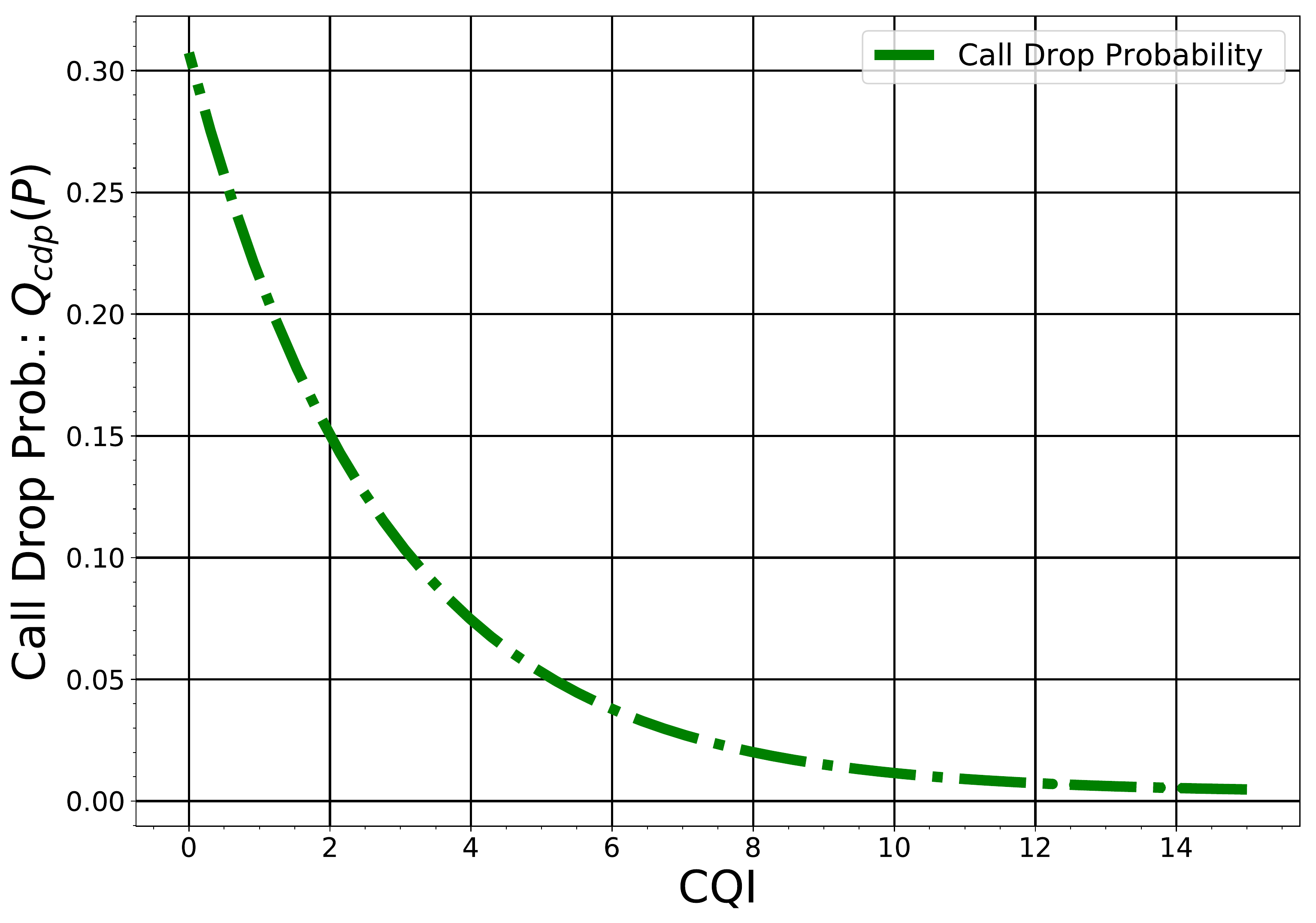} \label{fig:empirical_cdp_vs_cqi}}
	\caption{\footnotesize{ Call Drop Probability (CDP) \Qhat{$y$} as a function of KPIs.}\label{fig:empirical_cdp_vs_kpis} }
\end{figure}
{\em Mobile Coverage.} Detecting areas with weak/no signal (\emph{i.e.,} bad coverage) is a major problem for cellular operators. This is essentially a binary classification (per \cite{galindoCoverage:13}). We define the mobile coverage indicator as a function of RSRP for LTE~\cite{galindoCoverage:13}:
\begin{equation}
Q_c(y) =
\begin{cases}
0 & \text{if $ \text{\rsrp} <= -115$ dBm, \ie 0 or 1 bar}\\
1 & \text{otherwise}
\end{cases}       
\end{equation}
The rationale is that the call drop probability increases exponentially and the service deteriorates significantly below $-115$dBm~\cite{jia:15}.  
 We want to detect areas of bad coverage because undetected $Q_c(y) = 0$ could impact the operator's reputation, revenue and  overall performance.

\textbf{Minimizing the error that matters, not MSE.}  We show that prediction can be improved by training models directly on QoS observations \Qobj{$y$} and predicting \Qhat{$y$} instead of using the proxy \Qobj{$\widehat{y}$}; in other words, we minimize the error of ${f_{Q}}(\mathbf{x})$ instead of minimizing the error of ${f}_y(\mathbf{x})$.  This is equivalent to transforming the prediction problem from $P_B$ to some alternative $P$; intuitively, we are implicitly modifying the loss function used in estimation from one that treats errors at all $y$ values equally to one that emphasizes errors with practical consequences for cellular operators such as mis-identifying bad coverage areas or failing to predict areas with high call drop probability (\eg see Fig.~\ref{fig:empirical_cdp_vs_kpis}, \rsrp $ <= -100$dBm). Our experimental results in Section~\ref{sec:results} show how the prediction of bad coverage  ($Q_c(y) = 0$) can be improved for these regions that matter the most.

{\em Prediction of $\widehat{Q}$ using Random Forests.} We use \RFs~  to predict $\widehat{Q}$, similarly to predicting 
$\widehat{y}$ in the previous section.  Given prediction problem $P=(Q,k)$,
%
$Q(y) | \mathbf{x} \sim \mathcal{N} (RFs'_{\mu}(\mathbf{x}) , \sigma^2_{\mathbf{x}})\label{eq:RFs_Qmodel}$, 
%
where $RFs'_{\mu}$ is trained on quality-transformed observations $(\mathbf{x}_i,Q(y_i))$ instead of raw signal strength observations $(\mathbf{x}_i,y_i)$.  This simple procedural modification (using $\widehat{Q}(y)$ in place of $Q(\hat{y}(\mathbf{x}))$) allows us to transform the base problem to any element of $P$ involving constant weight function; this last restriction is lifted below.

\subsection{Importance Sampling (Weights $W$)}\label{sec:importancesampling}
We have argued above that not all values of $y$ are equally important from a QoS perspective.
 Similarly, not all inputs $\mathbf{x}$ are necessarily equally important. For instance, an operator may be particularly interested in accurate predictions in some times or locations, \eg 911 locations, health facilities, areas with high revenue or competitive advantage during business hours, areas with dense user population \etc We generally refer to points in the input space generically as ``locations''; however the dimensions involved may include space, time, frequency, device, \etcfree

Prior work~\cite{specsense:17,raik:18} has primarily minimized MSE for predicted signal strength via cross-validation (CV),  \ie report the error on held-out test data, after training on a sample of signal strength measurements.  This  implicitly assumes that all observations are equally important for both learning and evaluation, and, further, that the importance of error minimization in some subset of $\mathbb{X}$ is proportional to the number of observations in it.  These are strong assumptions that are often violated in practice.  For example, an operator might consider all locations within an areal unit having equal importance.  If, however, the available data is distributed according to population (which is highly uneven), then the weighting implicitly used in the analysis will be far from the desired (uniform) distribution.  By turns, an operator interested in population-weighted error may encounter problems when using data intensively collected by a small subset of users with residential locations or commuting patterns that are not reflective of the customer base.  Such mismatches between the \emph{sampling distribution} of signal strength observations in $\mathbb{X}$ and the \emph{target distribution} that captures the operator's desired loss function can be viewed as mismatches of the desired prediction problem $P(Q,W)$ versus the base problem $P_B$; to remove prediction bias, we show how direct estimation of $P=(Q,W)$ can be performed using  techniques from importance sampling.

\begin{table}[]
	\centering
	\setlength\tabcolsep{3pt} 
	\caption{{\small Importance sampling  for \mno(examples)}}
	\vspace{-5pt}
	\label{tab:importancesampling_methodology}
	{\small
	\begin{tabular}{|l|l|}
		\hline
		\begin{tabular}[c]{@{}l@{}}\textbf{Target Distribution}\\\textbf{ (Cellular Operator Objective)}\end{tabular} & \begin{tabular}[c]{@{}l@{}}\textbf{Importance Ratio}\\ \textbf{(Weights} $w_i$)\end{tabular} \\ \hline
		 Uniform distribution  $u(\mathbf{x})$                                           & $w_u \propto \frac{1}{s(\mathbf{x})}$                                      \\ \hline
		Population distribution   \population                                     & $\text{\wP} \propto \frac{\text{\population}}{s(\mathbf{x})}$                          \\ \hline
		Operator's custom target distr. $p(\mathbf{x})$                                & $w_c \propto  \frac{p(\mathbf{x})}{s(\mathbf{x})}$                         \\ \hline
	\end{tabular}
	}
	\vspace{-10pt}
\end{table}

\subsubsection{Importance-Reweighted Prediction Error}\label{sec:reweightedimportancesampling}
We are interested in assessing performance via error metric corresponding to an operator-defined objective, which is some measure of expected prediction error (i) integrated over the feature space $\mathbb{X}$, (ii) with some weight function that expresses how much the operator cares about different points in that space. 
 Consider the modified prediction problem $P=(I,W)$ (where, for now, we leave $Q=I$). The expected prediction error over the target data distribution of interest $p(\mathbf{x})$  can for this problem be written as:
\begin{align}
\varepsilon_p = \varepsilon \left( \mathbf{x},W, \widehat{y}, y \right)  &= \int_\mathbb{X} W(\mathbf{x}) \mathbb{E} \left[ L \left( \widehat{f}(\mathbf{x})- f(\mathbf{x}) \right) \right]   d\mathbf{x} \label{eq:generalerrormetric}
\end{align}
%
where, $W(\mathbf{x}) \rightarrow \mathbb{R} ^{+} $ is the weighting function for importance sampling,  $L$ is the loss function,  
$f(\mathbf{x}) = y\rightarrow \mathbb{R} ^{y}$ 
 and $\int _\mathbb{X}  W\left( \mathbf{x}\right)  d\mathbf{x} < \infty$.
 If we knew $\mathbb{E} \left[ L( (\widehat{f}(\mathbf{x}) - f(\mathbf{x})) ) \right] $, we could directly evaluate this integral, however we do not. 
We can sample from $\mathbf{x}$ and compare our predictions to true values under \eg cross-validation (CV). However, the mean CV error itself will not in general  give us $\varepsilon_p$,  because CV is based on the sampling distribution of the data $s(\mathbf{x})$, which may look nothing like $W(\mathbf{x}), \mathbf{x} \sim p(\mathbf{x})$ (which we can interpret as target distribution, though it may not be normalized).
In order to deal with the mismatch of the sampling and the target distributions, we use importance sampling techniques. 
\begin{align}
\widehat{\varepsilon_p}  =   \frac{1}{N}\sum^{N}_{i=1}\dfrac {W\left( \mathbf{x}_{i}\right) }{s\left( \mathbf{x}_{i}\right) }\left( \widehat {f}\left( \mathbf{x}_{i}\right) -f\left( \mathbf{x}_{i}\right) \right) ^{2} , \mathbf{x}_i \sim s(\mathbf{x}_i) \label{eq:rsserrorimportancesampling}
\end{align} 
where $N$ is the number of sampled data points, $s\left( \mathbf{x}_{i}\right)$ is the sampling distribution, $p\left( \mathbf{x}_{i}\right)$ is the target distribution and the adjustment factor $W(\mathbf{x}_i)/ s(\mathbf{x}_i)$ is the importance ratio.
Thus, we are able to estimate an error weighted by  $W(\mathbf{x})$, $\mathbf{x} \sim p(\mathbf{x})$, with data generated from the distribution $s(\mathbf{x})$. This procedure allows us to transform any method for solving $P_B$ into a method for solving $P=(I,W)$.  
The base problem weighting function is inappropriate for many practical tasks. 
Some intuitive examples of alternative choices of $W$ are summarized in Table \ref{tab:importancesampling_methodology} and described next.

{\indent  \bf (1)  {\em \erroruniform~uniform over $\mathbb{X}$.}} This is equivalent to the expected loss evaluated at a random location in $\mathbb{X}$, reflecting that the operator is equally concerned with performance over all portions of the target area. 
 To obtain this objective function, we need $W(\mathbf{x})$ proportional to a constant, \ie the uniform distribution $u(\mathbf{x}_i)$. This leads to an importance ratio  $w_u \sim \frac{1}{s(\mathbf{x})}$: we weigh each data point inversely by how often its region of the space is sampled, \ie the inverse of the weights implicitly used by naive estimation.

{\indent  \bf  (2) {\em \errorpopulation~proportional to population density.}} An intuitive target for operators is loss averaged over the user population, denoted by \population~at point $\mathbf{x}$ of the input space. We then want $W(\mathbf{x}) \sim \text{\population}$, thus  importance ratio $\text{\wP} \sim \frac{\text{\population}}{s(\mathbf{x})}$. This means that observations from parts of the user population that are rarely sampled need to be given more weight and those that are oversampled should be given less weight. It should be noted that if our sample is representative of our user population, then the naive error estimator is already an approximation of the target. However, if some groups of users are under or oversampled then the naive estimator may not perform well. Crowdsourced data collection is inherently biased due to human mobility and usage patterns.

\textbf{Estimating the sampling distribution $s(\mathbf{x})$.} 
 Our observed signal strength data may have come from a known or unknown sampling design $s(\mathbf{x})$, in which case  $s$ must generally be inferred.  In the experimental results Section~\ref{sec:results} we estimate $s(\mathbf{x})$ via adaptive bandwidth kernel density estimation (KDE)~\cite{lichman:14} on the 2D spatial space and the importance ratio is $w_u \propto \frac{u(\mathbf{l})}{s(\mathbf{l})}$. Our experimental results show that the main source of bias is location of devices. 

\textbf{Training Weighted Random Forests.}
  The \RFs~algorithm splits each node utilizing a random set of features. The criterion of each split is to maximize the  Information Gain (for classification), or to minimize the MSE (for regression). For $N$ training samples, 
  weighted \RFs~\cite{chenweightedRFs} adjust MSE for each split according to the samples weight vector $\mathbf{w} = \left( w_1, \cdots, w_N\right)$, (\ie implicitly turning loss function to a $wMSE$) while the default setting would be $w_i = 1$.

\subsection{Shapley Values of Cellular Measurements \label{sec:shapley-theory}}

The {\bf Shapley Value} ~\cite{shapley1953value} is a celebrated framework in cooperative game theory, 
used to assign value to the contributions of individual players. Recently, it has inspired  {\em data Shapley~\cite{dshapIcml:2019}}, which quantifies the contribution of training data points in supervised machine learning. More precisely, data Shapley provides a measure of the value $\phi_i$ of each training data point (\aka datum) $(\text{\features}_i, y_i )$, for a supervised ML setting which consist of: (i) a training data set  $\mathcal{D}_{train} = \{ (\text{\features}_i, y_i) \}^N_1$ , (ii) a learning algorithm  $\mathcal{A}$ that produces a predictor $\widehat{y} = \widehat{f}_y(\mathbf{x})$ or $\widehat{f}_Q(\mathbf{x})$ and (iii) a performance metric $V(\mathcal{D},\mathcal{A})$. The prediction of the ML algorithm, and thus the value of the training data depend on all three. 
More precisely, the goal is to compute the data Shapley value $\phi_i(\mathcal{D}, \mathcal{A}, V)  \equiv \phi_i(V) = \phi_i \in \mathbb{R}~\forall i, (\text{\features}_i,y_i) \in \mathcal{D}_{train}$, which follows the equitable valuation properties (\ie null property, symmetry, summation and linearity); see Appendix A1 and ~\cite{dshapIcml:2019} for details}.

{\bf Intuition and Computation.}  Intuitively, a data point interacts and influences the training procedure, in conjunction with the other training points. Thus, conditions which formulate the interactions among the data points and an holistic data valuation should be considered.  A simple method is {\bf leave-one-out}, which calculates the datum value by leaving it out and calculating the performance score, \ie $\phi_i^{\text{LOO}} = V(D) - V(D - \{i\})$. However, this formulation does not consider all subsets the point may belong to, thus does not satisfy the equitable conditions \cite{dshapIcml:2019}. %
 According to~\cite{dshapIcml:2019}, the data Shapley value
 must have the following form: 
 \begin{equation}
 \phi_i = C \sum_{\mathcal{D} - \{ i \}} \frac{ V(\mathcal{D} \cup \{ i \})- V(\mathcal{D})}{ \binom{n-1}{|\mathcal{D}|} } 
 \label{eq:datashapleyform} 
 \end{equation}
In other words, the data Shapley value is the average of the leave-one-out value (\aka marginal contributions) of all possible training subsets of data in $D_{train}$. Data Shapley, in its closed form in Eq.~(\ref{eq:datashapleyform}), would require an exponential number of calculations. An approximation - truncated Monte Carlo algorithm (TMC-Shapley) is provided by~\cite{dshapIcml:2019}. We adapt and extend the library for our custom error metrics, in order to estimate the data Shapley value $\phi_i$ of each training data point $(\text{\features}_i, y_i)$. More specifically, we augment it with the recall $R_0$ performance metric for classification, \RFs~regression and our reweighted-spatial uniform error $\varepsilon_u$ for the performance metric $V$, as defined in Sec.~\ref{sec:importancesampling}.

{\bf Application to Cellular Measurements and Prediction.}
 In~\cite{dshapIcml:2019}, the Data Shapley value was tacitly defined for classification of  medical data. In this paper, we apply, for the first time, Data Shapley valuation to mobile performance data, and not only for the base problem $P_B$ but also for the general problem $P=(Q,W)$. 
The input to data Shapley in our context is: (i) the available cellular measurement data used for training (ii) the ML prediction algorithm (\RFs) and (ii) the performance metric $V$, which in our case we define as the re-weighted error based on the operators' objectives, presented earlier in the paper. 
The output is a value assigned to each individual measurement training data point, used for the training for the particular prediction task $(Q,W)$.

{\bf Data Shapley vs. Weight Functions:} We have already described earlier how both the choice of a loss function and of the evaluation metric really matter (Sec. \ref{sec:qualityfunctions} and~\ref{sec:reweightedimportancesampling}). Data Shapley and weight functions share common characteristics but they also have differences, and can complement each other creating a powerful framework. Each framework independently can inform us whether the data are scarce and valuable for our objective, hence can inform how to acquire future data to improve the predictor. However, weight functions on their own do not not quantify the contribution of training data points, \eg if a training data point is an outlier. On the other hand, data Shapley inherently requires a performance score to evaluate the test data.  There is no universal data valuation and for different learning tasks (\ie objectives) some data points might be more valuable than other. Our $(Q,W)$ framework 
  thus define the reweighted error, 
  which provides the performance score for the data Shapley value. 

{\bf Using data Shapley and preview of results.}
Having computed the Shapley values of our cellular measurement data, we can then use them to remove measurements with negative or low Shapley values, in order to  both (i) {\em improve prediction} and (ii) achieve {\em data minimization}. The results in Sec. \ref{sec:dshap_results} show that we can remove up to 65-70\% of data, while  improving the recall for coverage from 64\% to 99\%.

\begin{table*}[t!]
	\centering
	\scriptsize
	\caption{Overview of datasets used throughout the paper. \inhouse is collected by us on our university campus. \external and \externalsuburban were provided by a Mobile Analytics Company} \vspace{-5pt}
   \label{tab:datasets_overview}
   \setlength\tabcolsep{1.5pt} 
	\begin{tabular}{|l|l|l|l|c|}
		\hline
		\textbf{Dataset}                       & \textbf{Period}                                                                        & \textbf{Areas}                                                                                       & \textbf{Type of Measurements}                                                                                                                                                                                                                                                              & \textbf{Characteristics}                                                                                                                                                                                                                                                                                   \\ \hline \hline 
		\inhouse                      & \begin{tabular}[c]{@{}l@{}}02/10/17 - \\ 06/18/17\end{tabular}                & \begin{tabular}[c]{@{}l@{}}Univ. Campus \\ Area $\approxeq 3km^2$\end{tabular}            & \begin{tabular}[c]{@{}l@{}}LTE KPIs: RSRP, {[}RSRQ{]}.\\ Context: GPS Location, timestamp, $dev$, $cid$.\\ Features: $\mathbf{x} = \left( l_j^x, l_j^y,d, h,dev, out, ||\vec{l}_{\text{BS}} - \vec{l}_j||_2 \right)$\end{tabular}                                   & \begin{tabular}[c]{@{}l@{}}
		No. Cells $=25$\\ No. Meas $\approxeq 180$K\\ Density ($\frac{N}{m^2}$)\\ Per Cell: $0.01$ - $0.66$ (Table 3)\\ Overall Density: $0.06$\end{tabular}                                                                                   
		\\ \hline\hline
		\multirow{2}{*}{\external \& \externalsuburban} & \multirow{2}{*}{\begin{tabular}[c]{@{}l@{}}09/01/17-\\ 11/30/17\end{tabular}} & \begin{tabular}[c]{@{}l@{}} NYC Metropolitan \\ Area $\approxeq 300 km^2$\end{tabular} & \multirow{2}{*}{\begin{tabular}[c]{@{}l@{}}LTE KPIs: RSRP, RSRQ, CQI.\\ Context: GPS Location, timestamp, $dev$, $cid$. EARFCN.\\ Features:$\mathbf{x} = \left ( l_j^x, l_j^y,d, h, cid, dev, out, ||\vec{l}_{\text{BS}} - \vec{l}_j||_2, freq_{dl}\right)$\end{tabular}} & \begin{tabular}[c]{@{}l@{}}No. Meas NYC $\approxeq4.2$M\\ No. Cells NYC $\approxeq 88k$\\ Density NYC-all $\approxeq 0.014\frac{N}{m^2}$\end{tabular}  \\ \cline{3-3} \cline{5-5}
		&                                                                               & \begin{tabular}[c]{@{}l@{}} LA metropolitan \\ Area $\approxeq 1600 km^2$\end{tabular} &                                                                                                                                                                                                                                                                     & \begin{tabular}[c]{@{}l@{}}No. Meas LA $\approxeq 6.7$M\\ No. Cells LA $\approxeq 111$K\\ Density LA-all $\approxeq 0.0042 \frac{N}{m^2}$\end{tabular}                                                                                                                                                                                                                 \\ \hline
	\end{tabular}
\end{table*}

\section{Performance Evaluation}\label{sec:results}

\subsection{Data Sets}\label{sec:datasets}
 \begin{figure*}[t!]
	\subfigure[\inhouse example cell x204: high density ($0.66$), low dispersion (325). \label{fig:campus_dense_x204}] {\includegraphics[height=1in]{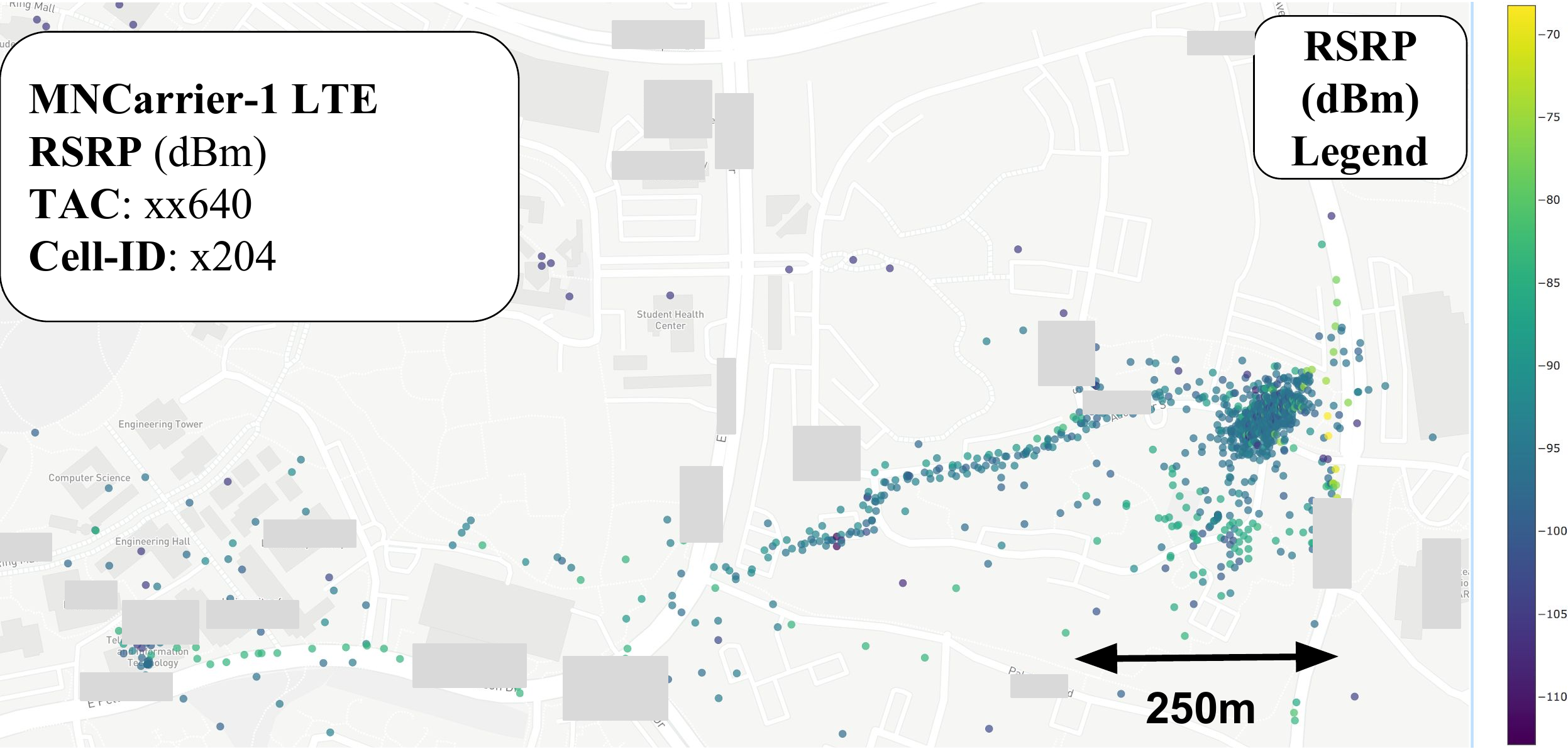} }
	\subfigure[\inhouse: example cell x355: small density ($0.12$) more dispersed data ($573$). \label{fig:campus_sparse_x355}] {\includegraphics[height=1in]{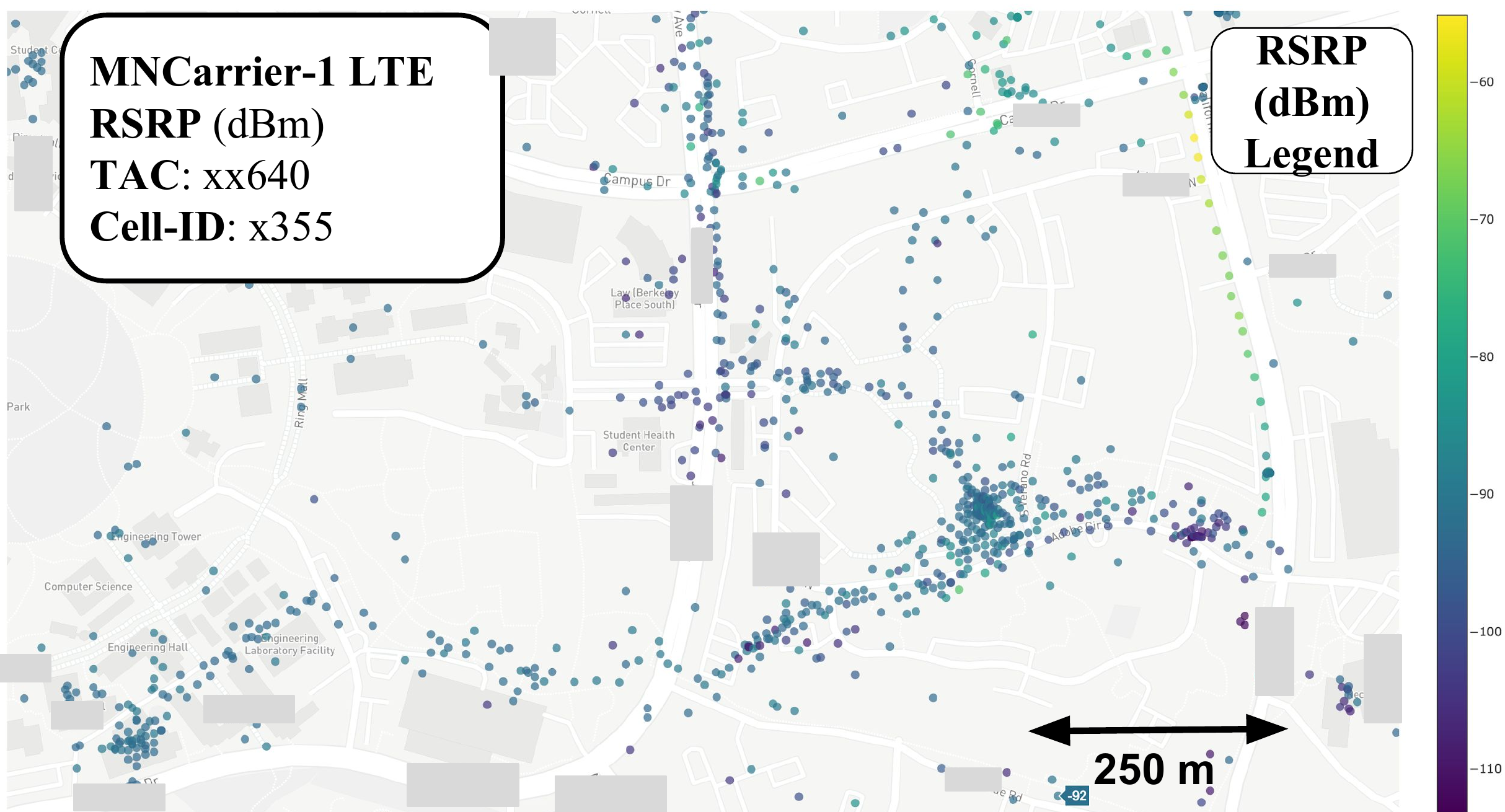}}
	\subfigure[\external: Manhattan \lteta  \label{fig:external_nyc_midtown_data}]{\includegraphics[height=1in]{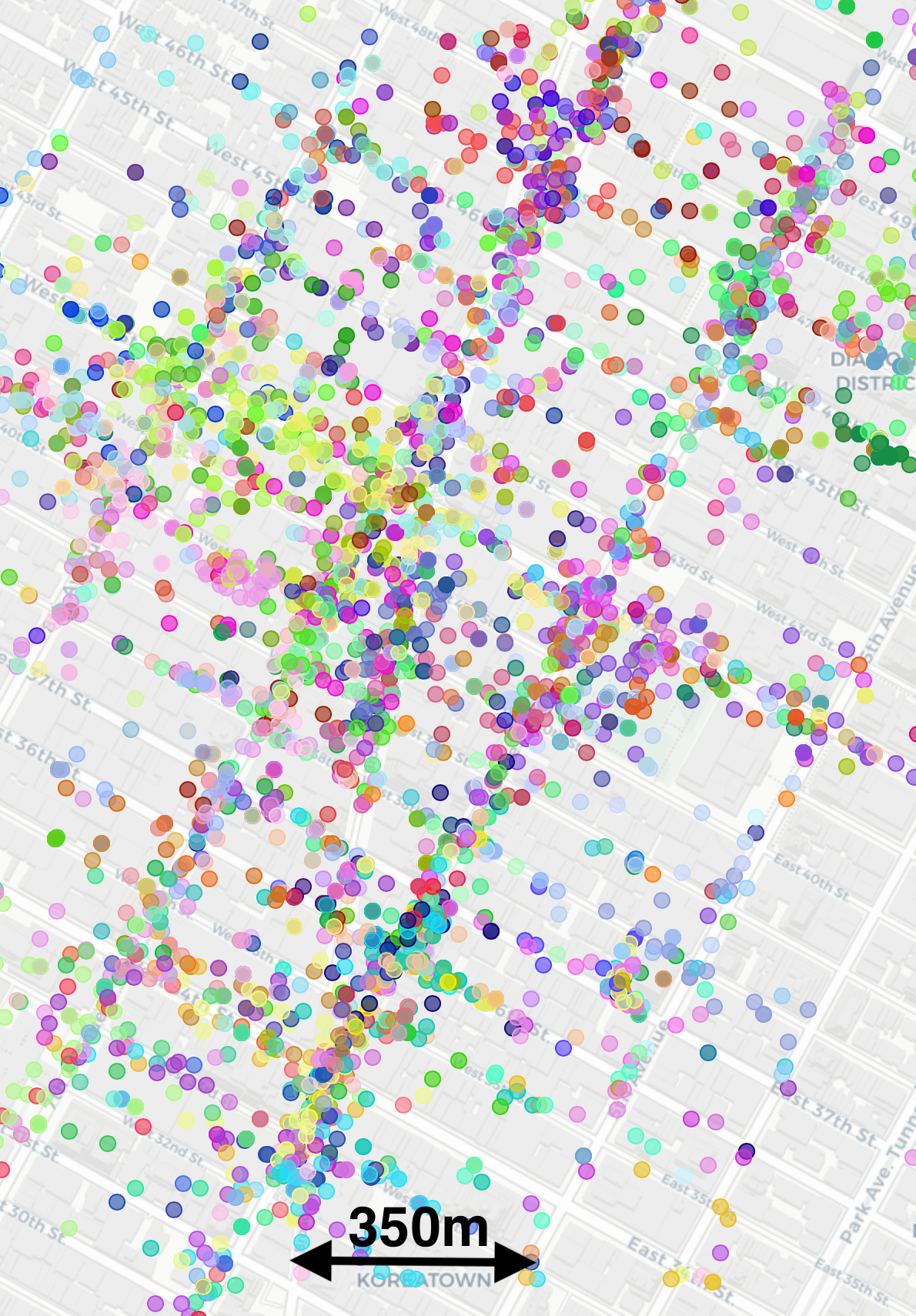}}
	\subfigure[\external: zooming in Manhattan Midtown (Time Square) for some of the available cells.\label{fig:external_nyc_midtown_data_zoomed}]{\includegraphics[height=1in]{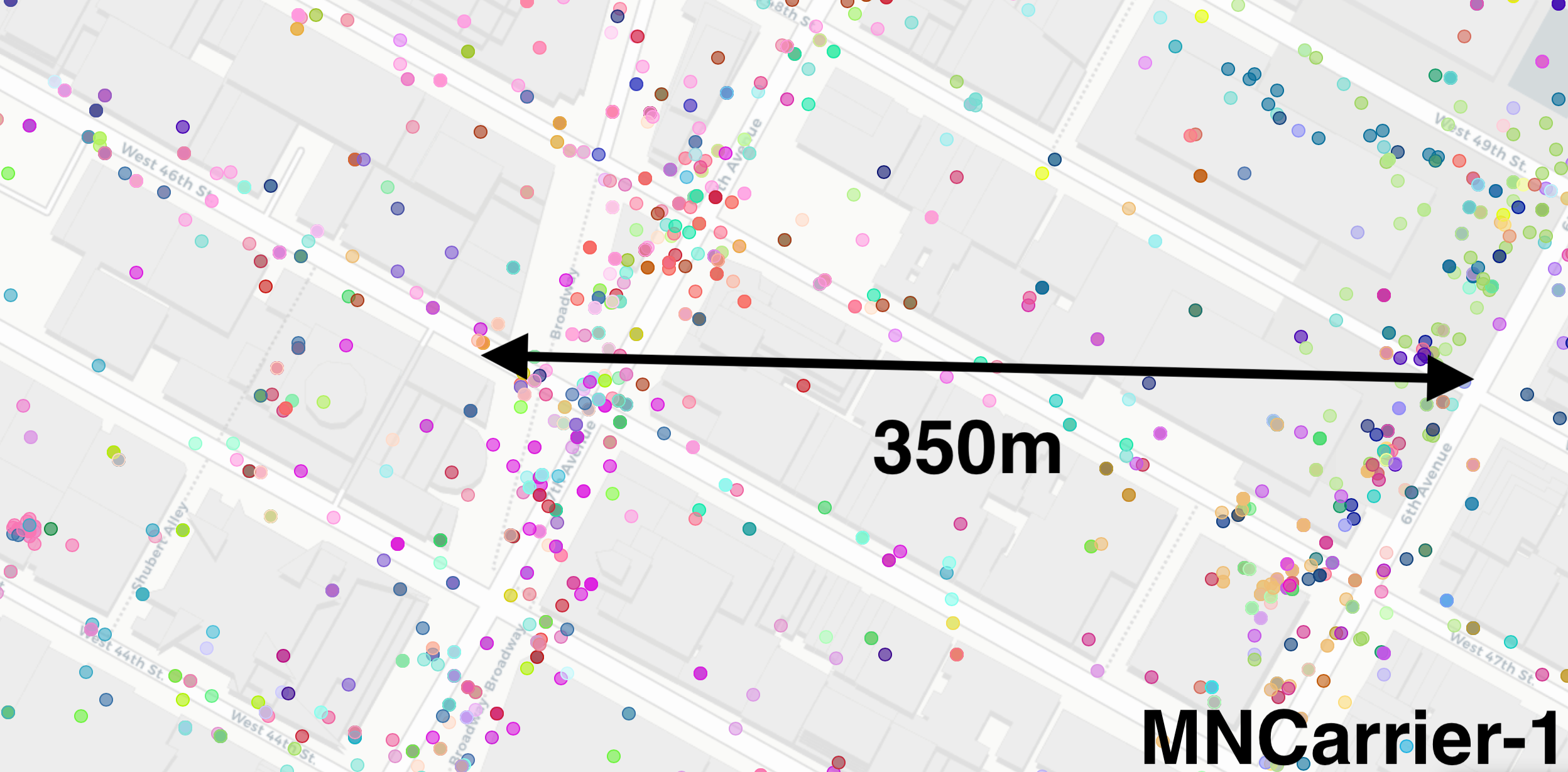}}
	\vspace{-10pt}
	\caption{LTE RSRP Map Examples from our datasets. {\bf (a)-(b):} \inhousedataset. Color indicates RSRP value. {\bf (c)-(d):} \externaldataset. Data for a group of LTE cells in the Manhattan Midtown  area. Different colors indicate different cell IDs.}
	\vspace{-10pt}
	\label{fig:map_examples} 
\end{figure*}

We evaluate our framework using two types of real-world LTE datasets obtained in prior work \cite{alimpertis:19}, the characteristics of which are summarized on Table~\ref{tab:datasets_overview}. 
Both datasets include cellular measurements, and we use the same subset of information from all datasets, \eg RSRP, RSRQ or CQI values and the corresponding features defined in Section \ref{sec:objectivesprediction}. User identifiers were neither collected nor stored for this study.

\textbf{Data Format.} 
We use the same subset of information from all datasets, \eg RSRP, RSRQ or CQI values and the corresponding features defined in Section \ref{sec:objectivesprediction}.
An example, in GeoJSON, with some of the KPIs fields (obfuscated) is shown below:

\begin{mdframed}
\scriptsize
\begin{lstlisting}[language=java,firstnumber=1, caption={\footnotesize{GeoJSON example with LTE KPIs and location, in MongoDB (obfuscated for presentation)}.},captionpos=b]
{"type": "Feature", "properties": {
"timestamp": "2017-09-11T17:54:35EDT",
"lteMeasurement": {"rsrp": -89, "rsrq": -20,
		 "cqi": 9, "pci": 169, "earfcn": 9820},
"cell": { "ci": xxxxx710, "mnc": 410, "mcc": 310,
	"tac": xx22,  "networkType": 4}, 
"device" : {"manufacturer":"samsung",
		"model":"SM-G935P", "os":"android70"},
"locationMetaData": {"city": "New York",
		 "accuracy": "x","velocity":"x"}},
"geometry": {"type": "Point","coords": [-73.9x,40.7x]}}\end{lstlisting}
\end{mdframed}

\textbf{Campus Dataset.} We collected the first dataset on our university campus, via a user-space Android app we developed ourselves and used to collect data from volunteers ~\cite{alimpertis:17}.
This \inhousedataset  is relatively small: $180,000$ data points, collected by seven Android devices that belong to graduate students, using 2 cellular providers, and moving between student housing, offices and other locations on campus (approx. $3km^2$). However, this is a dense dataset, with multiple measurements over time on nearby locations. Due to space limitations, we refer to ~\cite{alimpertis:17, alimpertis:19} for details. 

\textbf{NYC and LA datasets.} These were collected by a major mobile crowdsourcing and data analytics company and shared with us (after appropriate anonymization).
They contain $10.9$M  measurements, covering  approx. $300km^2$ and $1600km^2$ in the metropolitan areas of NYC and LA, respectively,  for a period of 3 months with tens of thousands of unique cells (\ie unique \cid).  An example of the  \external neighborhood in East \external is depicted in Fig.~\ref{fig:rsrpmap_eastnyc_x540}. Examples of representative LTE TAs from \externalcitiesdataset used in our evaluation are summarized later in Table~\ref{tab:external_rsrp_reweighted_uniform}. 
These are large datasets  in terms of any metric, such as number of measurements, geographical scale, number of cells \etc  
As such, they provide novel insight into the problem at a scale that is relevant to operators and mobile analytics companies.

\subsection{Evaluation Setup}\label{sec:results_setup}
{\bf \RFs~predictor.} We train Random Forest (\RFs) to predict the KPI $\widehat{y}$; then we compute \Qobj{$\widehat{y}$}) or $\widehat{Q}(y)$ directly. 
We use state-of-the-art \RFs~\cite{alimpertis:19} as the underlying predictor, but it could be replaced by other ML models.

 The most important hyper-parameters for \RFs~are the number of decision trees  (\ie $n_{trees}$) and the maximum depth of each tree (\ie $\text{max}_{depth}$). 
  We used a grid  search over the parameter values of the \RFs~ estimator~\cite{pedregosaScikit:2011}  in a small hold-out part of the data to select the best values. For the \inhousedataset,  we select $n_{trees}=20$ and $\text{max}_{depth}=20$  via 5-Fold CV; larger $\text{max}_{depth}$  values could result in overfitting of \RFs. For the \externalcitiesdataset, we select $n_{trees}=1000$ and $\text{max}_{depth}=30$; more and deeper trees are required for larger datasets.

 An important design choice is the granularity we choose to build our \RFs~models. As we  demonstrated in~\cite{alimpertis:19}, using a model per cell (\ie train a separate \RFs~ model per cell \cid~ with $\mathbf{x_j^{-cID}} = \{ x: x \in \mathbf{x_j^{full}}  , \sim  x \notin \{cID \}  \} $) is beneficial for large number of measurements per \cid. In sparser data, such as \externalcitiesdataset, it is better to train a model per \lteta~ using $\mathbf{x_j^{full}}$. In this paper, we utilize models per \cid~for the \inhousedataset and per \lteta~models for \externalcitiesdataset  as~\cite{alimpertis:19}.

 To improve the reweighted prediction error $\varepsilon_{p}$ according to \mno'~ objectives (Section~\ref{sec:importancesampling}),  we train weighted random forests \RFsWeights, 
 with $w_i=\{ w_{u_i}, w_{d_i}\}$ proportionally to the target distribution (see Table~\ref{tab:importancesampling_methodology}). In essence, the ML training weights are set equal to the importance ratio of each sample~\cite{deepnetsimportancesampling}. We compare \RFsWeights~with the default \RFs, where all samples are weighted equally. 

{\bf Data Shapley Setup.} For $\widehat{\phi_i}$ estimation with the TMC-Shap algorithm, work in~\cite{dshapIcml:2019} suggests a convergence (stopping) criterion of  $\frac{1}{n} \sum_{i=1}^{n} \frac{| \phi_i^t - \phi_i^{t-100} | }{|\phi_i^t|} < 0.05$ with an observation that the algorithm usually convergences with up to $3N_{train}$ iterations. However, our datasets are significantly larger;~\cite{dshapIcml:2019} use approx. up to $3000$ points and on the contrary the cell x$901$ demonstrated later contains approx. $15000$ measurements. Thus, we relax the convergence criterion to save execution time and we set a 30\% convergence if we exceed 2$N_{train}$ iterations.

{\bf Splitting  Training vs. Testing.} We select randomly $70\%$ of the data as the training set $\mathcal{D}_{train} =  \{\mathbf{X}_{train}, \mathbf{y}_{train}\}$ and $30\%$ of the data as the $~$testing $~$set $\mathcal{D}_{test} = \{\mathbf{X}_{test}, \mathbf{y}_{test}\}$  for the problem of predicting missing signal map values (\ie KPIs $y = \{ \text{\rsrp, \rsrq, \cqi} \}$ or QoS \coverage, \cdp{y}). The reported results are averaged over $S=10$ random splits. 

Our choices differ for data Shapley where  we split the data as following: 60\% of the data for $\mathcal{D}_{train} = \{\mathbf{X}_{train}, \mathbf{y}_{train}\}$, 20\% for $\mathcal{D}_{test} =  \{\mathbf{X}_{test}, \mathbf{y}_{test}\}$ and 20\% for the held-out set $\mathcal{D}_\text{held-out}$. Data Shapley values $\phi_i$ are being calculated per training point $(\text{\features}_i, y_i)$ w.r.t. the performance score $V$ of the prediction on $\mathcal{D}_{test}$. We use the $\mathcal{D}_\text{held-out}$ dataset to report the final data minimization results, \ie use some completely unseen data  since $\mathcal{D}_{test}$ was used for the data Shapley $\phi_i$ itself.

{\bf Evaluation Metrics - Coverage Classification.} We evaluate the performance of $Q_c(y)$ in terms of binary classification metrics, \ie recall, precision, F1 score and balanced accuracy. 
{\em Recall} is defined as $R = \frac{T_p}{T_p + F_n}$ where $T_p$ is the true positive rate and $F_n$ is the false negative rate, \emph{for the class of interest}. {\em Precision} is defined as  $Pr= \frac{T_p}{T_p + F_p} $. {\em F1 Score} is the weighted average of precision and recall and {\em Balanced Accuracy}  the average of recall for each class.

{\bf Evaluation Metrics for  Regression.}
{\em (I) Root MSE ($RMSE$)}. If $\widehat{y}$ is an  estimator for $y$, then  $RMSE(\widehat{y}) = \sqrt{MSE(\widehat{y})} = \sqrt{E((y - \widehat{y})^2)}$, in dB for RSRP \rsrp and RSRQ \rsrq and unitless for CQI \cqi. 
{\em (II) Reweighted Error $\varepsilon_{p}$ for a target distribution $p(\mathbf{x})$.} According to Eq.~(\ref{eq:rsserrorimportancesampling}), $\varepsilon_{p} = \frac{1}{N} \sum_{i=1}^{N} w_i \left( \widehat{y}_i - y_i\right)^2$, with $w_i=\{ w_{u_i}, w_{d_i}\} \propto \{\frac{1}{s(\mathbf{l}_i)}, \frac{\text{\pop{$\mathbf{l}_i$}}}{s(\mathbf{l}_i)}  \}$, as defined in Table~\ref{tab:importancesampling_methodology}, where $w_u$ corresponding to the importance ratio for error in a random location in $\mathbb{X}$ and $w_P$ the weighting proportional to population density. We use only location density $s(\mathbf{l})$ to calculate the (i) uniform error $\varepsilon_{u}$ or (ii) \wP, over the space $\mathbb{X}$, but our methodology is applicable to an arbitrary space $\mathbb{X}$.

\subsection{Results for Base Problem \problem{$I,k$}}
\label{sec:numerical_results_baseproblem}

Prior work has exclusively focused on RSRP  prediction minimizing MSE. Evaluating the performance of \RFs~ for this base problem is necessary  to show that our \RFs~based predictor \cite{alimpertis:19} 
 is a good choice for the underlying ``workhorse'' ML predictor on top of which, we can develop our $P=(Q,W)$ framework. We compare \RFs~against several baselines: model-driven (\LDPLknn~and \LDPLhom), geospatial interpolation (OK and OKD),  and a multilayer DNN (trained with $l^x,l^y $ features). We show that \RFs~achieves lower MSE than all those alternatives: Table \ref{tab:campus_allcells} shows that comparison of all predictors for the cells of the Campus dataset.  Due to space constraints, and since this is not a core contribution of this paper, we defer additional evaluation details to Appendix A2 and our prior work in~\cite{alimpertis:19}.  We emphasize that our framework builds on top of the base predictor, which can be swapped for other state-of-the art square loss predictors, as those become available. For the rest of the paper, we focus on evaluating our framework for  \problem{Q,W}, which essentially modifies the loss function on the base problem \baseproblem, and the data Shapley valuation ($\phi$).

\begin{table}[t!]
  \centering
  \setlength\tabcolsep{1.5pt}
  \caption{\small \inhousedataset: Comparing predictors per cell ($cID$) for the Base Problem, \problem{$I,k$}, for all cells of the \inhousedataset. 
  One can see that the \RFs~ predictor (last columns) achieves lower MSE than all alternative predictors (\LDPL, OK, OKD, DNN), and is therefore a good choice.}
  \label{tab:campus_allcells}
  \scriptsize
	\begin{tabular}{|l|r|r|l|r|r||r|r|r|r|r|r|r|r|}
		\hline
		& \multicolumn{4}{l|}{Cell Characteristics}           & \multicolumn{8}{l|}{$RMSE$ (dB)}  \\ \hline
		$cID$ & $N$                   & \multicolumn{1}{l|}{$\frac{N}{\text{sq } m^2}$} & $\E[P]$                 & $\sigma^2$            & \multicolumn{1}{l|}{\begin{tabular}[c]{@{}l@{}}{\tiny LDPL}\\ {\tiny hom}\end{tabular}} & \multicolumn{1}{l|}{\begin{tabular}[c]{@{}l@{}}{\tiny LDPL}\\ $\scriptstyle{kNN}$\end{tabular}} & \multicolumn{1}{l|}{\tiny OK} & \multicolumn{1}{l|}{{\tiny OKD}} & \multicolumn{1}{l|}{{\tiny DNN}} & \multicolumn{1}{l|}{\begin{tabular}[c]{@{}l@{}}$RFs$\\ $\scriptstyle{x,y}$\end{tabular}} & \multicolumn{1}{l|}{\begin{tabular}[c]{@{}l@{}}$RFs$\\ $\scriptstyle{x,y,t}$\end{tabular}} & \multicolumn{1}{l|}{\begin{tabular}[c]{@{}l@{}}$\scriptstyle{RFs}$\\ $\scriptscriptstyle{all}$\end{tabular}} \\ \hline
		x312  & 10140 & 0.015                                                                    & -120.6 & 12.0       & 17.5                                                                     & 1.63                                                                      & 1.70                    & 1.37   & 2.05                      & 1.58                                                                         & 0.93                                                                           & \textbf{0.92}                                                                    \\ \hline
		x914  & 3215  & 0.007                                                                    & -94.5  & 96.3       & 13.3                                                                     & 3.47                                                                      & 3.59                    & 2.28 & 6.48                    & 3.43                                                                         & 1.71                                                                           & \textbf{1.67}                                                                         \\ \hline
		x034  & 1564  & 0.010                                                                   & -101.2 & 337.5      & 19.5                                                                     & 7.82                                                                      & 7.44                    & 5.12  & 11.59                    & 7.56                                                                         & 3.82                                                                           & \textbf{3.84}                                                                        \\ \hline
		x901  & 16051 & 0.162                                                                   & -107.9 & 82.3       & 8.9                                                                      & 4.60                                                                      & 4.72                    & 3.04  & 5.69                    & 4.54                                                                         & 1.73                                                                           & \textbf{1.66}                                                                        \\ \hline
		x204  & 55566 & 0.666                                                                    & -96.0  & 23.9       & 6.9                                                                      & 3.84                                                                      & 3.85                    & 2.99 & 4.44                    & 3.83                                                                         & 2.30                                                                           & \textbf{2.27}                                                                         \\ \hline
		x922  & 3996  & 0.107                                                                    & -102.7 & 29.5       & 5.6                                                                      & 3.1                                                                       & 3.16                    & 2.01 & 4.51                    & 3.10                                                                         & 1.92                                                                           & \textbf{1.82}                                                                         \\ \hline
		x902  & 34193 & 0.187                                                                    & -111.5 & 8.1        & 21.0                                                                     & 2.60                                                                      & 2.47                    & 1.64 & 2.8                    & 2.50                                                                         & 1.37                                                                           & \textbf{1.37}                                                                         \\ \hline
		x470  & 7699  & 0.034                                                                    & -107.3 & 16.9       & 24.8                                                                     & 3.64                                                                      & 2.73                    & 1.87 & 3.33                    & 2.78                                                                         & 1.26                                                                           & \textbf{1.26}                                                                         \\ \hline
		x915  & 4733  & 0.042                                                                    & -110.6 & 203.9      & 14.3                                                                     & 7.54                                                                      & 7.39                    & 4.25 & 9.94                    & 7.31                                                                         & 3.29                                                                           & \textbf{3.15}                                                                         \\ \hline
		x808  & 12153 & 0.035                                                                    & -105.1 & 7.7        & 4.40                                                                     & 2.41                                                                      & 2.42                    & 1.60 & 2.84                    & 2.34                                                                         & 1.75                                                                           & \textbf{1.59}                                                                          \\ \hline
		x460  & 4077  & 0.040                                                                    & -88.0  & 32.8       & 11.2                                                                     & 2.35                                                                      & 2.28                    & 1.56 & 3.60                    & 2.31                                                                         & 1.84                                                                           & \textbf{1.84}                                                                         \\ \hline
		x306  & 4076  & 0.011                                                                    & -99.2  & 133.3      & 18.3                                                                     & 4.85                                                                      & 4.30                    & 2.80 & 7.07                    & 3.94                                                                         & 3.1                                                                            & \textbf{3.06}                                                                         \\ \hline
		x355  & 30084 & 0.116                                                                    & -94.3  & 42.6       & 9.3                                                                      & 2.42                                                                      & 2.31                    & 1.85 & 3.28                    & 2.26                                                                         & 1.79                                                                           & \textbf{1.79}                                                                         \\ \hline
	\end{tabular}
  \end{table}

\subsection{Results for QoS functions $P=(Q,k)$}\label{sec:numerical_results_qos}

\begin{figure*}[th!]
	\subfigure[Actual Spots with Bad Coverage. \label{fig:badcoverage_uci}] {\includegraphics[height=1.1in]{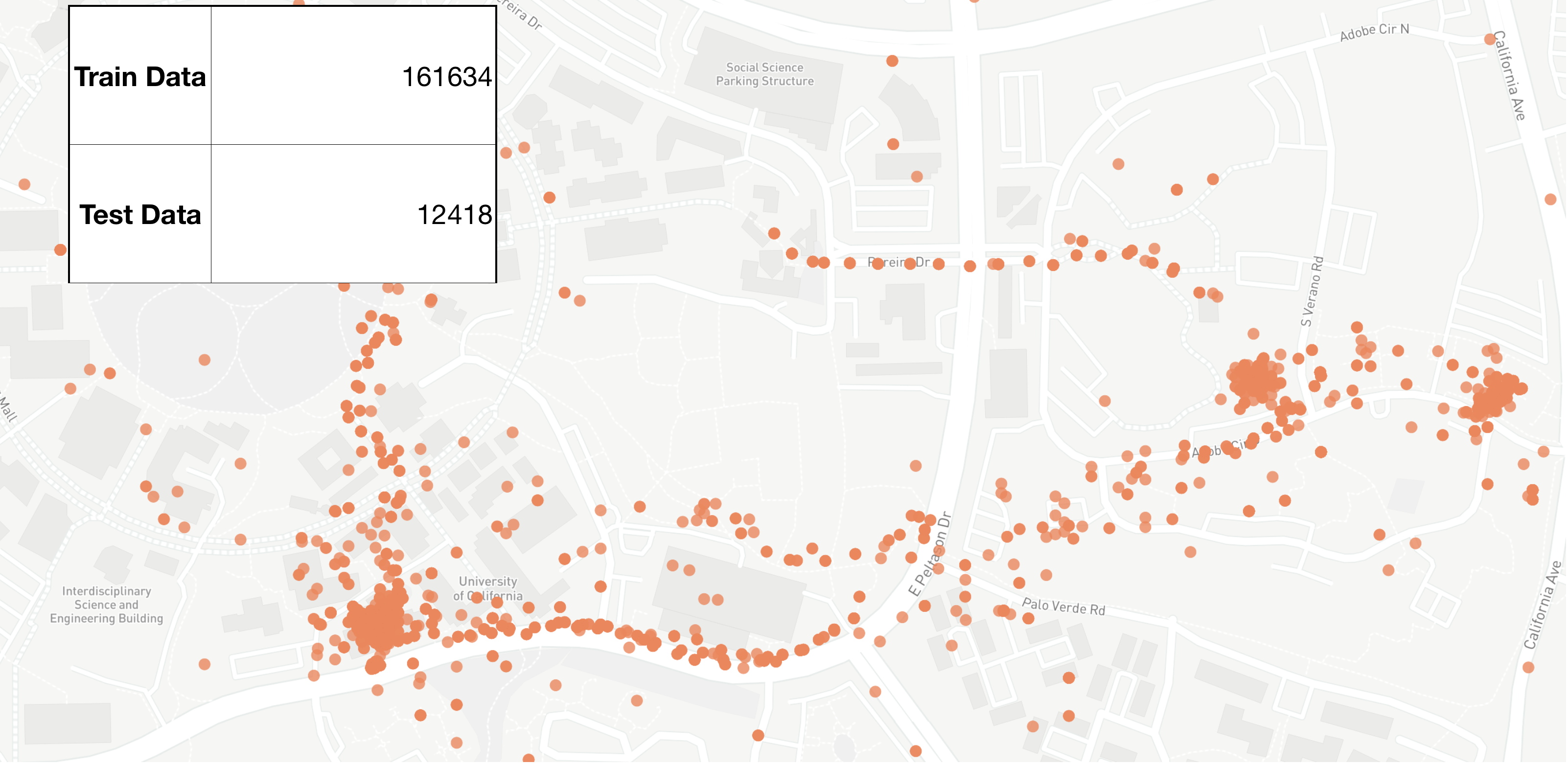} }
		\hspace{-10pt}
	\subfigure[Baseline Prediction \Qobj{$\widehat{y}$}. \label{fig:baseline_uci0}] {\includegraphics[height=1.1in]{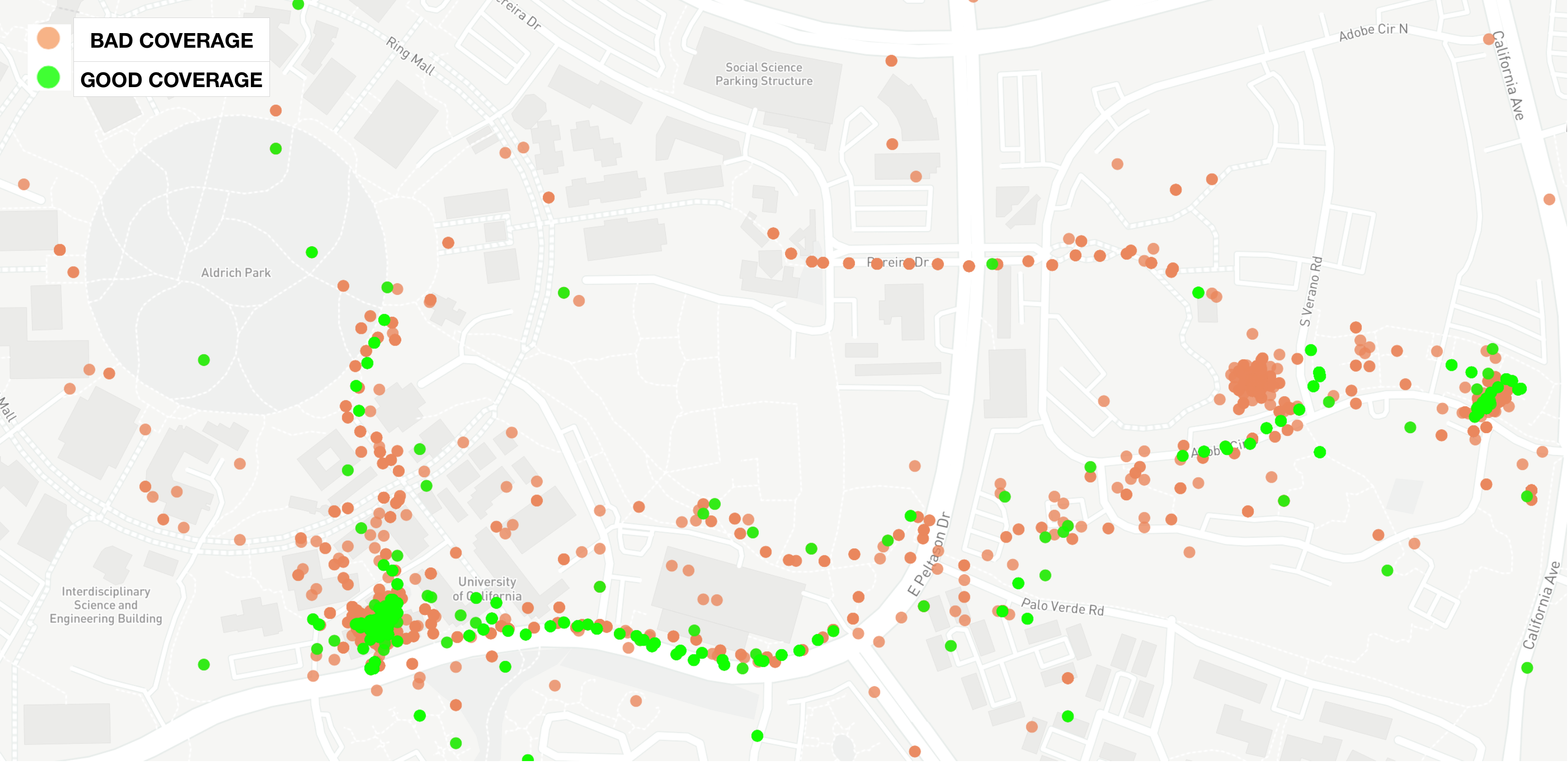}}
	\hspace{-10pt}
	\subfigure[Our Approach $\widehat{Q_c}(y)$  \label{fig:ourapproach_uci}]{\includegraphics[height=1.1in]{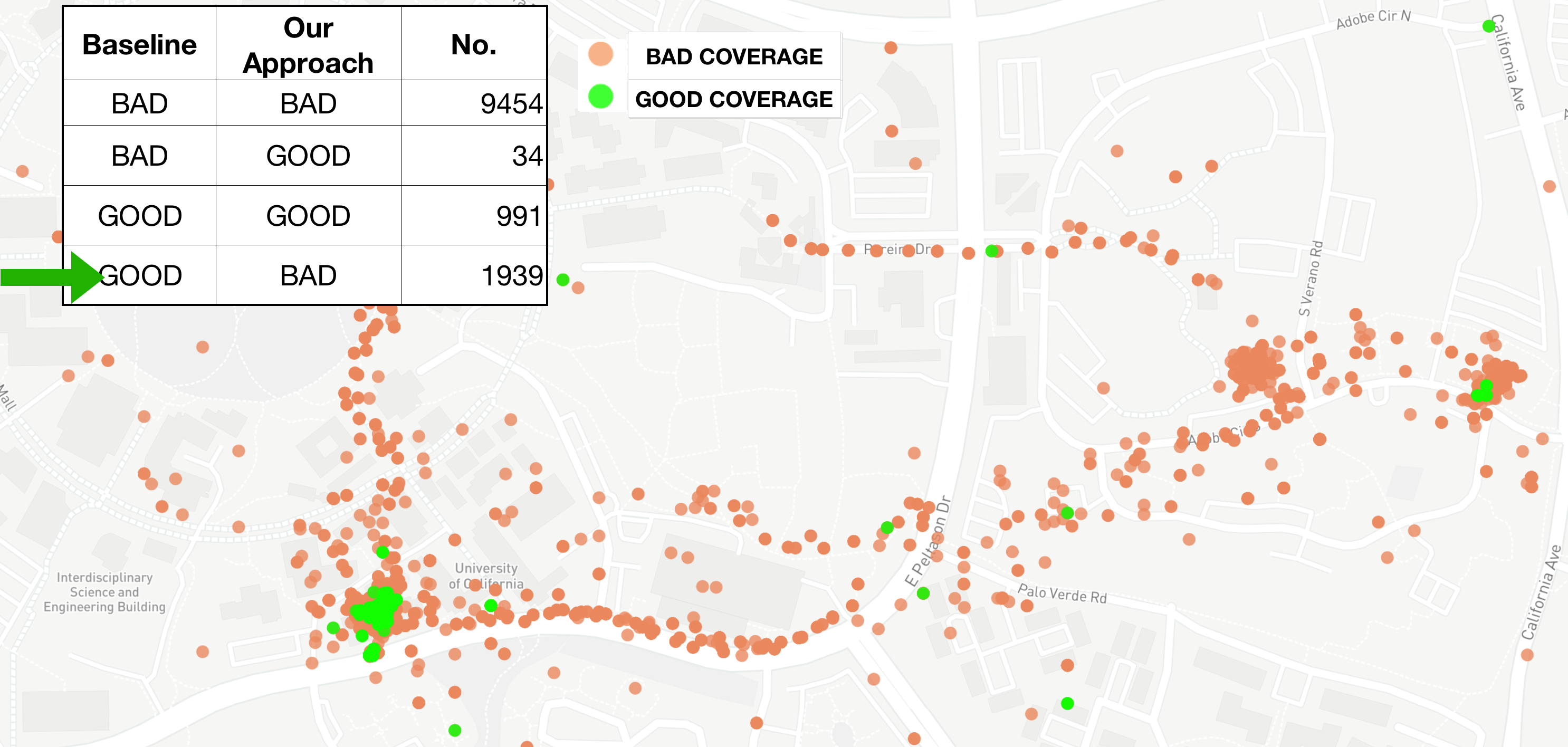}}
	\vspace{-10pt}
	\caption{\footnotesize LTE Coverage Map for our own \inhousedataset. Display only Test Data. (a) Bad Coverage from Test Data (b) Baseline-Proxy- Prediction \Qobj{$\widehat{y}$} (c) Our Model Prediction. It can be seen that (c) has more red points than (b), implying better classification. For this example, we discover 1939 data points which the baseline would not detect (16\% of the total 12418 bad coverage points). Best viewed in color.}
	\vspace{-10pt}
	\label{fig:uci_coverage} 
\end{figure*}

\subsubsection{Coverage Domain, \problem{$Q_c,k$}}
This setup is a typical binary classification problem, where class 0 corresponds to bad coverage  and class 1 corresponds to good coverage. As a baseline, we train the \RFs {} regression models, we predict $\widehat{y}$ and compute the proxy \Qobj{$\widehat{y}$}. We compare that with our proposed approach, which is to train \RFs~classifiers, with the same features, on quality-transformed observations $(\mathbf{x}_i,Q(y_i))$ and predicting $\widehat{Q}(y)$. For coverage indicator, we employ $y=$\rsrp since it is defined on RSRP. \RFs~use the default training ($\forall i, w_i = 1$).

In this setup, bad coverage  (class-0) misclassified as  good coverage areas (class-1)  impact reputation, revenue, and overall performance. 
 Therefore, from the \mno' perspective, it is desirable to maximize the Recall for class-0 $R_0$ because  higher  Recall means fewer  false negatives, \ie our algorithm did not classify a bad coverage ($Q(y) = 0$) as a good coverage area ($\hat{Q}(y) = 1$).

\begin{figure}[t!]
	\centering
	\subfigure[Baseline-Proxy $Q_c(\widehat{y})$\label{fig:Q_c_rsrpPred_confmatrix}] {\centering\includegraphics[scale = 0.28]{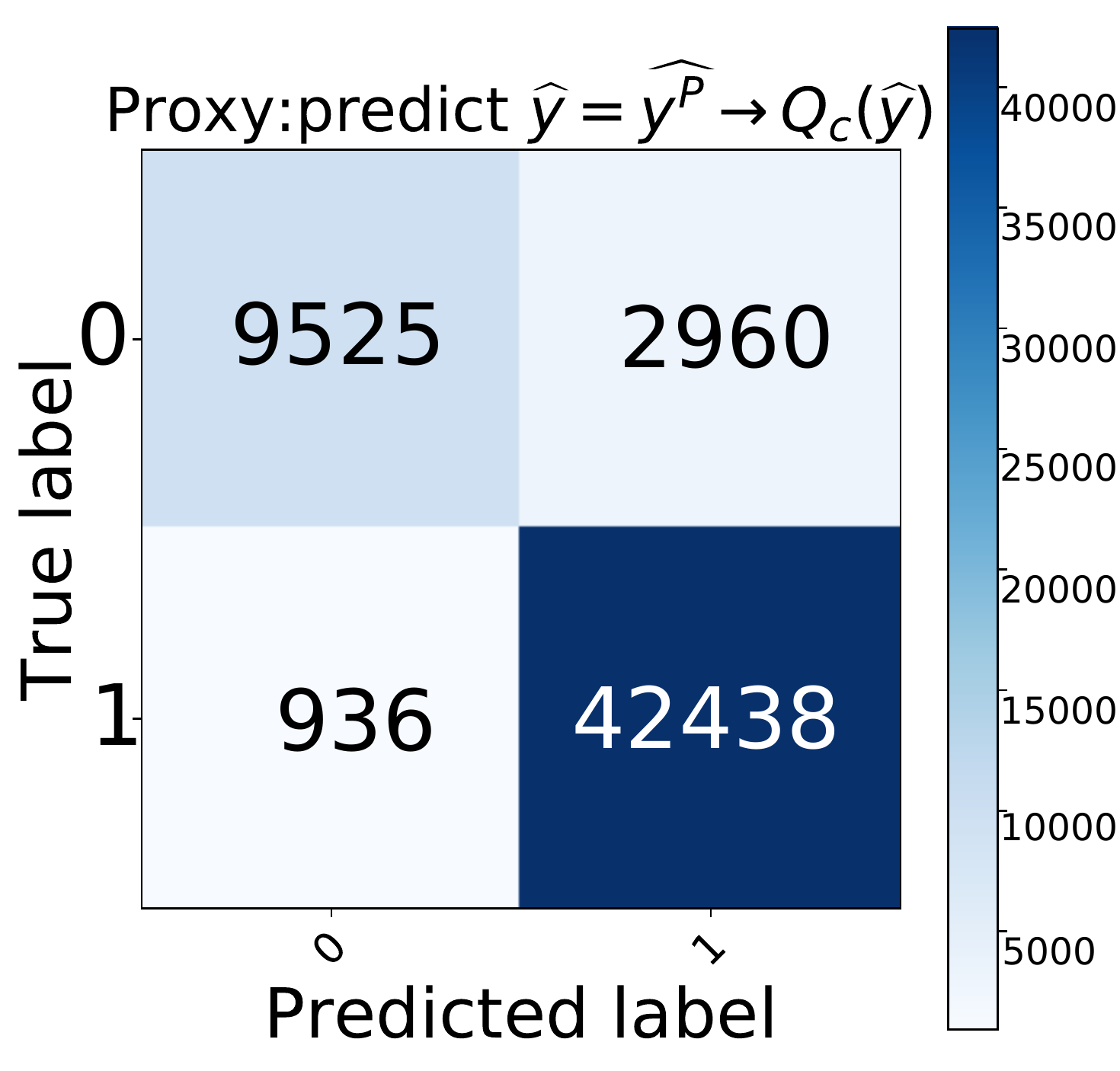}}
	\subfigure[Our method $\widehat{Q}_{cdp}(y)$.\label{fig:Q_c_pred_confmatrix}] {\centering\includegraphics[scale = 0.28]{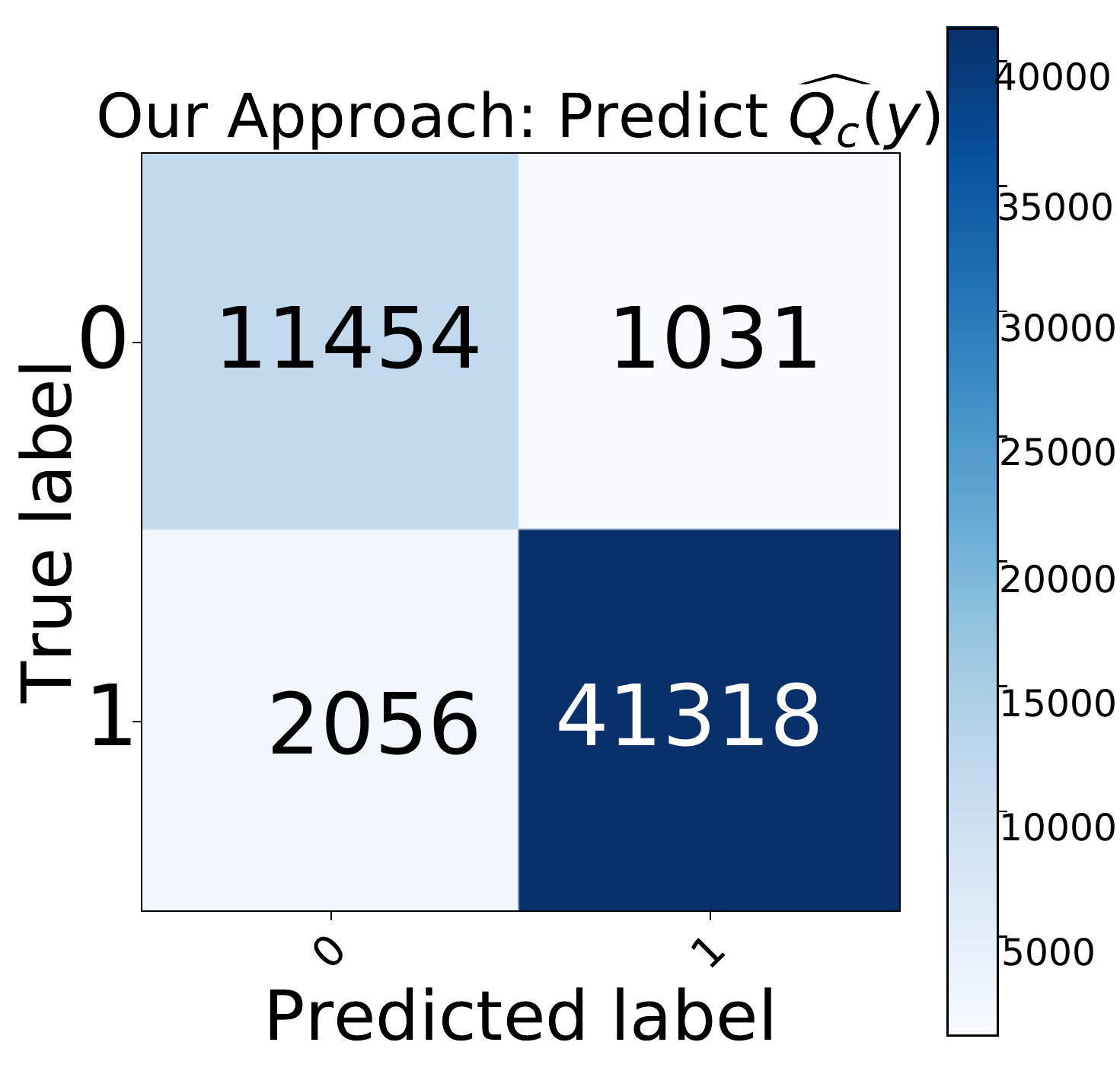}}
	\vspace{-10pt}
	\caption{ \footnotesize For \inhousedataset: Confusion Matrix for coverage, \coverage~(\problem{$Q_c,k$}). The points incorrectly classified as ``good coverage'' by the baseline $Q_c(\widehat{y})$ predictor are shifted to the ``bad coverage'' class under our predictor $\widehat{Q}(y)$.}
	\vspace{-5pt}
	\label{fig:coverage_confmatrix_campus} 
\end{figure}
\begin{table}[]
	\centering
	\setlength\tabcolsep{1.5pt} 
	\footnotesize
	\begin{tabular}{|l|l|l|l|l|l|l|l|l|}
		\hline
		& \multicolumn{2}{l|}{\textbf{Recall}} & \multicolumn{2}{l|}{Precision} & \multicolumn{2}{l|}{F-1} & \multirow{2}{*}{Accuracy} & \multirow{2}{*}{\begin{tabular}[c]{@{}l@{}}\textbf{Balanced} \\ \textbf{Accuracy}\end{tabular}} \\ \cline{1-7}
		Class Label & \textbf{0}            & 1            & 0              & 1             & 0           & 1          &                           &                                                                               \\ \hline
		$Q_c(\widehat{y})$             & \textbf{0.762}        & 0.978        & 0.910          & 0.934         & 0.830       & 0.956      & 0.930                     & \textbf{0.870}                                                                         \\ \hline
		$\widehat{Q_c}(y)$             & \textbf{0.917}        & 0.952        & 0.847          & 0.975         & 0.881       & 0.963      & 0.944                     & \textbf{0.935}                                                                         \\ \hline
	\end{tabular}
\caption{\footnotesize \inhouse Coverage $Q_c(y)$ results: (i) Recall for Class-0 (No-Coverage) 76\% $\rightarrow$ $92\%$, (ii) Accuracy and (iii) Balanced Accuracy Improve. The improved Recall  ($R_0$) is of immense importance for Cellular Providers; higher $R_0$ means less false negatives for \coverage~(\ie miss-classifications of bad coverage to good coverage).  \label{tab:campus_coverage_results}}
\vspace{-5pt}
\end{table}

{(a) \inhousedataset:}  Fig.~\ref{fig:uci_coverage} illustrates the improvement in the \inhouse dataset from utilizing our predictor $\widehat{Q}(y)$ instead of the naive proxy {$Q_c(\widehat{y})$} for bad coverage spots. For this example, we discover 1939 bad coverage sites that the baseline did not detect (16\% of the total 12418
 bad coverage points). Moreover, Fig.~\ref{fig:ourapproach_uci} shows how the bad coverage spots which were mis-classified as good coverage spots have been reduced by our predictor $\widehat{Q_c}(y)$, especially in areas of densely sampled data and commute traces (note the road and path trajectories).  The confusion matrix for these results is shown in Fig.~\ref{fig:coverage_confmatrix_campus}, where we can see again the shift of points incorrectly classified as ``good coverage'' by the baseline $Q_c(\widehat{y})$ predictor to the ``bad coverage'' class under  our predictor.
The overall classification results, in terms of the binary classification metrics, are shown in Table~\ref{tab:campus_coverage_results}. We see an improvement of 16\% for Recall $R_0$, per Fig.~\ref{fig:uci_coverage}, as well an improvement in balanced accuracy from 87\% to 94\%. These improvements do not come at the expense of F1 and Accuracy, which improve by approx. 1\%.

{(b) \externalcitiesdataset:} Table~\ref{tab:external_coverage_results} lists the classification results for some  examples from \externalcitiesdataset. We see  a similar increase up to 12\% in terms of $R_0$ for our predictor $\widehat{Q_c}(y)$ compared to the baseline proxy. 
\begin{table}[t!]
	\centering
	\setlength\tabcolsep{1.5pt} 
	\footnotesize
\begin{tabular}{|l|l|l|l|l|l|l|l|l|}
\hline
                                        & \multicolumn{2}{l|}{\textbf{Recall}} & \multicolumn{2}{l|}{\textbf{Precision}} & \multicolumn{2}{l|}{\textbf{F-1}} & \multirow{2}{*}{\textbf{Accuracy}} & \multirow{2}{*}{\textbf{\begin{tabular}[c]{@{}l@{}}Balanced \\ Accuracy\end{tabular}}} \\ \cline{1-7}
Class Label & 0                 & 1                & 0                  & 1                  & 0               & 1               &                                    &                                                                                        \\ \hline
\multicolumn{9}{|l|}{MNC-1, LTE-TA: x552, Eastern Brooklyn}                                                                                                                                                                                                                           \\ \hline
\textbf{$Q_c(\widehat{y})$}             & 0.55              & 0.98             & 0.80               & 0.93               & 0.65            & 0.96            & 0.93                               & 0.77                                                                                   \\ \hline
\textbf{$\widehat{Q_c}(y)$}             & \textbf{0.67}              & 0.95             & 0.70               & 0.95               & 0.68            & 0.96            & 0.93                               & \textbf{0.81}                                                                                   \\ \hline
\multicolumn{9}{|l|}{MNC-1, LTE-TA: x641, LA, Covina - Hacienda Heights}                                                                                                                                                                                                                              \\ \hline
\textbf{$Q_c(\widehat{y})$}             & 0.58              & 0.90             & 0.73               & 0.82               & 0.65            & 0.86            & 0.80                               & 0.74                                                                                   \\ \hline
\textbf{$\widehat{Q_c}(y)$}             & \textbf{0.70}              & 0.86             & 0.70               & 0.86               & 0.70            & 0.86            & 0.81                               & \textbf{0.78}                                                                                   \\ \hline
\end{tabular}

\caption{ \footnotesize \externalcitiesdataset Coverage \coverage~results. Recall $R_0$ improves up to 12\%. Operators would ideally minimize the false negatives of class-0. Similar results observer in other \lteta s . \label{tab:external_coverage_results}} 
\vspace{-10pt}
\end{table}

\subsubsection{Call Drop Probability  Domain, \problem{$Q_{cdp},k$}} \label{sec:cdpresults}
CDP \cdp{y} estimation is a continuous value prediction problem   on the [0,1] interval. As with the coverage domain, we train \RFs~models in order to predict $\widehat{y}$ and use the proxy $Q_{cdp}(\widehat{y})$ as a baseline. We compare that with our approach, which is to train \RFs, using the same features, on quality-transformed observations $(\mathbf{x}_i,Q(y_i))$ and predict $\widehat{Q_{cdp}}(y)$. We report the relative reduction in $RMSE$.

\begin{figure}[t!]
  \begin{center}
  	\subfigure[ $RMSE$ Relative Reduction.\label{fig:CDP_rmse_vs_rsrp_campus}] {\includegraphics[width=0.325\textwidth]{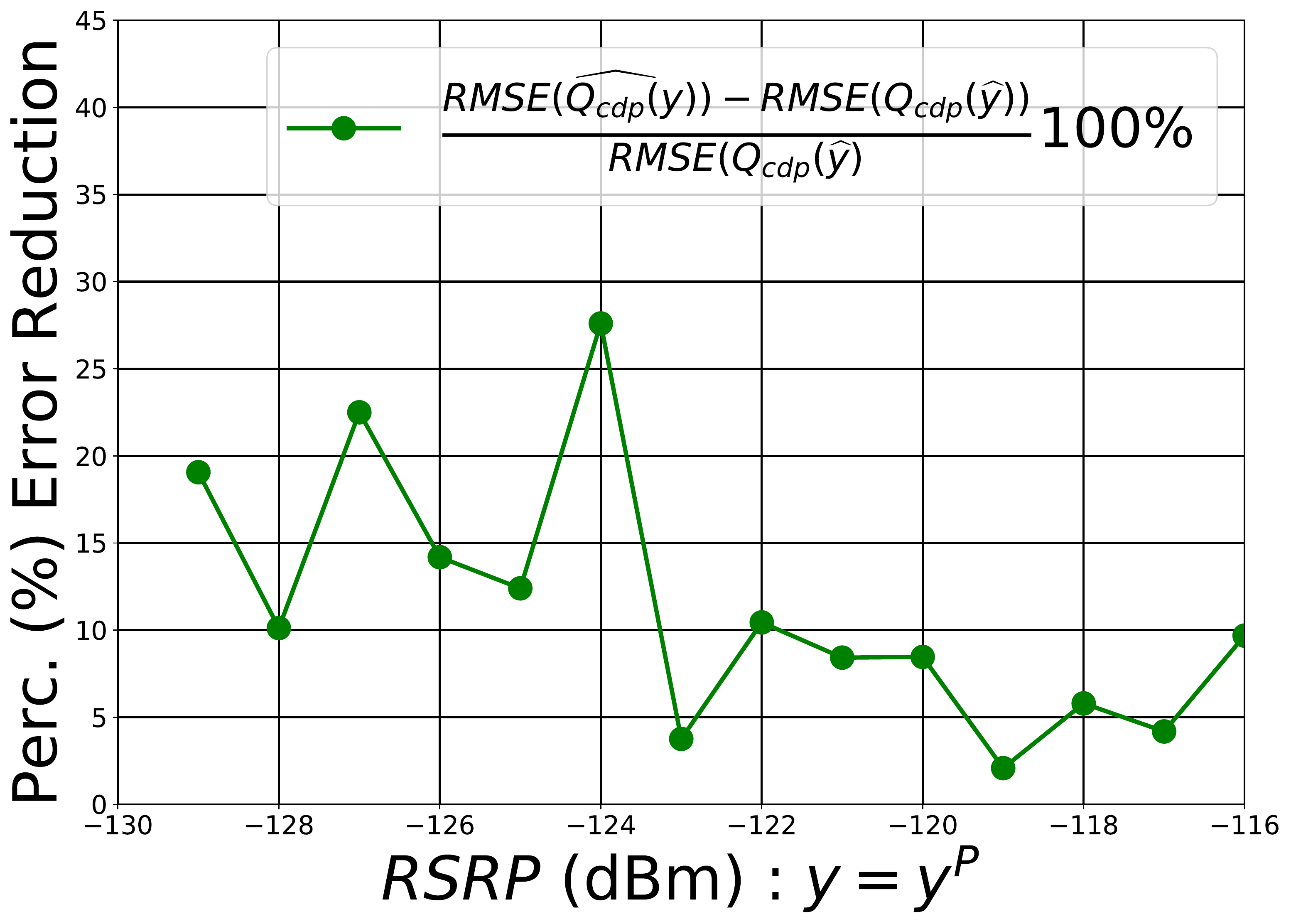}}
	\subfigure[$Q_{cdp}(y) \leftrightarrow$ RSRP $y$\label{fig:rsrp_RFs_vs_Qinv_campus}] {\includegraphics[width=0.325\textwidth]{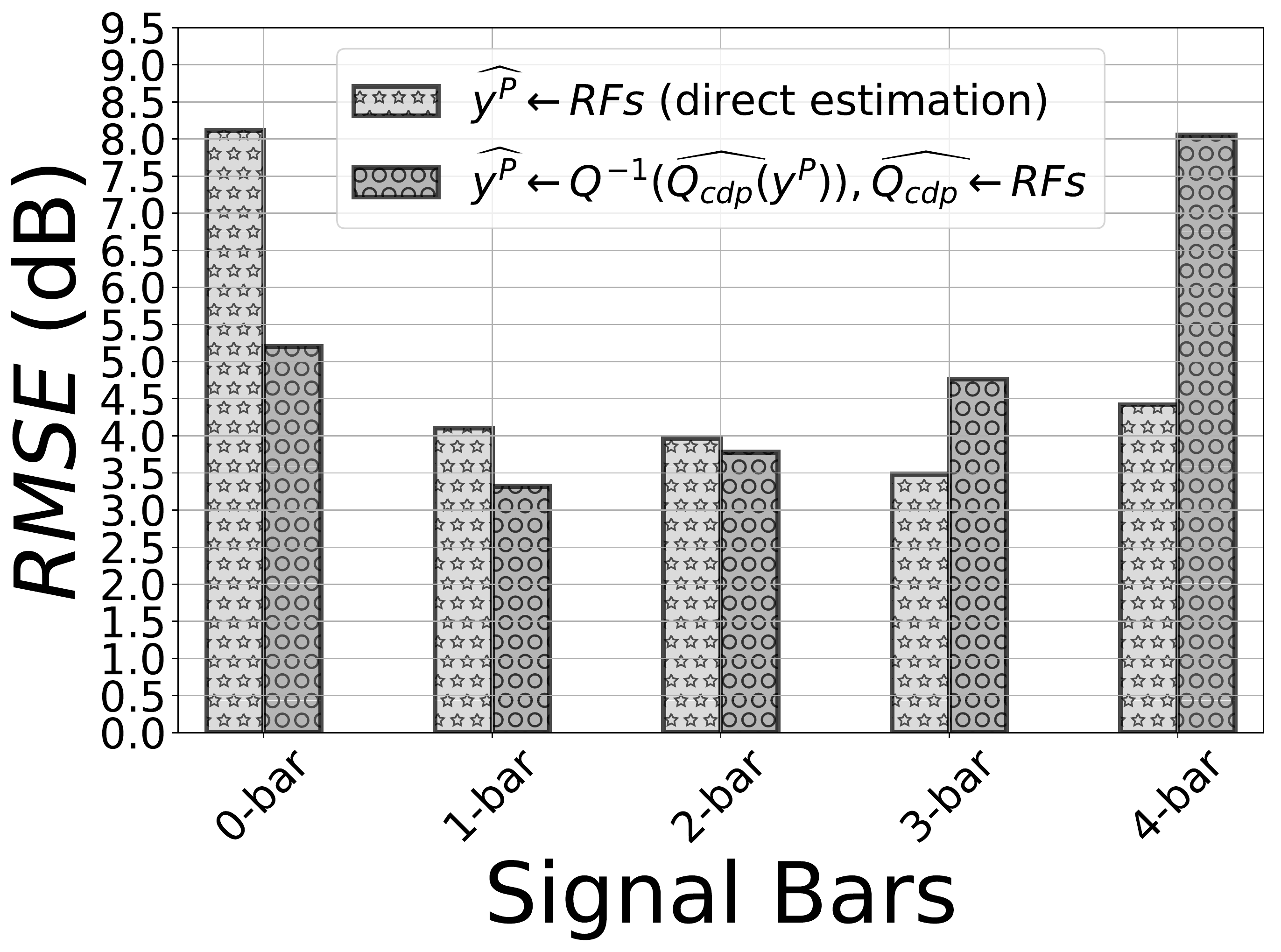}}
	\vspace{-15pt}
	\caption{\footnotesize Improving the RSRP prediction itself via $P=(Q,k) \rightarrow P_B=(I,k)$ transform, for the \inhousedataset. (a) \cdp{$y$} $RMSE$ \vs RSRP. Our methodology reduces the error up to 27\%  for lower RSRP values (0-1 bars). (b) $Q^{-1}(\widehat{Q}_{cdp})$ $RMSE$ \vs Predicting directly RSRP values: the improvement has shifted towards the lower RSRP (fewer bars) accordingly to the new QoS function $Q_{cdp}(y)$ we trained for. 
	}
	\label{fig:cdp_rmse_vs_rsrp_campus} 
		\vspace{-10pt}
  \end{center}
\end{figure}

{(a) \inhousedataset:} In Fig. \ref{fig:cdp_rmse_vs_rsrp_campus}, we report results for estimating CDP, when using the proxy baseline vs. predicting CDP directly. Fig.~\ref{fig:CDP_rmse_vs_rsrp_campus} shows the relative reduction in $RMSE$ error of CDP estimation \vs RSRP,  which confirms our design choice. Our estimation $\widehat{Q}_{cdp}(y)$ reduces the relative estimation error up to 27\% in the lower reception regime (0-1 vars,  \rsrp $\leq -115$dBm), where the error function being minimized is highly sensitive to predictive performance. 

\paragraph{\em From $Q_{cdp}(y)$ to the RSRP $y^P=y$ domain (\ie transform $ P=(Q,k) \rightarrow P_B=(I,k)$):} 
It is important to highlight  that  our QoS domain methodology can also improves \emph{RSRP prediction itself} for values with high CDP that matter the most. An example is shown in  Fig.~\ref{fig:rsrp_RFs_vs_Qinv_campus}:  we compare the prediction error of $\widehat{y}$  values (RSRP) themselves, \vs inverting $\widehat{Q}_{cdp}(y)$ to return back to the original \rsrp~space. We group the error by signal bars and we observe that the change in learning objective shifts the effort to reducing error where it is most critical (in lower signal strength range). We basically exploit the fact that we can tolerate higher uncertainty at high RSRP (where a large error has little impact on predicted CDP). We can hence view our procedure as allowing us to train on an application-specific loss function, without modifying our underlying learning algorithm.

\begin{figure}[t!]
	\subfigure[\external:Manhattan Uptown \label{fig:cdp_rsrqVSrmse_percell_tai22561}] {\includegraphics[scale = 0.14]{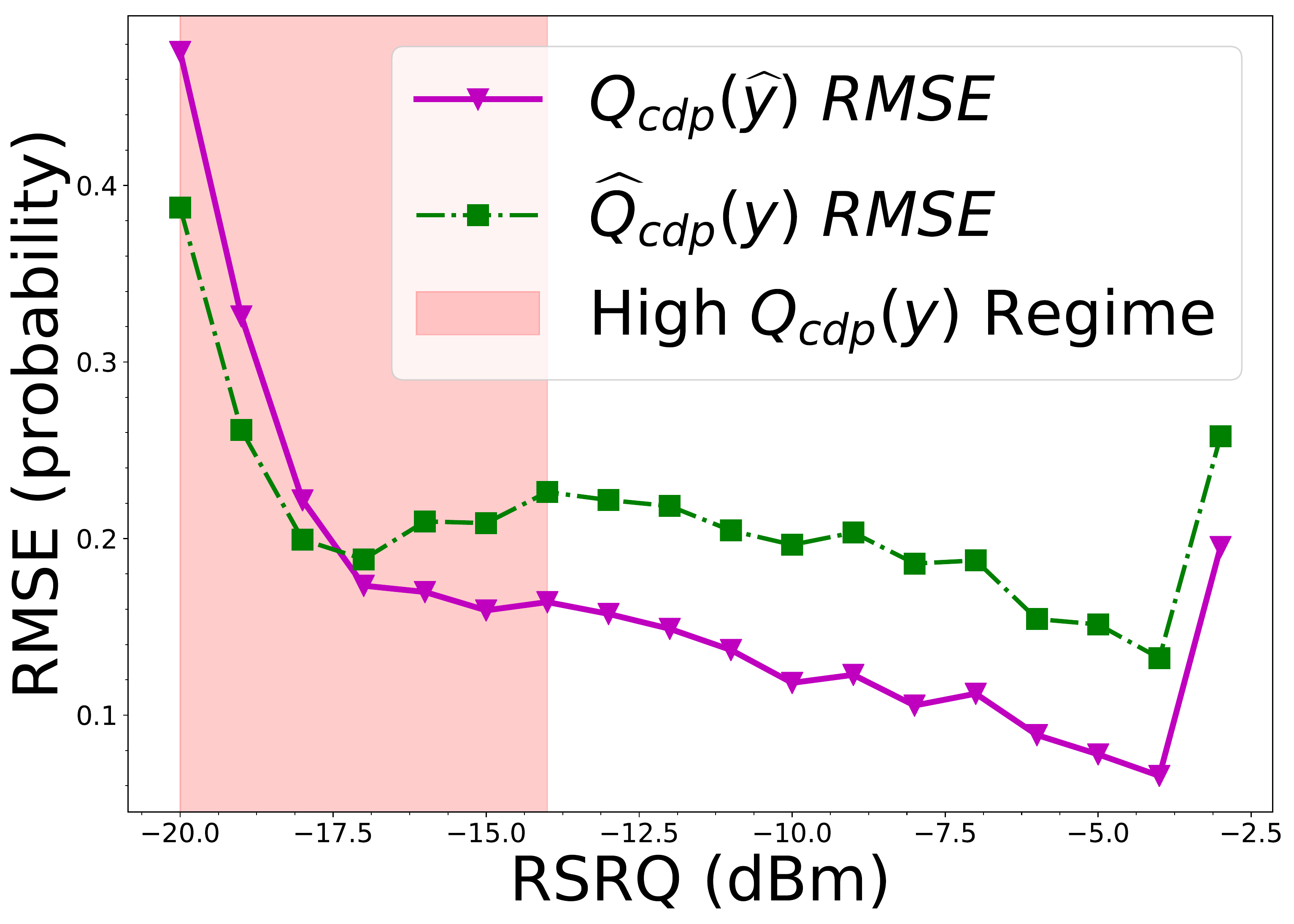}}
	\subfigure[\externalsuburban: Suburb x470 \label{fig:cdp_cqiVSrmse_tai15470}] {\includegraphics[scale = 0.14]{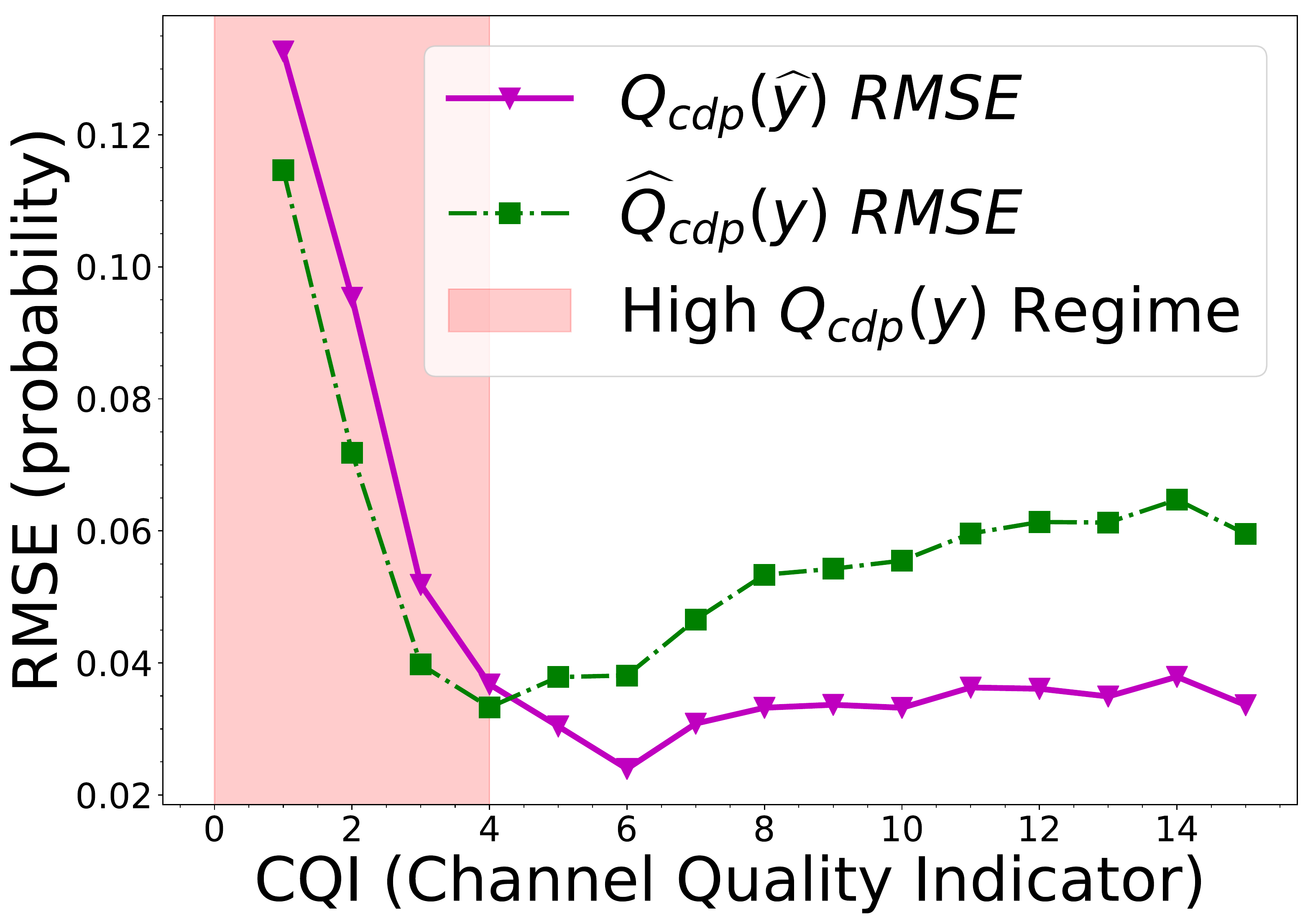}}
	\vspace{-10pt}
	\caption{\footnotesize Call Drop Probability \cdp{$y$} estimation for \externalcitiesdataset. (a)  RSRQ ($y = $ \rsrq), Models Per \cid. (b)  CQI ($y = $ \cqi).}
	\label{fig:cdp_kpisVSrmse_externaldataset} 
\end{figure}

{(b) \externalcitiesdataset:} We also present results for CDP prediction on the \externalcitiesdataset, and using RSRQ \rsrq and CQI \cqi data in addition to RSRP \rsrp. Fig.~\ref{fig:cdp_rsrqVSrmse_percell_tai22561}  and  Fig.~\ref{fig:cdp_cqiVSrmse_tai15470} show $RMSE$ of CDP estimation with RSRQ \rsrq and CQI \cqi respectively. The different KPIs and the use of per-\cid~models in one case (RSRQ \rsrq) do not change the improvements from our technique. We improve in the low KPI $y$ regime up to 0.1 in absolute error value (in the probability domain); in terms of relative error our method $\widehat{Q}_{cdp}(y)$ performs up to 32\% better than the baseline $Q_{cdp}(\widehat{y})$ for CDP estimation. For \cdp{$y$} and \coverage, similar results were observed for other \lteta s for \externalcitiesdataset but omitted due to lack of space.

\begin{figure}[h!]
	\subfigure[CDP with CQI $y=$\cqi  \label{fig:cdp_cqi_logratio}] {\includegraphics[scale = 0.3]{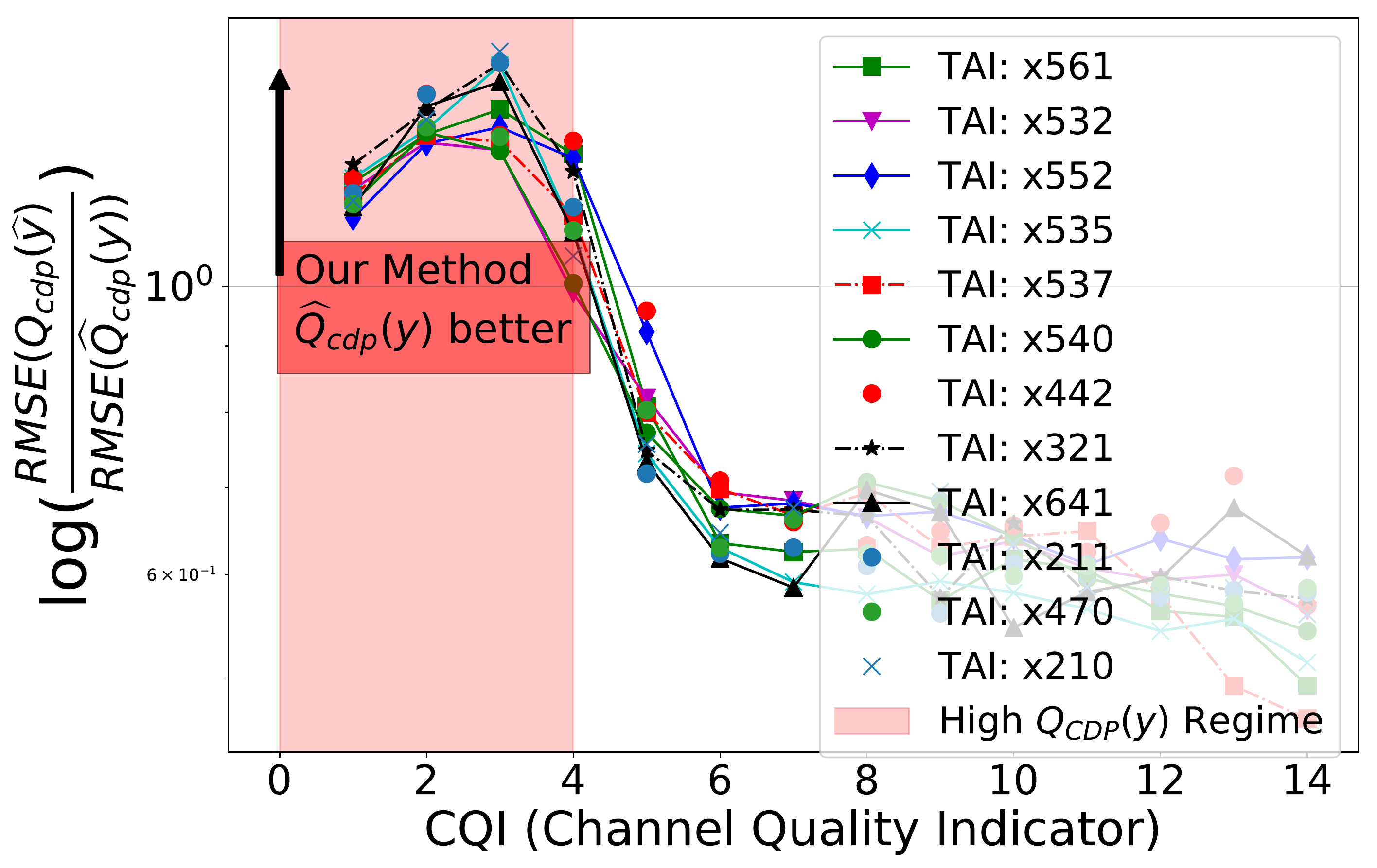}}
	\subfigure[CDP with RSRP $y=$\rsrq  \label{fig:cdp_rsrq_logratio}] {\includegraphics[scale = 0.3]{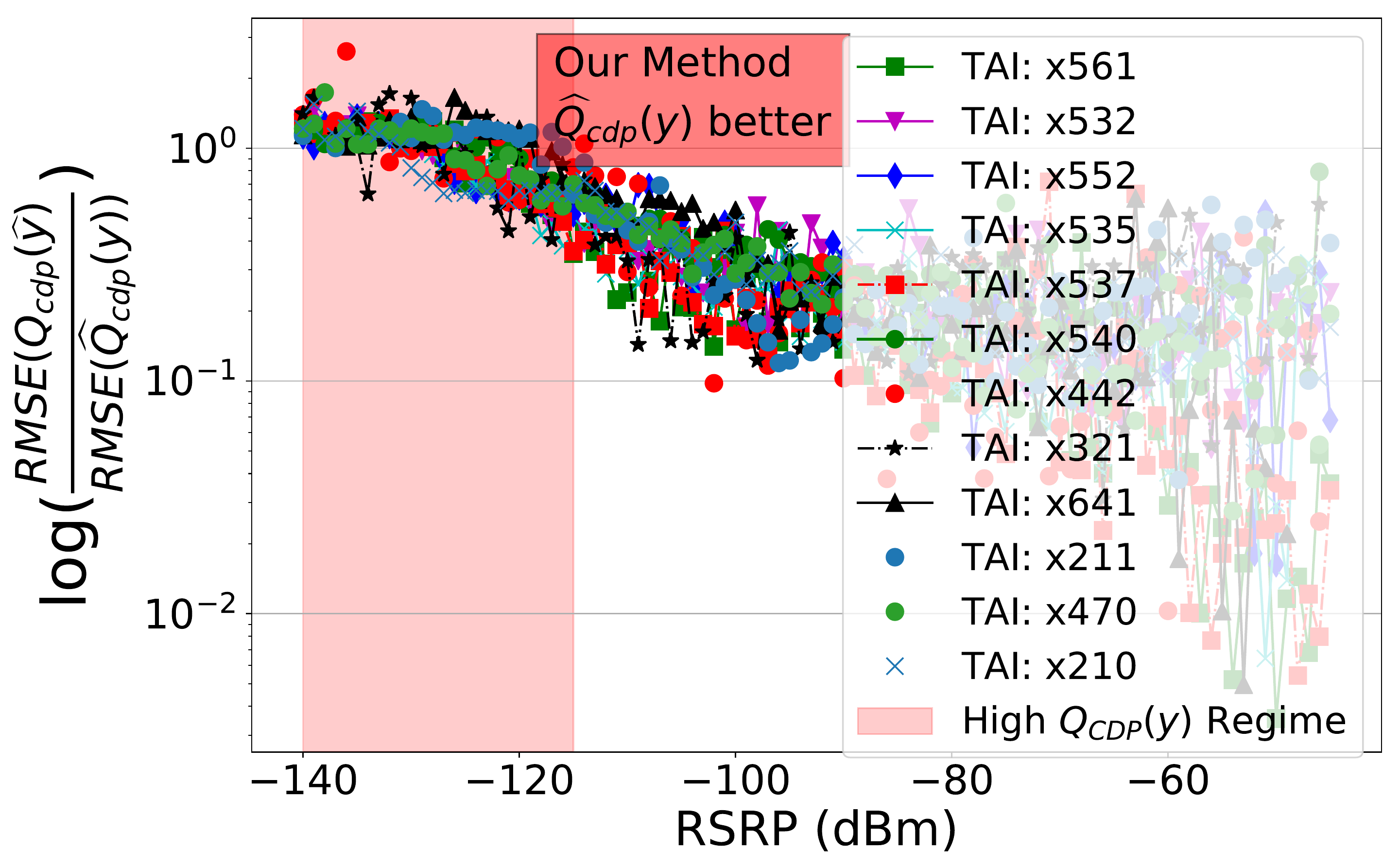}}
	\caption{\small \externalcitiesdataset Call Drop Probability \cdp{$y$} estimation.} 
	\label{fig:cdp_logratios_externaldataset} 
\end{figure}

Fig.~\ref{fig:cdp_logratios_externaldataset} summarizes the performance of CDP estimation for the different \lteta s (\ie areas) available in our dataset. We plot the $\log$-ratio of the $RMSE$ of the baseline \vs our approach. Values greater than 1 indicate improvement for our model as in the other examples, we see that our procedure successfully focuses improvement where it is needed for CDP prediction, rather than wasting statistical power on the high signal strength regime.

\subsubsection{Discussion: why minimizing MSE can be naive.}
In signal strength prediction, an error of few (say 5) dB will not reflect much change in QoS when the user's received signal strength is high (\eg -50 to -60 dBm, see Fig.~\ref{fig:empirical_cdp_vs_kpis}). The user experiences excellent QoS in that regime, and hence even moderately large errors in predicted RSRP would not greatly impact predictions of QoS. In contrast, an error of $5$dB would substantially affect QoS  prediction in the weak reception regime (\eg for -120dBm \vs -125dBm you can notice the large difference in CDP in Fig.~\ref{fig:empirical_cdp_vs_kpis}). For QoS prediction, it can hence be worth trading greater RSRP error in the high-strength regime for lower error in the low-strength regime, as we demonstrated.  Working directly with \Qobj{$y$} alters our application loss function so as to focus performance where it is most needed, but without requiring us to modify the \RFs{} procedure to change its nominal loss function).  The result is improved performance for QoS outcomes, here up to 32\% for the values that matter more to cellular \mno.

\subsection{Results for Importance Sampling $P=(I, W)$}\label{sec:importance_sampling_rsrp}
Next, we evaluate our framework in terms of the reweighted error $\varepsilon_p$. We predict RSRP  
and we compare a default setting \RFs~vs. \RFsWeights~(\ie $w_i$ are set to importance ratio as described in Section \ref{sec:results_setup}).

\subsubsection{$\varepsilon_{u}$ over Uniform Spatial Distribution, \problem{$I,w_u$}.} 

{(a) \inhousedataset:} To calculate the importance ratio $w_u = 1/s(\mathbf{l})$ we estimate $s(\mathbf{l})$ with adaptive bandwidth KDE over the spatial dimensions as we describe in~\ref{sec:importancesampling}. 
  Table~\ref{tab:campus_rsrp_reweighted_uniform} reports the error $\varepsilon_{u}$ for both the default \RFs~predictor and  the \RFsWeights. We observe an improvement of up to 20\% for $\varepsilon_{u}$ for cells that are oversampled in just few locations; the average relative improvement is approx. 5\%, which demonstrates the benefit of readjusting the training loss.

\begin{table}[]
	\centering
	\setlength\tabcolsep{1.5pt} 
	\footnotesize
	\begin{tabular}{|l|r|r|r|r|r|}
		\hline
		\multicolumn{2}{|l|}{\textbf{Cell Characteristics}} & \multicolumn{1}{l|}{\textbf{Default $RFs$ $\rightarrow \widehat{y}$}} & \multicolumn{1}{l|}{\textbf{$RFs_{w_{u}}$ $\rightarrow \widehat{y}_w$}} & \multicolumn{2}{l|}{\textbf{Improvement}}                                      \\ \hline
		\textbf{$cID$}           & \textbf{$N$}             & \textbf{$\sqrt{\varepsilon_u}$}                                       & \textbf{$\sqrt{\varepsilon_u}$}                                         & \multicolumn{1}{l|}{\textbf{Diff.}} & \multicolumn{1}{l|}{\textbf{Diff. (\%)}} \\ \hline \hline
		x922                     & 3955                     & 0.86                                                                  & \textbf{0.69}                                                           & 0.17                                & \textbf{19.6}                            \\ \hline
		x808                     & 12153                    & 1.54                                                                  & \textbf{1.25}                                                           & 0.28                                & \textbf{18.5}                            \\ \hline
		x470                     & 7688                     & 0.71                                                                  & \textbf{0.59}                                                           & 0.12                                & \textbf{17.0}                            \\ \hline
		x460                     & 4069                     & 1.66                                                                  & \textbf{1.44}                                                           & 0.22                                & \textbf{13.1}                            \\ \hline
		x355                     & 29608                    & 1.77                                                                  & \textbf{1.57}                                                           & 0.20                                & \textbf{11.5}                            \\ \hline
		x306                     & 4027                     & 2.21                                                                  & \textbf{2.03}                                                           & 0.18                                & \textbf{8.1}                             \\ \hline
		x901                     & 16049                    & 0.94                                                                  & \textbf{0.91}                                                           & 0.03                                & \textbf{3.4}                             \\ \hline
		x902*                    & 34164                    & 1.93                                                                  & \textbf{1.90}                                                           & 0.03                                & \textbf{1.5}                             \\ \hline
		x914                     & 3041                     & 1.66                                                                  & \textbf{1.64}                                                           & 0.02                                & \textbf{1.0}                             \\ \hline
		x915                     & 4725                     & 1.81                                                                  & \textbf{1.80}                                                           & 0.00                                & \textbf{0.2}                             \\ \hline
		x312                     & 9727                     & 0.64                                                                  & \textbf{0.65}                                                           & -0.01                               & \textbf{-0.6}                            \\ \hline
		x204*                    & 55413                    & 0.91                                                                  & \textbf{0.94}                                                           & -0.03                               & \textbf{-3.2}                            \\ \hline
		x034                     & 1554                     & 2.43                                                                  & \textbf{2.68}                                                           & -0.24                               & \textbf{-10.0}                           \\ \hline
		\textbf{All}             & \textbf{186173}          & \textbf{1.34}                                                         & \textbf{1.28}                                                           & \textbf{0.06}                       & \textbf{4.89}                            \\ \hline
	\end{tabular}
\caption{\footnotesize  \inhousedataset, ~RSRP $y$ prediction: $\varepsilon_u$ Error (\ie reweighted according to the uniform distribution $\equiv$ \problem{$I,w_u$}): (i) Train on Default \RFs{} \vs~ (ii) train on \RFsWeights{} $w_i = w_u \propto \frac{1}{s(\mathbf{l})}$. Models per \cid. For each \cid{} and training case, we pick the best performing adaptive bandwidth KDE for estimating $s(\mathbf{l})$. Our methodology shows improvement up to approx. $20\%$. For cells * with  extremely high sampling density in few locations -see~\cite{alimpertis:19}-, we utilize fixed bandwidth estimation both in space and time. \label{tab:campus_rsrp_reweighted_uniform}}
\end{table}

\begin{figure}
	\centering
	\subfigure[Actual Measurements $s(\mathbf{l})$.  \label{fig:data_sampling_kde}] {\includegraphics[height = 2.1in]{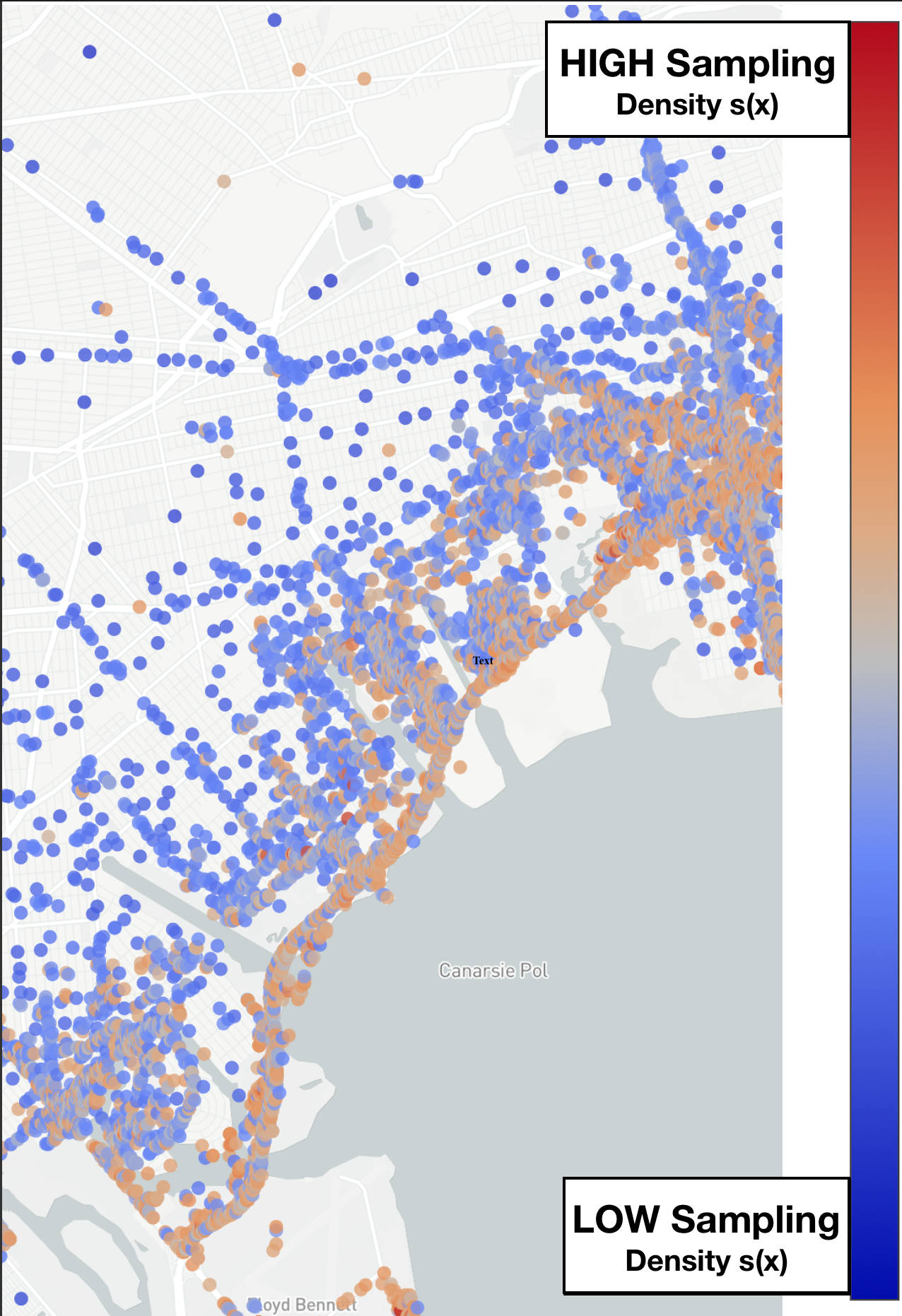}}
	\hspace{20pt}
	\subfigure[Importance sampling.  \label{fig:reweighted_wu}] {\includegraphics[height = 2.1in]{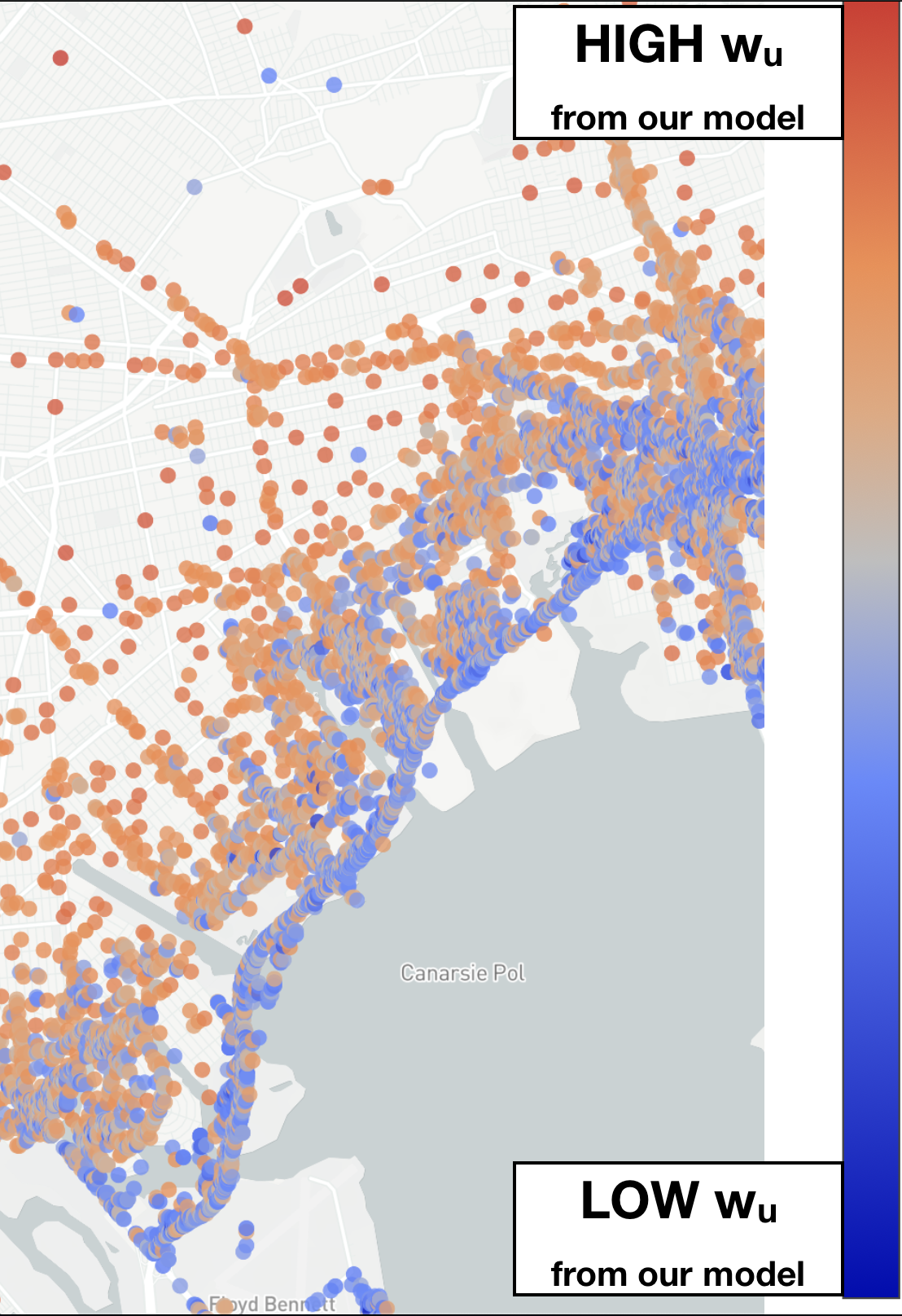}}
	\caption{\footnotesize \externaldataset: RSRP prediction and uniform error $\varepsilon_{u}$. (a) Actual sampling ($s(\mathbf{l})$) estimated by an adaptive bandwidth KDE. (b) Reweighting with $w_u$ from importance sampling. It can be seen that the collected data from the Mobile analytics companies oversample devices during commute (GPS Apps push locations updates - power plugged) and undersample other residential areas. } 
	\vspace{-10pt}
	\label{fig:sampling_vs_reweighting_external} 
\end{figure}

{(b) \externalcitiesdataset:} 
 Fig.~\ref{fig:data_sampling_kde} depicts the sampling distribution $s(\mathbf{l})$ in spatial dimensions (estimated by adaptive bandwidth KDE as we described in~\ref{sec:importancesampling}), in East \external nearby JFK airport. It can be observed that the data are primarily being collected on the highway (Belt Pkwy) adjacent to the sea; the sampling density is much higher compared to nearby residential blocks. Although the specifics of the data collection for \externaldataset are proprietary, we can speculate that the data collection is more frequent when the devices are plugged to power and the users utilize a location navigator app which pushes location updates to other applications;  it is a  common practice to minimize the impact on users' devices ~\cite{alimpertis:17}. Fig.~\ref{fig:reweighted_wu} illustrates the importance ratio weights $w_u$ and how our model readjusts for the sampling-target distribution mismatch. Similar patterns are observed throughout many different areas in \externalcitiesdataset: \eg see the 405 highway in the Long Beach area x210.

Table~\ref{tab:external_rsrp_reweighted_uniform} reports the error $\varepsilon_{u}$ for different \lteta s in \externalcitiesdataset. The average performance improvement by training \RFs$_{w_u}$ is approx. 3\%, with up to 5\% in some units. We also examined the area x532 where the benefit of our method was small and as expected the spatial distribution was indeed approx. uniform (omitted for space limitations). At the other extreme, regions with highly biased data collection  (\ie x540 East \external and x210 Long Beach in \externalsuburban) show the highest error reduction (3.6\% and 5.3\% respectively).
Overall, we find higher gains to reweighting on \inhousedataset, as it is collected from a small number of users and hence more unevenly sampled.  We expect that this feature will be common for small-scale data sets as well as setups with biased sampling because of mobile analytics companies practices, making reweighting especially important to correct for sampling bias.

\begin{table}
	\centering
	\setlength\tabcolsep{0.5pt} 
	\footnotesize
		\begin{tabular}{|l|r|l|r|r|l|r|}
			\hline
			\multicolumn{3}{|l|}{\textbf{LTE-TA Characteristics}}                                                          & \multicolumn{1}{l|}{\textbf{$RFs$ $: \widehat{y}$}}& \multicolumn{1}{l|}{\textbf{$RFs_{w_{u}}$ $:  \widehat{y}_w$}} & \multicolumn{2}{l|}{\textbf{Improvement}} \\ \hline
			\textbf{$TAI$} & \textbf{$N$}     & \textbf{Info}                                                              & \textbf{$\sqrt{\varepsilon_u}$}                                       & \textbf{$\sqrt{\varepsilon_u}$}                                         & \textbf{Diff.}    & \textbf{Diff. (\%)}   \\ \hline \hline
			x210           & 197521           & \begin{tabular}[c]{@{}l@{}}Long Beach\\ Lakewood\end{tabular}              & 4.06                                                                  & 3.84                                                                    & 0.21              & \textbf{5.3}          \\ \hline
			x552           & 97942            & Eastern Brooklyn                                                           & 5.38                                                                  & 5.12                                                                    & 0.26              & \textbf{4.9}          \\ \hline
			x540           & 136105           & E. Long Island                                                             & 5.01                                                                  & 4.83                                                                    & 0.18              & \textbf{3.6}          \\ \hline
			x535           & 121159           & W. Queens                                                                  & 5.36                                                                  & 5.17                                                                    & 0.19              & \textbf{3.6}          \\ \hline
			x641           & 10663            & \begin{tabular}[c]{@{}l@{}}Covina \\ Hacienda Heights\end{tabular}         & 1.8                                                                   & 1.74                                                                    & 0.06              & \textbf{3.5}          \\ \hline
			x561           & 62448            & Manhattan Mid.                                                         & 5.64                                                                  & 5.46                                                                    & 0.18              & \textbf{3.2}          \\ \hline
			x470           & 198252           & \begin{tabular}[c]{@{}l@{}}LA Downtown\\ Hollywood\end{tabular}            & 4.56                                                                  & 4.43                                                                    & 0.13              & \textbf{2.8}          \\ \hline
			x211           & 77049            & Suburban S. LA                                                             & 4.06                                                                  & 3.96                                                                    & 0.1               & \textbf{2.4}          \\ \hline
			x442           & 14538            & \begin{tabular}[c]{@{}l@{}}Manhattan Uptown \\ Queens - Bronx\end{tabular} & 3.23                                                                  & 3.19                                                                    & 0.05              & \textbf{1.5}          \\ \hline
			x537           & 37247            & \begin{tabular}[c]{@{}l@{}}Manhattan \\Midtown East\end{tabular}                                                    & 7.62                                                                  & 7.53                                                                    & 0.09              & \textbf{1.1}          \\ \hline
			x321           & 5111             & Eastern Brooklyn                                                           & 3.87                                                                  & 3.83                                                                    & 0.04              & \textbf{1.1}          \\ \hline
			x532           & 136508           & Brooklyn                                                                   & 5.46                                                                  & 5.43                                                                    & 0.03              & \textbf{0.5}          \\ \hline
			\textbf{ALL}   & \textbf{1094543} & \textbf{NYC \& LA}                                                         & \textbf{4.88}                                                         & \textbf{4.72}                                                           & \textbf{0.16}     & \textbf{3.16}         \\ \hline
		\end{tabular}
	\caption{\footnotesize \externalcitiesdataset,   RSRP $y$ prediction: $\varepsilon_u$ Error (\ie reweighted according to the uniform distribution $\equiv$ \problem{$I,w_u$}): (i) Train on Default \RFs{} \vs~ (ii) train on \RFsWeights{} $w_i = w_u \propto \frac{1}{s(\mathbf{l})}$. Models per \lteta. We use adaptive bandwidth KDE for estimating $s(\mathbf{l})$\cite{lichman:14}. Our methodology shows improvement up to $5.3\%$. \label{tab:external_rsrp_reweighted_uniform}}
	\vspace{-5pt}
\end{table}

\subsubsection{$\varepsilon_{P}$ : reweighting for Population Density \problem{$I,w_d$}.}
Weighing errors by local population density, instead of uniformly, results in a metric that places more emphasis on accuracy in regions where more potential users reside. To that end, we utilize public APIs to retrieve the census data and estimate the population density \pop{$\mathbf{l}_i$}. Reweighted \errorpopulation~for RSRP data by using the weighted train \RFs$_\text{\wP}$ \vs the the default \RFs{ } show an improvement of up to 5.7\%.

Table~\ref{tab:qosdomain_jointly_reweightedflavors_population} reports the error weighted proportional to the actual user population density  over the same  spatial area.
\begin{table}[h]
	\centering
	\setlength\tabcolsep{1.5pt} 
	\scriptsize

	\begin{tabular}{|c||l|l|l|l|l|}
		\hline
		\multirow{4}{*}{\textbf{\begin{tabular}[c]{@{}c@{}}KPI: CQI\\ All LTE-TA \\regions\end{tabular}}}  & \textbf{Training Options}                                             & \multicolumn{2}{l|}{\textbf{$y$ domain $\rightarrow$ $Q(\widehat{y})$}} & \multicolumn{2}{l|}{\textbf{$Q(y)$ domain}} \\ \cline{2-6} 
		& \textbf{$w_i = 1,  \forall i$}                                        & $Q_{cdp}(\widehat{y})$                      & 0.0107                    & $\widehat{Q}_{cdp}(y)$        & 0.0088      \\ \cline{2-6} 
		& \textbf{$w_i = w_P \propto \frac{\text{\pop{$\mathbf{l}_i$}}}{s(\mathbf{l}_i)} $} & $Q_{cdp}(\widehat{y}_w)$                    & 0.0109                    & $\widehat{Q}^w_{cdp}(y)$      & 0.0085      \\ \cline{2-6} 
		& \textbf{Relative Difference}                                          & \multicolumn{2}{r|}{-2.3\%}                                             & \multicolumn{2}{r|}{3.07\%}                 \\ \hline \hline
		\multirow{3}{*}{\textbf{\begin{tabular}[c]{@{}c@{}}KPI: RSRP\\ All LTE-TA \\regions\end{tabular}}} & \textbf{$w_i = 1, \forall i$}                                         & $Q_{cdp}(\widehat{y})$                      & 0.0045                    & $\widehat{Q}_{cdp}(y)$        & 0.0036      \\ \cline{2-6} 
		& \textbf{$w_i = \text{\wP} \propto \frac{\text{\pop{$\mathbf{l}_i$}}}{s(\mathbf{l}_i)} $} & $Q_{cdp}(\widehat{y}_w)$                    & 0.0047                    & $\widehat{Q}^w_{cdp}(y)$      & 0.0034      \\ \cline{2-6} 
		& \textbf{Relative Difference}                                          & \multicolumn{2}{r|}{-5\%}                                               & \multicolumn{2}{r|}{4\%}                    \\ \hline
	\end{tabular}

\caption{\footnotesize \externalcitiesdataset. The error   
	\errorpopulation ~ is re-weighted according to the population distribution) is computed on the $Q$ domain,  \ie \problem{$Q_{cdp},\text{\wP}$}. When predicting $\widehat{y}$ with weights and then converting to $Q(y)$,   information is lost from the transformation.  When training with the importance sampling weights, then predicting $\widehat{Q}(y)$, can further reduce the error up to $5\%$. \label{tab:qosdomain_jointly_reweightedflavors_population}}
\end{table}

Table~\ref{tab:external_rsrp_reweighted_population} includes the reweighted \errorpopulation~for RSRP data by using the default RFs vs. the weighted train \RFs$_\text{\wP}$ ; we see perfor- mance improvement up to 5.7
\begin{table}[h]
		\centering
	\setlength\tabcolsep{1.5pt} 
	\vspace{-5pt}
	\scriptsize

	\begin{tabular}{|l|r|l|r|r|r|r|}
		\hline
			\multicolumn{3}{|l|}{\textbf{LTE-TA Characteristics}}                                                          & \multicolumn{1}{l|}{\textbf{$RFs$ $\rightarrow \widehat{y}$}} & \multicolumn{1}{l|}{\textbf{$RFs_{w_{P}}$ $\rightarrow \widehat{y}_w$}} & \multicolumn{2}{l|}{\textbf{Improvement}} \\ \hline
\textbf{$TAI$} & \textbf{$N$}     & \textbf{Info}                                                              & Default \textbf{$\sqrt{\varepsilon_P}$}                                       & \textbf{$\sqrt{\varepsilon_P}$}                                         & \textbf{Diff.}    & \textbf{Diff. (\%)}   \\ \hline \hline
		x561          & 63303        & Manhattan Midtown                                                          & 7.23                            & 6.82                                                               & 0.41                 & \textbf{5.7}             \\ \hline
		x321           & 7014         & Eastern Brooklyn                                                           & 4.94                            & 4.8                                                                & 0.14                 & \textbf{2.8}             \\ \hline
		x535          & 122071       & W. Queens                                                                  & 6.03                            & 5.87                                                               & 0.15                 & \textbf{2.5}             \\ \hline
		x552          & 98240        & Eastern Brooklyn                                                           & 5.35                            & 5.29                                                               & 0.06                 & \textbf{1.2}             \\ \hline
		x532          & 137962       & Brooklyn                                                                   & 6.24                            & 6.22                                                               & 0.02                 & \textbf{0.3}             \\ \hline
		x537          & 37964        & \begin{tabular}[c]{@{}l@{}}Manhattan Midtown \\ East\end{tabular}                                                    & 8.82                            & 8.81                                                               & 0.01                 & \textbf{0.1}             \\ \hline
		x540          & 138495       & E. Long Island                                                             & 5.09                            & 5.09                                                               & 0.00                 & \textbf{0.0}             \\ \hline
		x442           & 16372        & \begin{tabular}[c]{@{}l@{}}Manhattan Uptown \\ Queens - Bronx\end{tabular} & 3.97                            & 4.21                                                               & -0.24                & \textbf{-6.1}            \\ \hline
		\textbf{ALL}   & \textbf{621421} & \textbf{NYC}                                                         & \textbf{5.98}                                        & \textbf{5.90}                                                               & \textbf{0.08}        & \textbf{1.35}            \\ \hline
	\end{tabular}
	\caption{\footnotesize \externalcitiesdataset  RSRP $y$ prediction: \errorpopulation~Error (\ie reweighted according to the population distribution): (i) Train on Default \RFs{} \vs~ (ii) train on \RFsWeights{} $w_i = \text{\wP} \propto \frac{\text{\pop{$\mathbf{l}$}}}{s(\mathbf{l})}$. Models per \lteta. We use adaptive bandwidth KDE for estimating  $s(\mathbf{l})$\cite{lichman:14}. Our methodology shows improvement up to $5.7\%$. \label{tab:external_rsrp_reweighted_population}}
\end{table}

\subsection{Reweighted Error for QoS functions: $P=(Q,W)$ \label{sec:results-reweighting}}

So far, we have separately evaluated the improvement from (1) predicting QoS directly and (2) re-weighting by importance ratio.
Here, we combine the two and calculate the reweighted error $\varepsilon_p$ (how we handle the input space) for a QoS function (how we handle the output space) of interest. Due to lack of space, we only show results for \cdp{y}. 
 In Table \ref{tab:customfunctions_methodology_v2}, we show four cases to be compared. First, $Q_{cdp}(\widehat{y})$ is the baseline, where we first predict the signal map value $y$ of interest and then get an estimate of the CDP. Second, $\widehat{Q}_{cdp}(y)$ is our prediction directly on the function of interest. Third, we can train a weighted \RFsWeights{} for $y$, and get $Q_{cdp}(\widehat{y}_w)$. Last, we can have $\widehat{Q}^w_{cdp}(y)$ which is the weighted trained model \RFsWeights{} for estimating CDP (\ie \problem{$Q_{cdp},W$}).

Table~\ref{tab:qosdomain_jointly_reweightedflavors_uniform} reports the errors for uniform loss over a spatial area, and shows  improvements up to 5.5\%. Interestingly, the baseline  performance deteriorates when we train on the adjusted weights. It tries to minimize MSE for $y$, therefore the weights can  have very little or even negative effect for mapping back to CDP. Similar results are observed for error proportional to user population density, but omitted due to lack of space.
This demonstrates the importance of choosing the loss function, jointly controlled by $w$ and $Q$, to optimize performance for a specific prediction problem.

\begin{table}
		\centering
	\setlength\tabcolsep{1.5pt} 
	\scriptsize

	\begin{tabular}{|c||l|l|l|l|l|}
		\hline
		\multirow{4}{*}{\textbf{\begin{tabular}[c]{@{}c@{}}KPI: CQI\\ All LTE-TA\\ regions\end{tabular}}}  & \textbf{Training Options}                               & \multicolumn{2}{l|}{\textbf{$y$ domain $\rightarrow$ $Q(\widehat{y})$}} & \multicolumn{2}{l|}{\textbf{$Q(y)$ domain}} \\ \cline{2-6} 
		& \textbf{$w_i = 1,  \forall i$}                          & $Q_{cdp}(\widehat{y})$                       & 0.018                    & $\widehat{Q}_{cdp}(y)$        & 0.0169      \\ \cline{2-6} 
		& \textbf{$w_i = w_u \propto \frac{1}{s(\mathbf{l}_i)} $} & $Q_{cdp}(\widehat{y_w})$                     & 0,018                    & $\widehat{Q}^w_{cdp}(y)$      & 0.0160      \\ \cline{2-6} 
		& \textbf{Relative Difference}                            & \multicolumn{2}{r|}{0.5\%}                                              & \multicolumn{2}{r|}{5.5\%}                  \\ \hline \hline
		\multirow{3}{*}{\textbf{\begin{tabular}[c]{@{}c@{}}KPI: RSRP\\ All LTE-TA\\ regions\end{tabular}}} & \textbf{$w_i = 1, \forall i$}                           & $Q_{cdp}(\widehat{y})$                       & 0.028                    & $\widehat{Q}_{cdp}(y)$        & 0.023       \\ \cline{2-6} 
		& \textbf{$w_i = w_u \propto \frac{1}{s(\mathbf{l}_i)} $} & $Q^w_{cdp}(\widehat{y})$                     & 0.029                    & $\widehat{Q}^w_{cdp}(y)$      & 0.022       \\ \cline{2-6} 
		& \textbf{Relative Difference}                            & \multicolumn{2}{r|}{-2\%}                                               & \multicolumn{2}{r|}{2.3\%}                  \\ \hline
	\end{tabular}

\caption{\footnotesize \problem{$Q_{cdp},w_u$},  \externalcitiesdataset, error $\varepsilon_u$ (\ie reweighted according to the uniform distribution), results on the $Q$ domain. Predicting $\widehat{y}$ with weights and then converting to $Q(y)$ does not help because information is lost from the transformation. Predicting $\widehat{Q}(y)$ after training with the importance sampling weights further reduces error up to $5\%$. \label{tab:qosdomain_jointly_reweightedflavors_uniform}}
\vspace{-10pt}
\end{table}


\subsection{Data Shapley Results\label{sec:dshap_results}}

In this section, we compute the Data Shapley values (using the techniques presented in Section \ref{sec:shapley-theory} and Appendix A1) of datapoints in  our datasets. For each  prediction problem $P=(Q,W)$ and dataset of interest,  we compute the performance score $V$, and eventually the shapley value $\phi_i$ value of every measurement datapoint $i$ in that training dataset. We then order all  datapoints in decreasing shapley values, and remove datapoints with negative or low values.
First, we show that by removing measurements  with negative Shapley values, we can actually {\em improve the prediction} performance up to 30\%; the intuition is that these are noisy/erroneous measurements. Second, we show that we can further remove a large percentage of data points (up to 65\%, depending on the dataset and problem $P$) with low Shapley values, while maintaining high prediction performance. The latter enables {\em data minimization} that can practically improve privacy, and can reduce the cost of collecting, storing, uploading or buying  cellular measurements. 

{\em Data Minimization Setup:} As we described in Sec. \ref{sec:shapley-theory}, we utilize the TMC-Shapley algorithm to calculate the data Shapley values $\phi_i$ per training data point $(\text{\features}_i, y_i)$, for the problem of coverage indicator classification $\widehat{Q_c}(y)$, \ie whether there is  coverage in a location or not (as per  Sec.~\ref{sec:numerical_results_qos}). We remove batches of 5\% of $\mathcal{D}_{train}$ starting from the least valuable (\ie lowest $\phi_i$). At each step, we re-train the \RFsAll~model with the remaining $\mathcal{D}_{train}$ and calculate the performance of the prediction on the $\mathcal{D}_\text{held-out}$ data.

\begin{figure}[t!]
	\centering
		\subfigure[\footnotesize Cell x$034$. \label{fig:campus_dshap_removlowvalue_x034}] {\includegraphics[scale = 0.27]{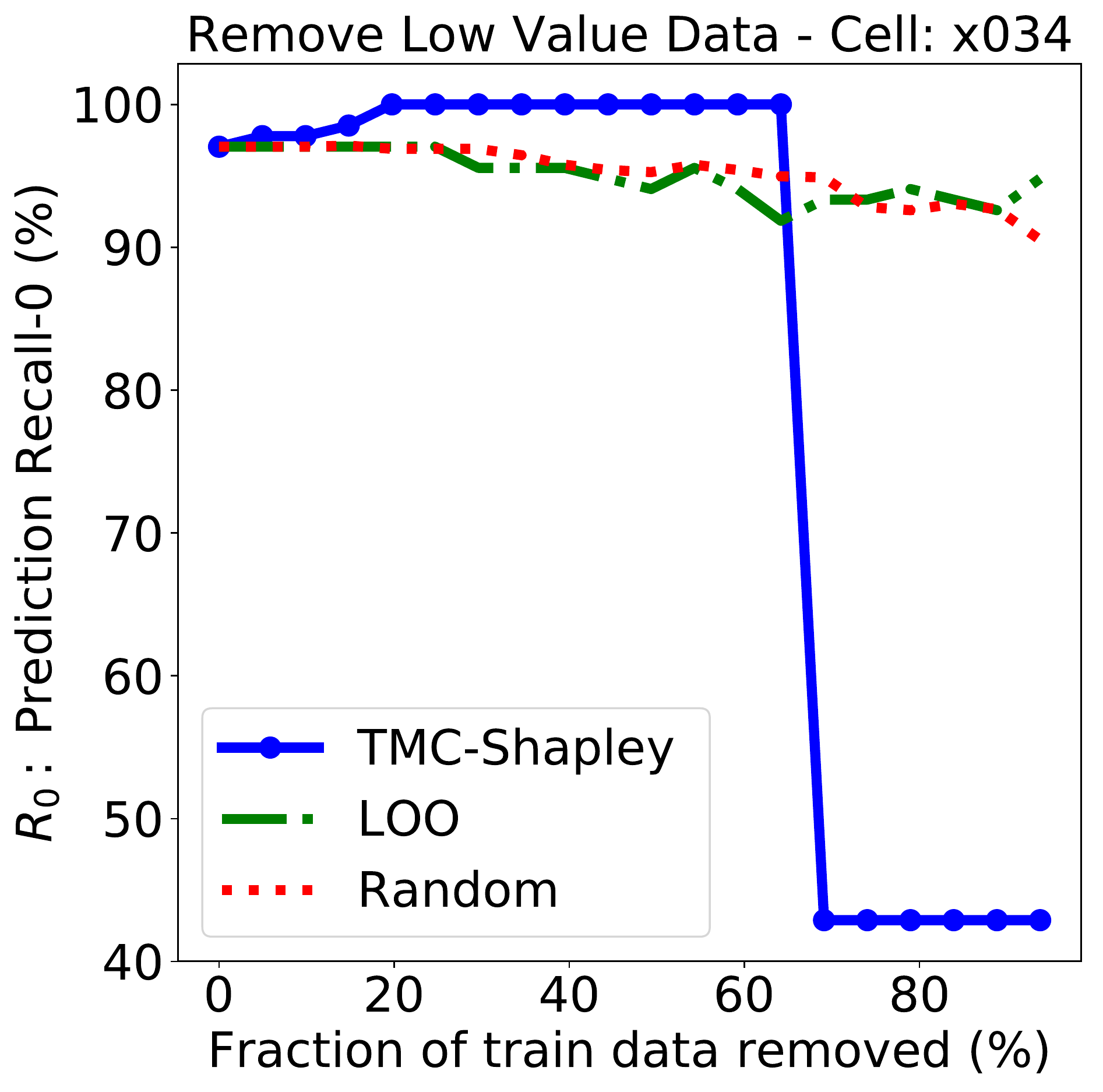}}
		\subfigure[\footnotesize Cell x$312$. \label{fig:campus_dshap_removlowvalue_x312}] {\includegraphics[scale = 0.27]{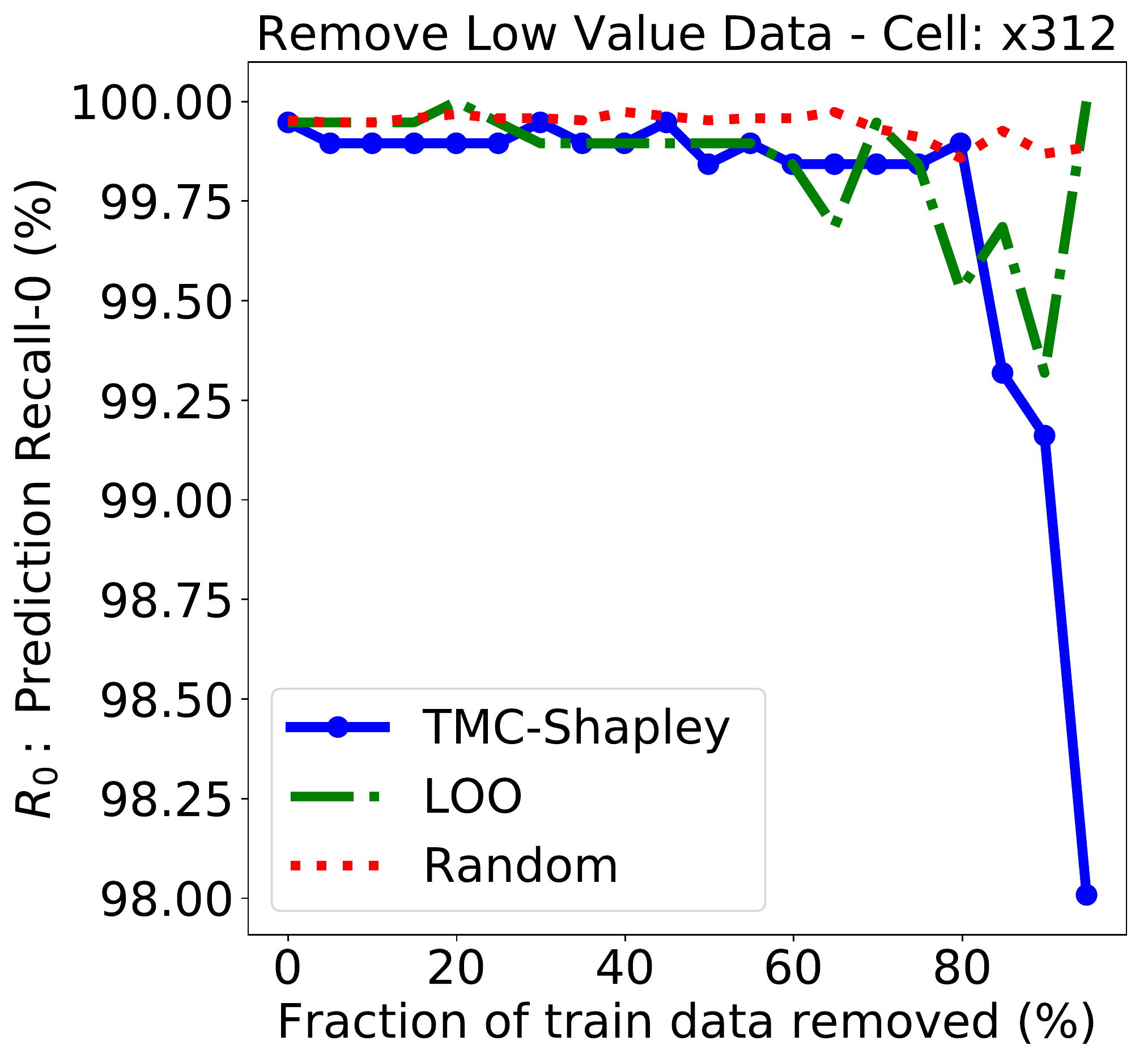}} \\
	\caption{\inhousedataset: Removing low valued data points (for Data-Shapley, LOO and Random) affects Recall $R_0$. \label{fig:campus_dshap_removlowvalue} }
\end{figure}

{\em Baselines for Comparison:} To compare against data Shapley valuation, we consider two baselines.  First, Leave-one-out (LOO) \cite{dshapIcml:2019}, calculates the datum's value by leaving it out and calculating the score $V$, \ie $\phi_i^{\text{LOO}} = V(D) - V(D - \{i\})$. Second, we remove randomly  batches of $\mathcal{D}_{train}$ at each step.

\subsubsection{Effect of removing low value data on Recall.}\label{sec:dshap_results_removelow}
\begin{figure*}[t!]
\centering{
	\subfigure[\footnotesize Removing \% Data. \label{fig:campus_dshap_detailed_removlowvalue_x901}] {\includegraphics[width= 1.7in, height=1.6in]{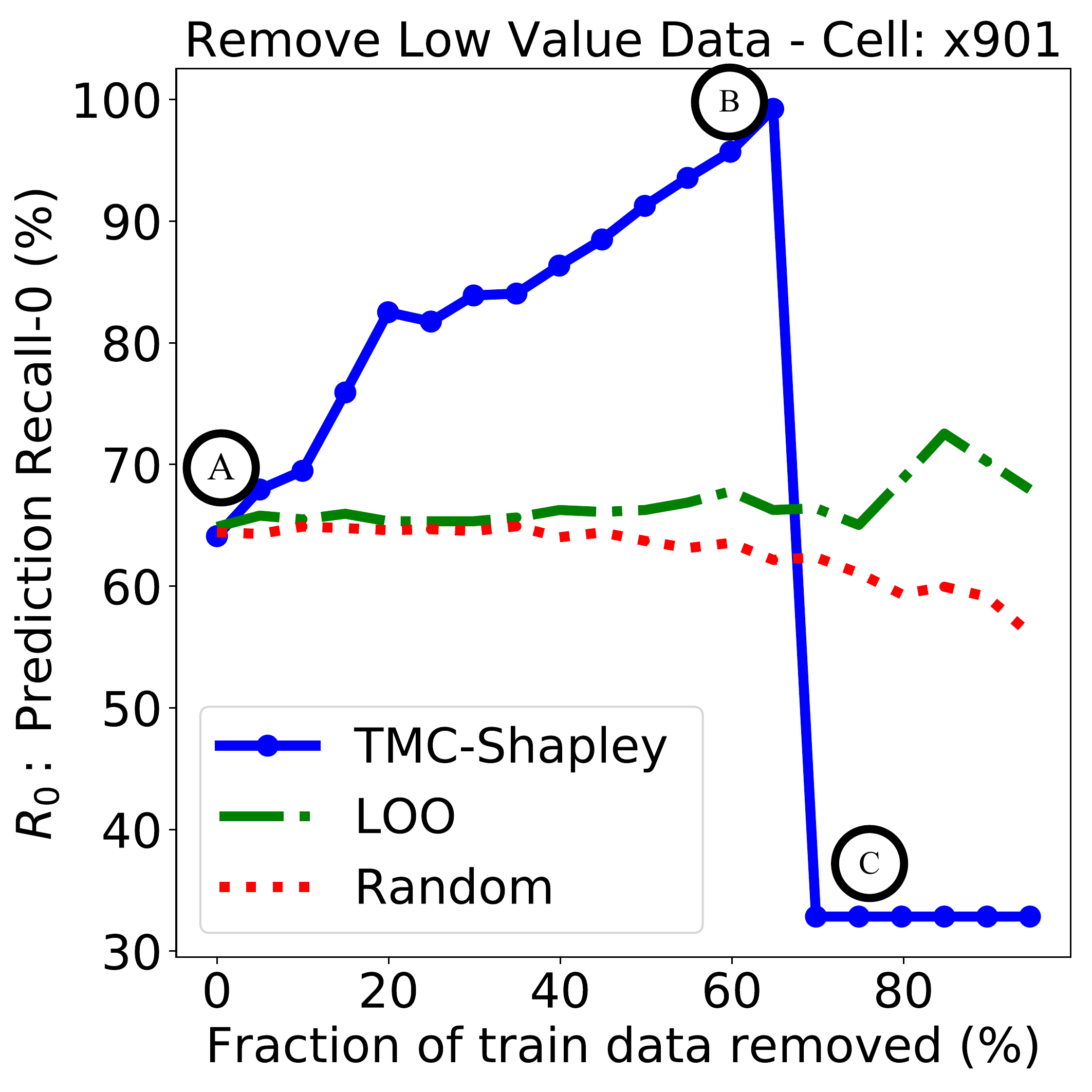}}
	\subfigure[\footnotesize Label A: all data in cell x901. \label{fig:campus_dshap_x901_phival_cdf_labelA}] {\includegraphics[width= 1.7in]{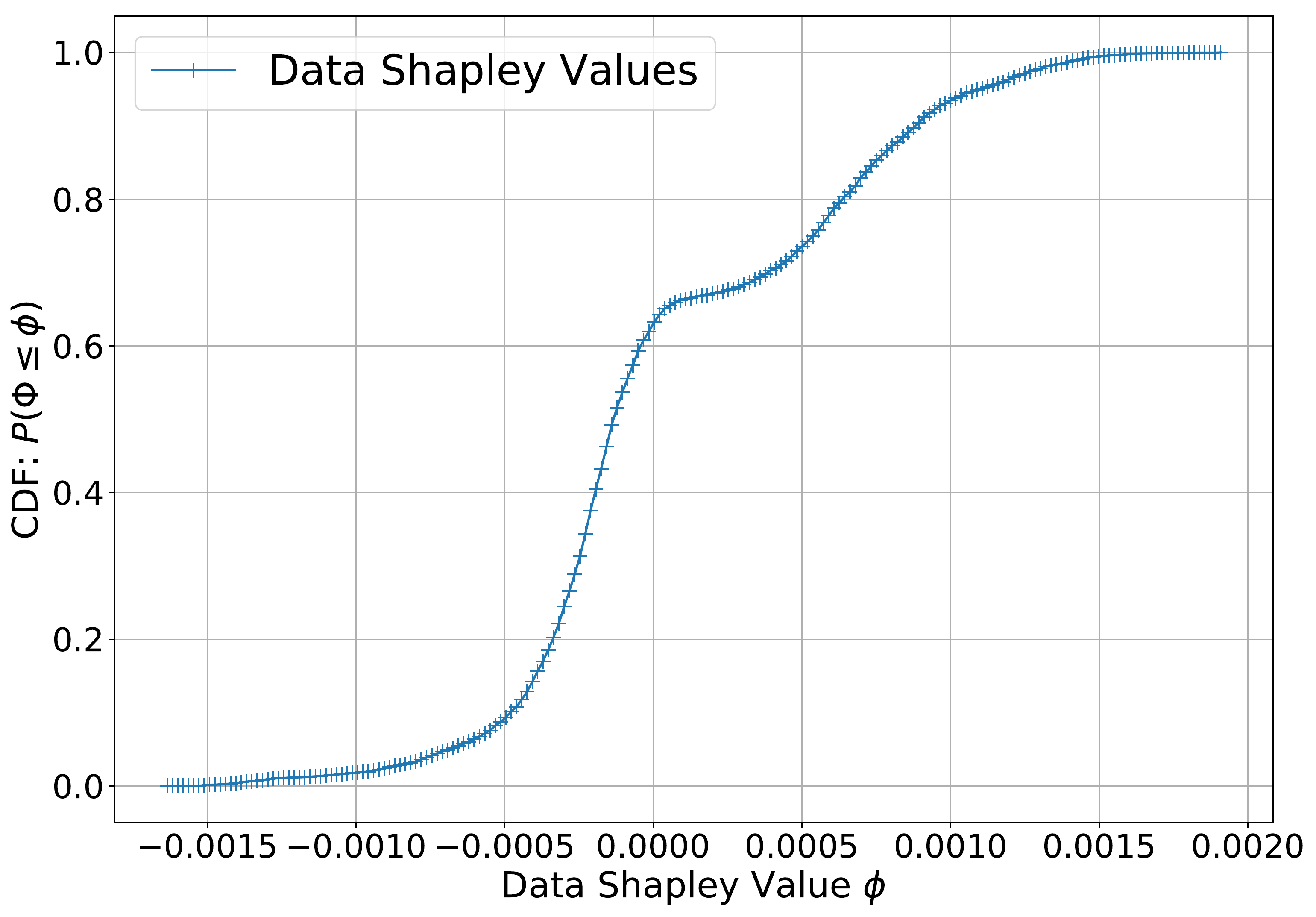}}
	\subfigure[\footnotesize data between Labels A and Label~B.\label{fig:campus_dshap_x901_phival_cdf_deluptolabelB}] 
	{\includegraphics[width= 1.7in]{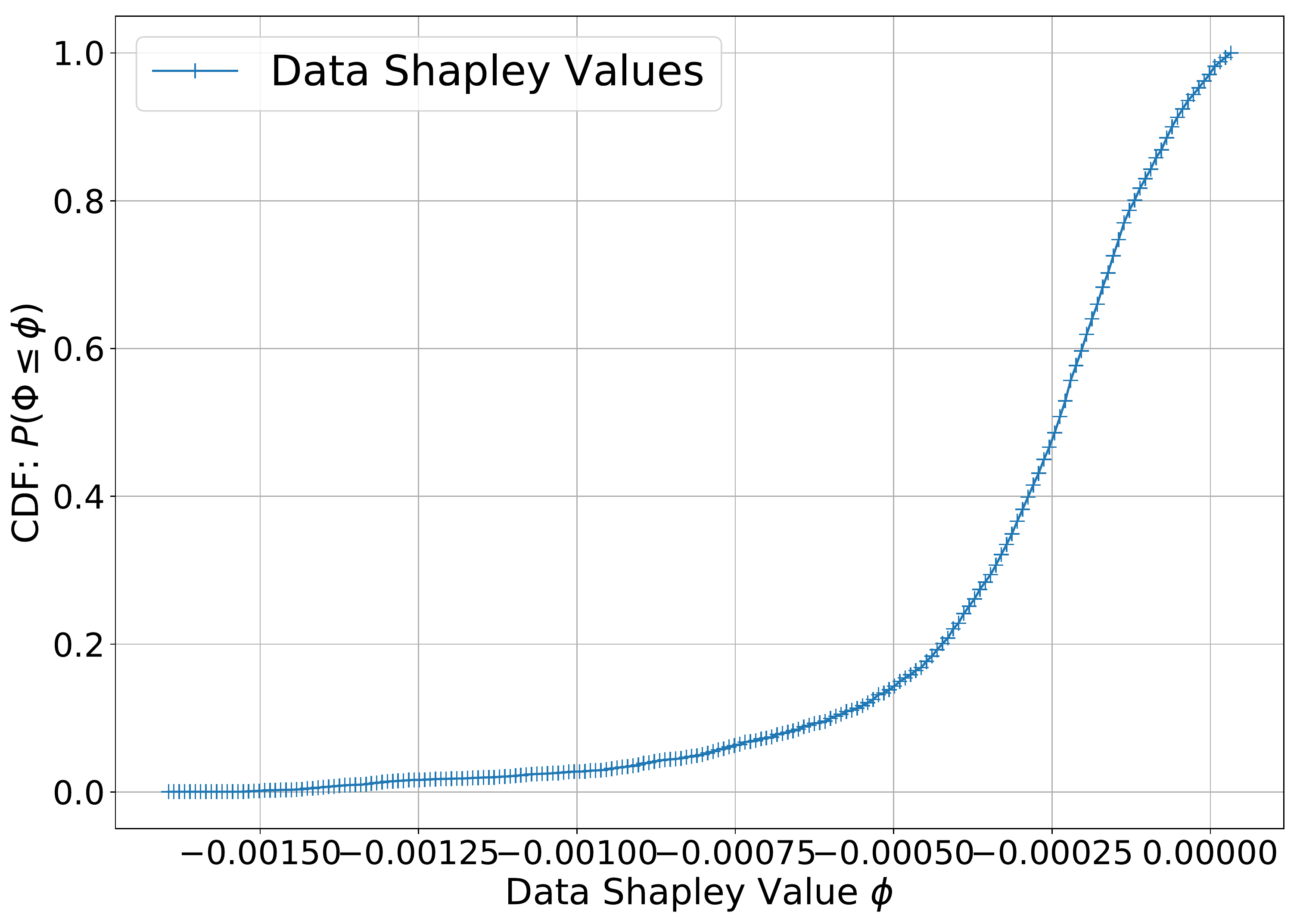}}
	\subfigure[\footnotesize Data at Label B. \label{fig:campus_dshap_x901_phival_lcdf_abelB}] {\includegraphics[width= 1.7in]{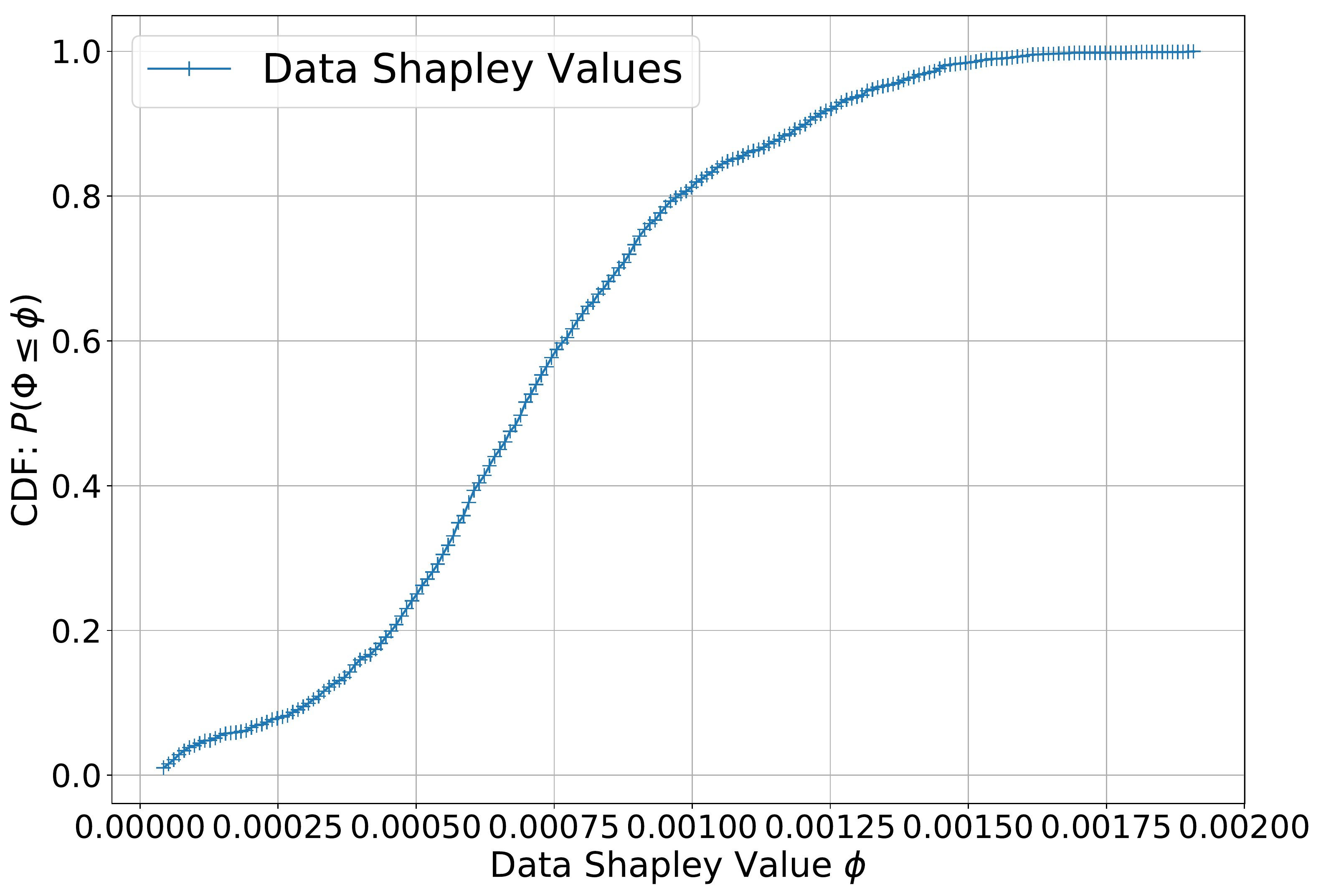}}
	\vspace{-10pt}
	\caption{{\bf Data minimization for \inhouse Cell x901.} {\bf (a)} Removing \% datapoints using different techniques: data Shapley, Leave-One-Out, Random. As we start removing data with low data Shapley (from A to B), recall improves. If we remove more data beyond a certain point (B), recall sharply decreases (from B to C). {\bf (b)(c)(d)} CDFs  of Data Shapley values $\phi_i$ fr the points among labels A,B,C in (a).  \label{fig:campus_dshap_x901_phival_CDFs} }
	}
\end{figure*}

For the \inhousedataset, Figures~\ref{fig:campus_dshap_removlowvalue}(a), \ref{fig:campus_dshap_removlowvalue}(b) and \ref{fig:campus_dshap_detailed_removlowvalue_x901} present results for three representative cells (c034, x312, x901), in terms of the recall $R_0$ (see Sec.~\ref{sec:numerical_results_qos} for its utility for \mno), as a function of the percentage of data removed from  $\mathcal{D}_{train}$, by all discussed methods (TMC-Shapley, LOO and Random). TMC-Shapley's performance either improves or remains the same when we remove low value data points compared to LOO and Random. There are two plausible explanations. First, the batches with low valued $\mathcal{D}_{train}$ contain outliers and corrupted data; the data Shapley has correctly identified these points compared to LOO which does not show any benefit. Second, the data points with low $\phi_i$ do not have much predictive power to maximize the \emph{defined performance} metric of interest for the \emph{particular} learning task; essentially their removal lets the best suited data points to train the predictor. Very interestingly, after a certain threshold, TMC-Shapley's performance drops with just a removal of single batch; this subset of points (highly ``influential'' points) hold significant predictive power. In contrast, by removing data randomly we keep bad quality data, however we might also keep some of these ``influential'' points and that explains that Random's performance neither improves nor decays fast.

Let us discuss in more depth Fig.\ref{fig:campus_dshap_detailed_removlowvalue_x901}, which presents data minimization results for cell x901 of the \inhousedataset. The removal of low valued data according to TMC-Shapley's valuation improves recall $R_0$ compared to Random and LOO. Moreover, in Fig.~\ref{fig:campus_dshap_detailed_removlowvalue_x901} we annotate with the label ``A'' the beginning of the process (\ie still using the entire $\mathcal{D}_{train}$); label ``B'' indicates the step where $65\%$ of the data have been removed and $R_0$ has achieved its maximum value. Fig.~\ref{fig:campus_dshap_x901_phival_CDFs} depicts the $\phi_i$ CDF values for all  $\mathcal{D}_{train}$ (\ie label ``A''). Interestingly,  upon closer inspection of the data, as the CDF in Fig.~\ref{fig:campus_dshap_x901_phival_cdf_deluptolabelB} shows, the data points removed between ``A''  and ``B'' have overwhelming negative values.  Fig. \ref{fig:campus_dshap_detailed_removlowvalue_x901} shows the CDF of $\phi_i$ at label ``B'' (\ie $\mathcal{D}_{train}$ for the max $R_0$ achieved) and we observe that are all positive.  Fig.~\ref{fig:campus_dshap_x901_phival_cdf_deluptolabelB} includes a few positive values that were removed and $R_0$ was keep increasing. This occurs because the TMC-Shap algorithm itself is an approximation of the closed-form data Shapley $\phi_i$ (eq.~\ref{eq:datashapleyform}) and we relaxed the convergence rates due to the size of our datasets.

\begin{table}[]
	\centering
	\scriptsize
	\setlength\tabcolsep{1pt} 
	\begin{tabular}{|l|l|l|l|l|l|l|l|l|}
	\hline
	\begin{tabular}[c]{@{}l@{}}\% Train Data \\ Removed\end{tabular} & $N$  & $R_0$ & userID-0 & UserID-1 & 0s   & 1s &  $\widehat{0}$ & $\widehat{1}$   \\ \hline
	0.0                                                             & 5777 & 0.64  & 5521     & 256      & 1938 & 3839 &541 &1384  \\ \hline
	0.05                                                            & 5489 & 0.68  & 5246     & 243      & 1855 & 3634&601 &1324 \\ \hline
	0.1                                                             & 5201 & 0.69  & 4967     & 234      & 1752 & 3449&622 &1303 \\ \hline
	0.15                                                            & 4913 & 0.76  & 4697     & 216      & 1651 & 3262&733 &1192 \\ \hline
	0.2                                                             & 4625 & 0.83  & 4429     & 196      & 1550 & 3075&889 &1036 \\ \hline
	0.25                                                            & 4337 & 0.82  & 4159     & 178      & 1448 & 2889&885 &1040 \\ \hline
	0.3                                                             & 4049 & 0.84  & 3882     & 167      & 1337 & 2712&916 & 1009\\ \hline
	0.35                                                            & 3761 & 0.84  & 3611     & 150      & 1226 & 2535&918 & 1007\\ \hline
	0.4                                                             & 3473 & 0.86  & 3335     & 138      & 1146 & 2327&1007 & 918\\ \hline
	0.45                                                            & 3185 & 0.88  & 3059     & 126      & 1058 & 2127&1062 &863 \\ \hline
	0.5                                                             & 2897 & 0.91  & 2780     & 117      & 976  & 1921&1183 &742 \\ \hline
	0.55                                                            & 2608 & 0.94  & 2502     & 107      & 872  & 1737& 1274& 651\\ \hline
	0.6                                                             & 2321 & 0.96  & 2226     & 95       & 768  & 1553 &1393 &532 \\ \hline
	0.65                                                            & 2032 & 0.99  & 1948     & 85       & 674  & 1359 & 1631 &294 \\ \hline
	0.7                                                             & 1745 & 0.33  & 1671     & 74       & 585  & 1160& 195&1730 \\ \hline
\end{tabular}
	\caption{x901 Cell: Data Minimization Results per removal step.} \vspace{-5pt}
	\label{tab:x901_details}
\end{table}

Table~\ref{tab:x901_details} reports additional detail for data minimization in cell x901, in Fig.~\ref{fig:campus_dshap_detailed_removlowvalue_x901}, including: the removed fraction and number of training data, the recall $R_0$, number of measurements per users as well as the number of 0s and 1s of both the held-out data and the predicted $\widehat{y}$, per each step of the removal process. We make the following observations.
First, for the label B, where 65\% of the data have been removed and $R_0$ has peak at $0.99$, we notice that the predictor $\widehat{Q_c}(y)$ has predicted significant higher number of 0s than 1s ($1631$ 0s \vs  $294$ 1s). This does not surprise us, because, the predictor $\widehat{Q_c}(y)$ at label B is being trained with data points of higher quality for maximizing $R_0$. Essentially, in this scenario, data Shapley $\phi_i$ encodes the ability of the data to result in training predictors that would \emph{minimize} the false negatives (\ie maximize recall) and tend to over-predict 0s than 1s. Apparently, for a different metric the low/high $\phi_i$ points could be different. Second, when $R_0$ drops from $99\%$ to $33\%$ there is still data availability for both classes and users.

\begin{figure}
{	\centering
	\subfigure[Label A, Fig~\ref{fig:campus_dshap_detailed_removlowvalue_x901}.  \label{fig:campus_dshap_x901_sampling_labelA}] {\includegraphics[scale = 0.045]{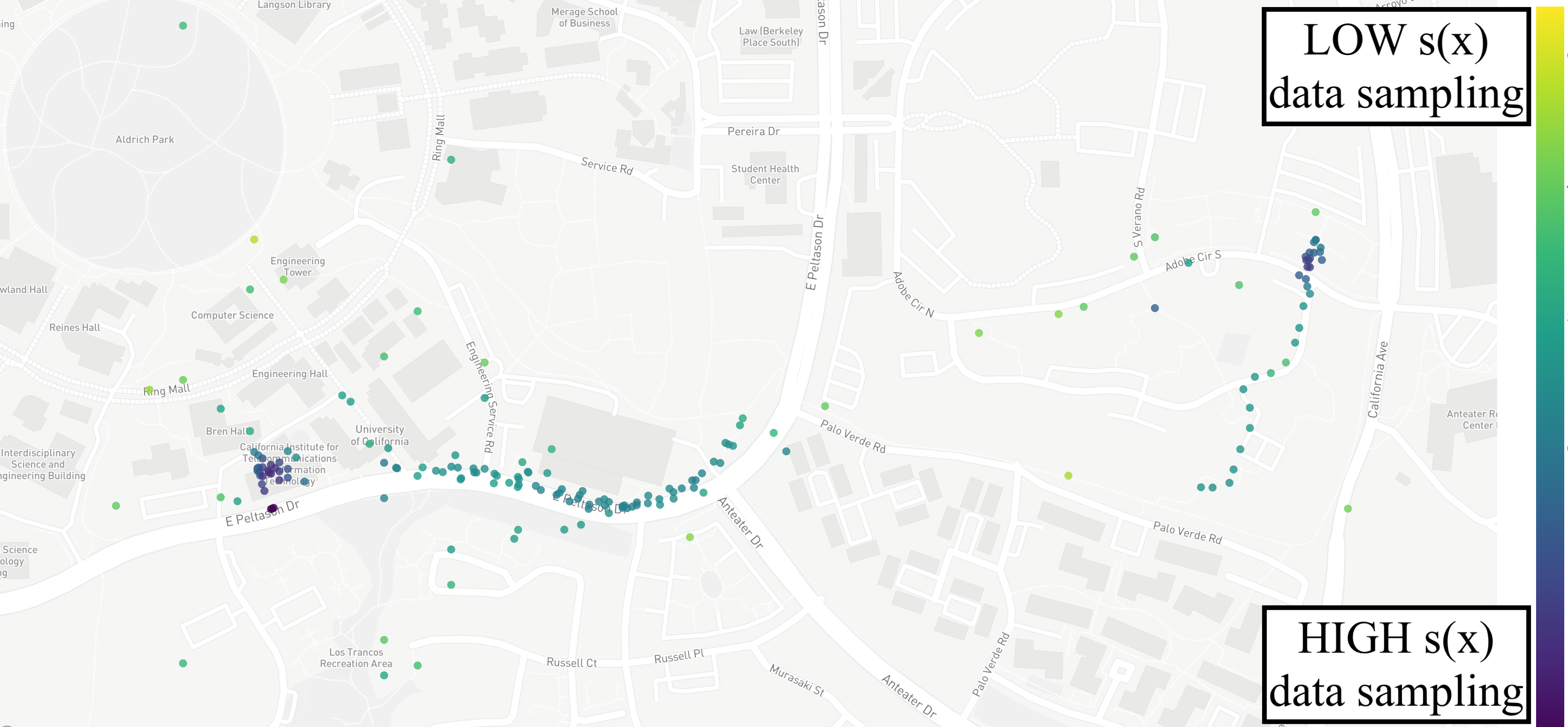}}
	\subfigure[Remaining Data's $s\text{\features}$ for performance at Label C, Fig~\ref{fig:campus_dshap_detailed_removlowvalue_x901}.  \label{fig:campus_dshap_x901_sampling_labelC}] {\includegraphics[scale = 0.09]{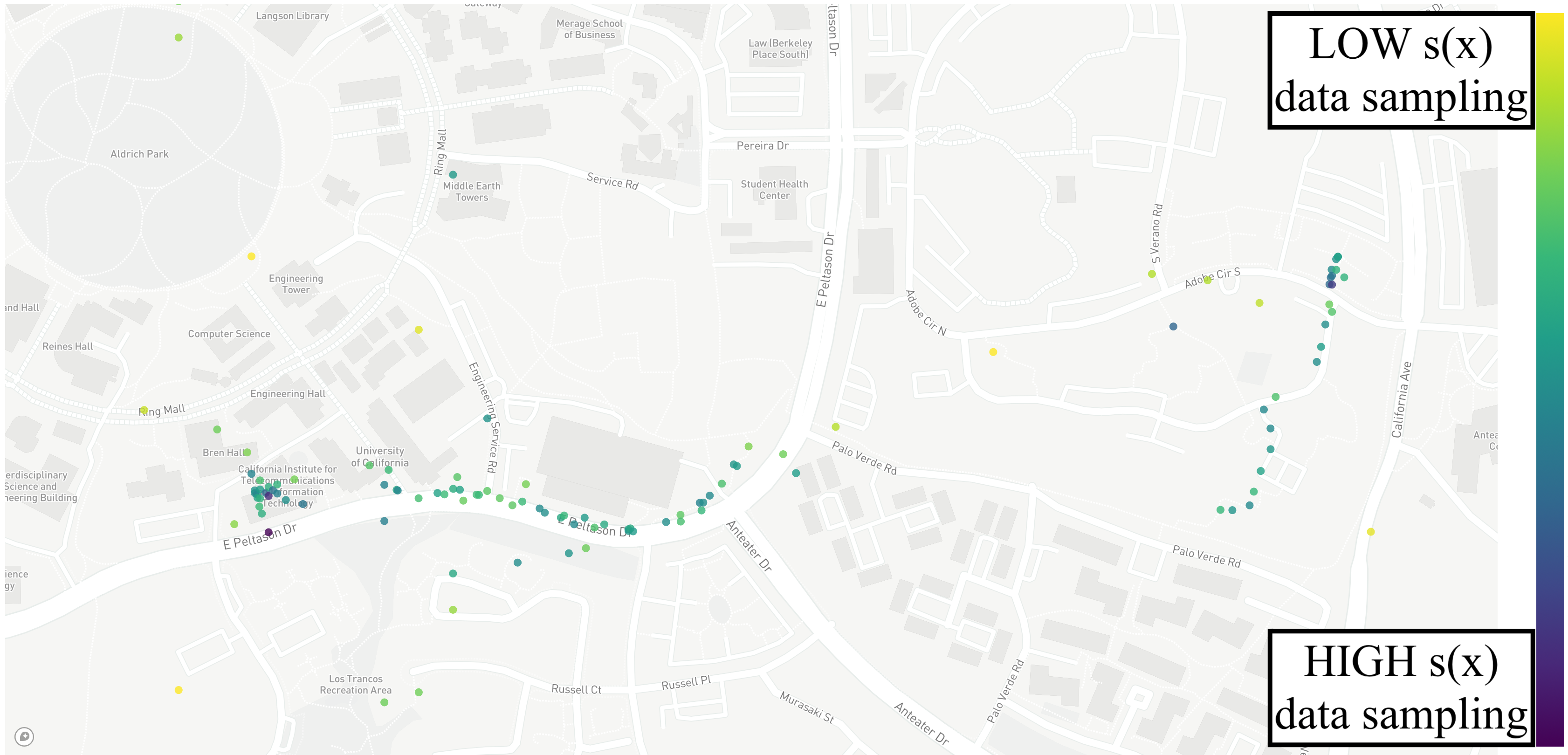}}
	\vspace{-10pt}
	\caption{\footnotesize  \inhousedataset cell x$901$. \textbf{Top}: Initial Sampling distribution $s(\text{\features})$ (Data for Label \textbf{A} in Fig.~\ref{fig:campus_dshap_detailed_removlowvalue_x901}). $\E[ \log s(\text{\features}) = -3.3 ] $  \textbf{Bottom}: Final Sampling distribution $s(\text{\features})$  (Data for Label \textbf{C} in Fig.~\ref{fig:campus_dshap_detailed_removlowvalue_x901}). The procedure of removing data points it eventually changed the sampling distribution of the data; at label A two regions were largely oversampled; at label C when the performance has finally been decreased the sampling distribution of the data look more uniform therefore it mismatches the original train distribution. $\E[ \log s(\text{\features}) = -9.3 ] $ \label{fig:campus_dshap_x901_sampling} }
}	
\end{figure}

\textbf{Dataset Shift and Data Shapley, for cell x901.} The dataset shift problem~\cite{datasetshiftNips:19} (\ie the mismatch of the training and the target distribution) for the labels ``A'' \vs ``C'' offers also significant insights of how the final performance is affected after a certain threshold of removing training data. 
Fig.~\ref{fig:campus_dshap_x901_sampling_labelA} shows $w_u \propto \frac{1}{s(\mathbf{l})}$ for the $\mathcal{D}_{train}$ data at label A; the home and work locations where data have been oversampled are illustrated clearly. The average data density is $\E[ \log s(\mathbf{l}) = -3.3 ] $. On the contrary, Fig.~\ref{fig:campus_dshap_x901_sampling_labelC} depicts $w_u \propto \frac{1}{s(\mathbf{l})}$ for the remaining $\mathcal{D}_{train}$ at label C and it can be clearly seen that the data distribution is closer to uniform and the average data density has been decreased to  $\E[ \log s(\mathbf{l}) = -9.3 ] $. The held-out data were randomly sampled from the original distribution, therefore, there is now a mismatch between the original and target distribution (in other words, the dataset shift problem we studied in Sec~\ref{sec:importancesampling}) which can explain the drop in the performance. Last but not least, the data that are being removed from label B $\rightarrow$ label C are primarily from the two oversampled regions, where we have a lot of held-out data to be tested.

\begin{figure}[t!]
	\centering
		\subfigure[\footnotesize Cell x$306$. \label{fig:campus_dshap_removlowvalue_x306}] {\includegraphics[scale = 0.27]{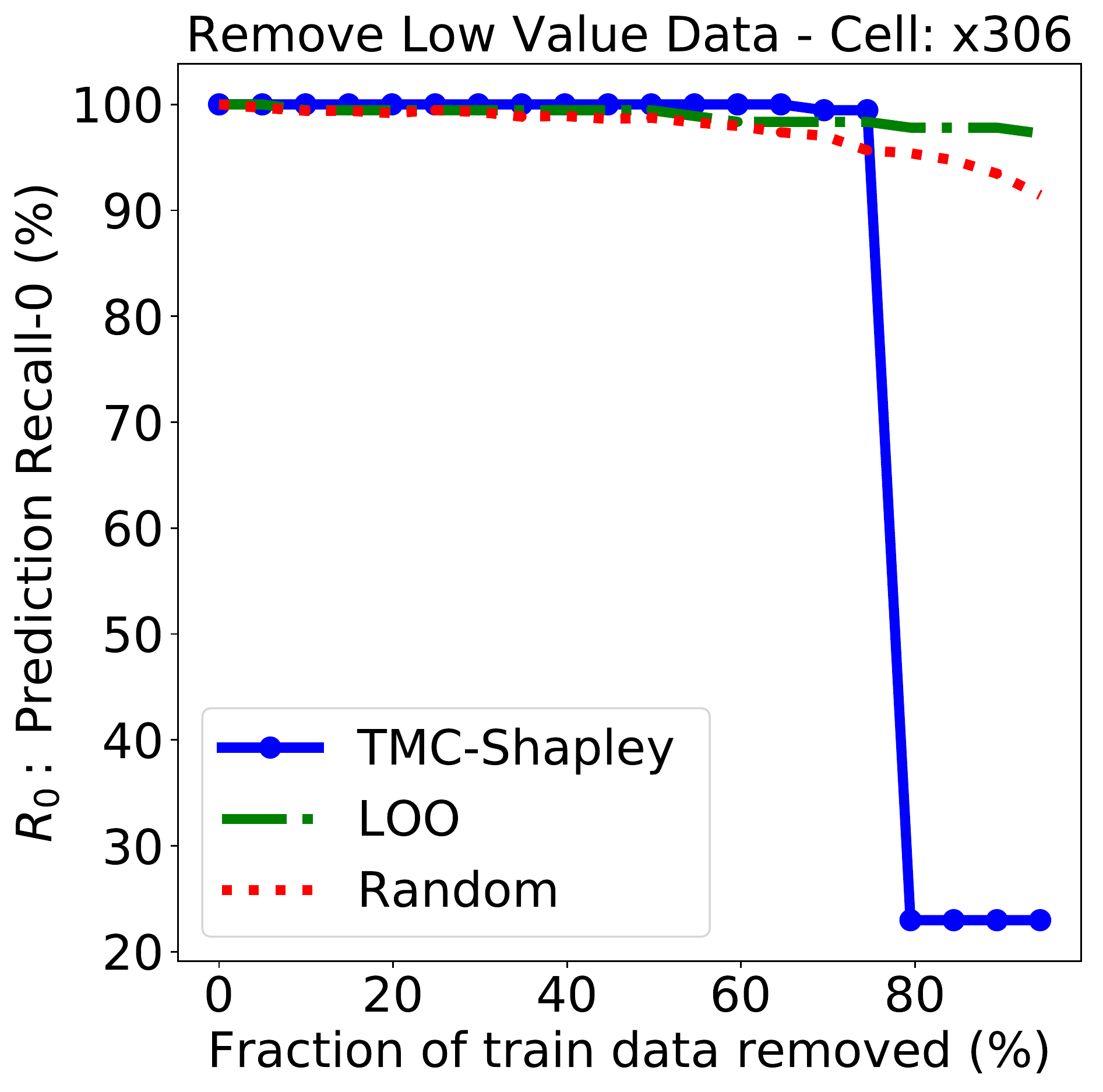}}
		\subfigure[\footnotesize Cell x$914$. \label{fig:campus_dshap_removlowvalue_x914}] {\includegraphics[scale = 0.27]{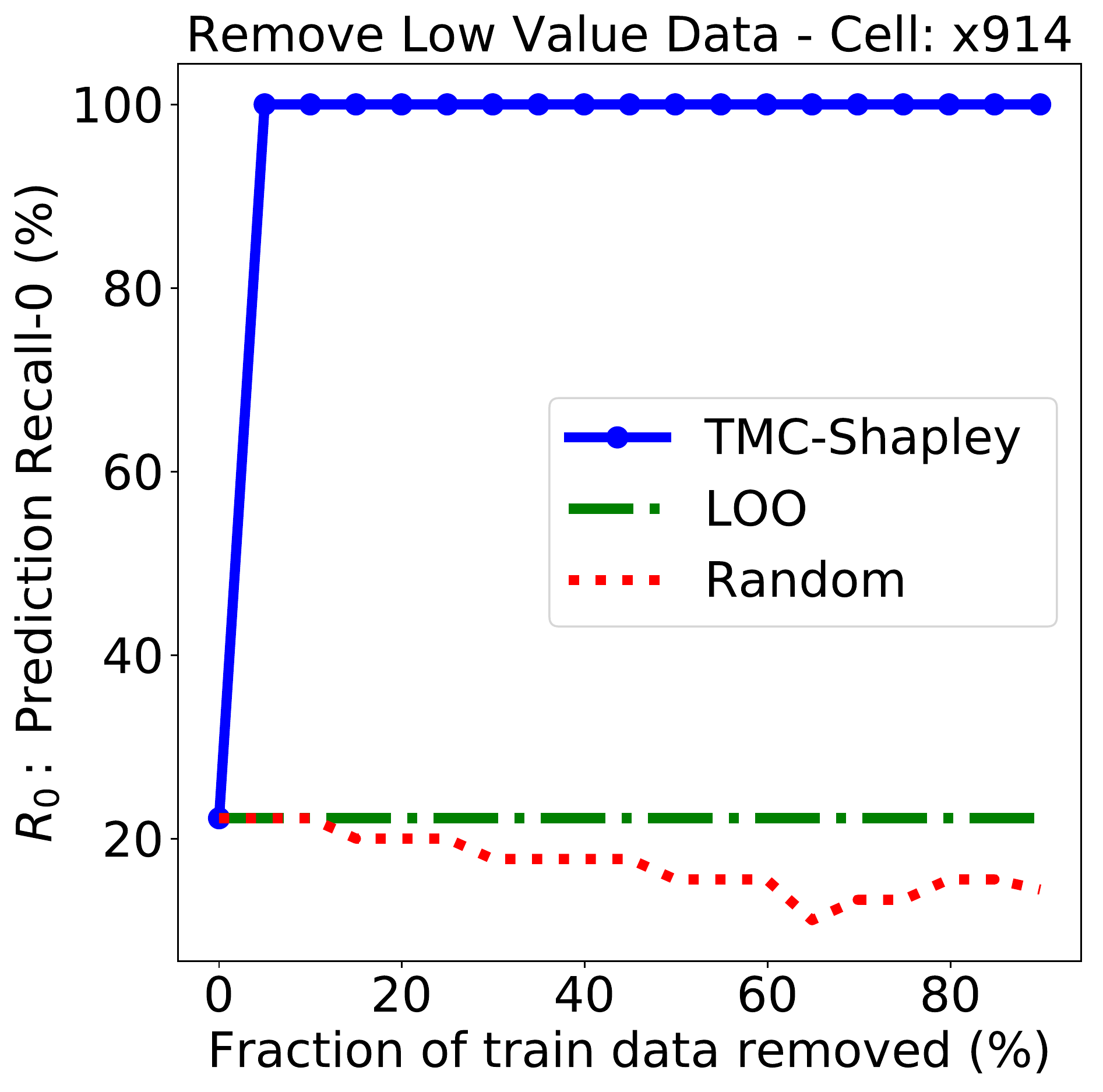}}
	\caption{\inhousedataset: Remove low valued data points (for Data-Shapley, LOO and Random) for various cells. Recall $R_0$ results. \label{fig:campus_dshap_removlowvalue_appendix} }
\end{figure}

Fig. \ref{fig:campus_dshap_removlowvalue_appendix} shows data minimization results for cells x306 and x914.

\subsubsection{Effect of Data Minimization on Metrics beyond $R_0$}
\begin{figure}[t!]
{\centering
	\subfigure[Accuracy $A$. \label{fig:campus_dshap_accuracy_x034}] {\includegraphics[scale = 0.27]{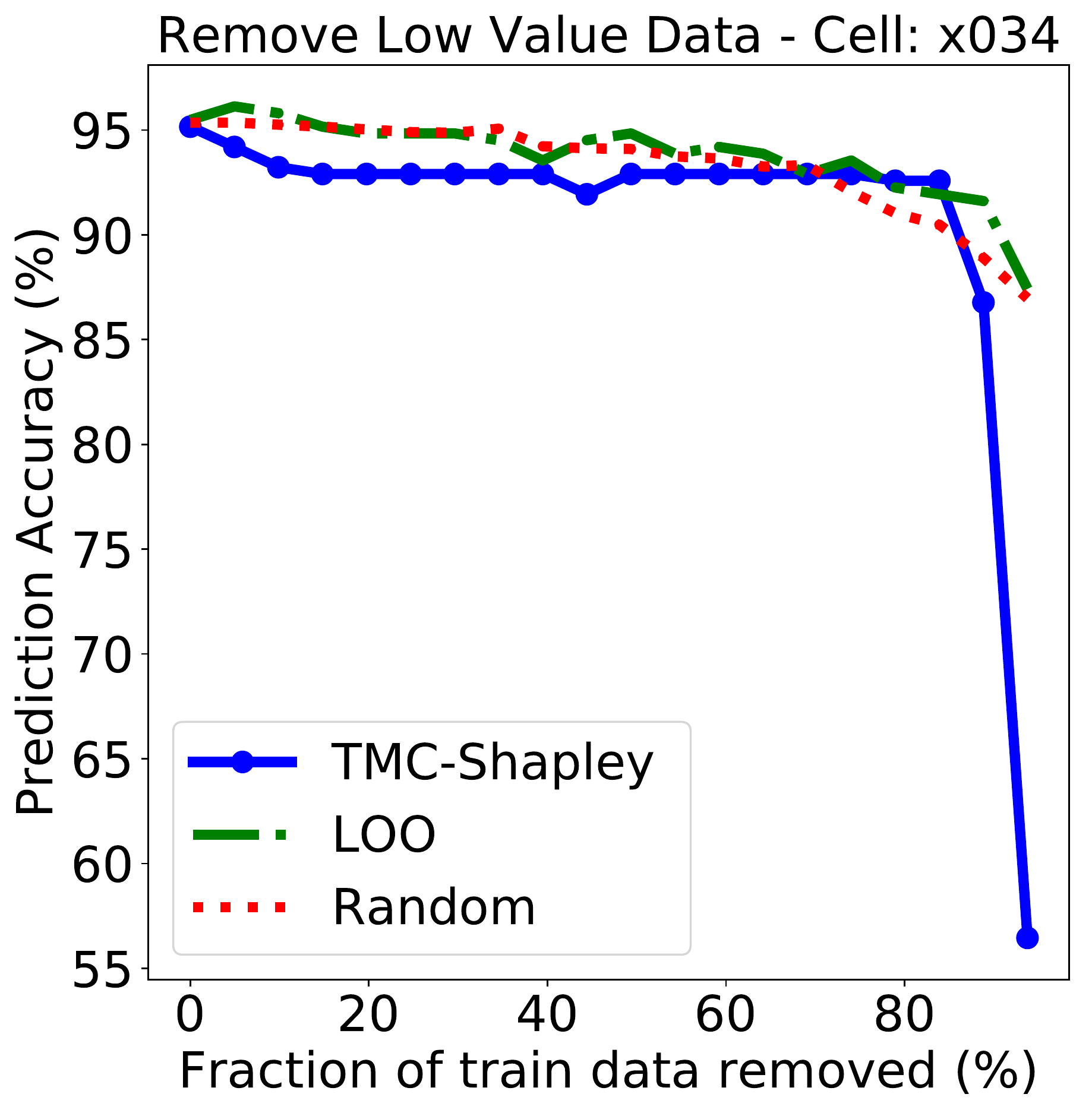}}
	\subfigure[Recall $R_0$. \label{fig:campus_dshap_recall0_x034}] {\includegraphics[scale = 0.27]{figs/dshap_remove_lowvalue/rm_leastDShapVal_cid_x034_metric_recall_0.pdf}}
	\vspace{-10pt}
	\caption{\small  \inhousedataset: Remove low valued data points (for Data-Shapley, LOO and Random) for  cell x$034$; \textbf{Left (a)} Results for Accuracy, \textbf{Right (b)} Results for Recall $R_0$ for coverage loss.\label{fig:campus_dshap_x034} }
}
\end{figure}

We also considered another cell x034 and the coverage classification task, but now for a different performance metric ($V$): accuracy ($A$). 
Fig.~\ref{fig:campus_dshap_accuracy_x034} reports results from the same data removal process as previously (\ie remove an increasing percentage of lowest valued datapoints). We observe that the TMC-Shapley's performance eventually outperforms LOO and Random when certain threshold of data removal has been reached. However after a certain threshold, the performance of TMC-Shap drops, as happened with the recall for the same cell (Fig.~\ref{fig:campus_dshap_recall0_x034}). That is expected because the portion of the data that can be removed depends on the dataset and the particular performance score/error metric, even for the same predictor. 

\subsubsection{Data Shapley with Reweigthed Error Metrics.}
\begin{figure}
	\centering
	\subfigure[$w_u \propto \frac{1}{s(\mathbf{l})}$. \label{fig:campus_dshap_map_x034_logwu}] {\includegraphics[width = 2in]{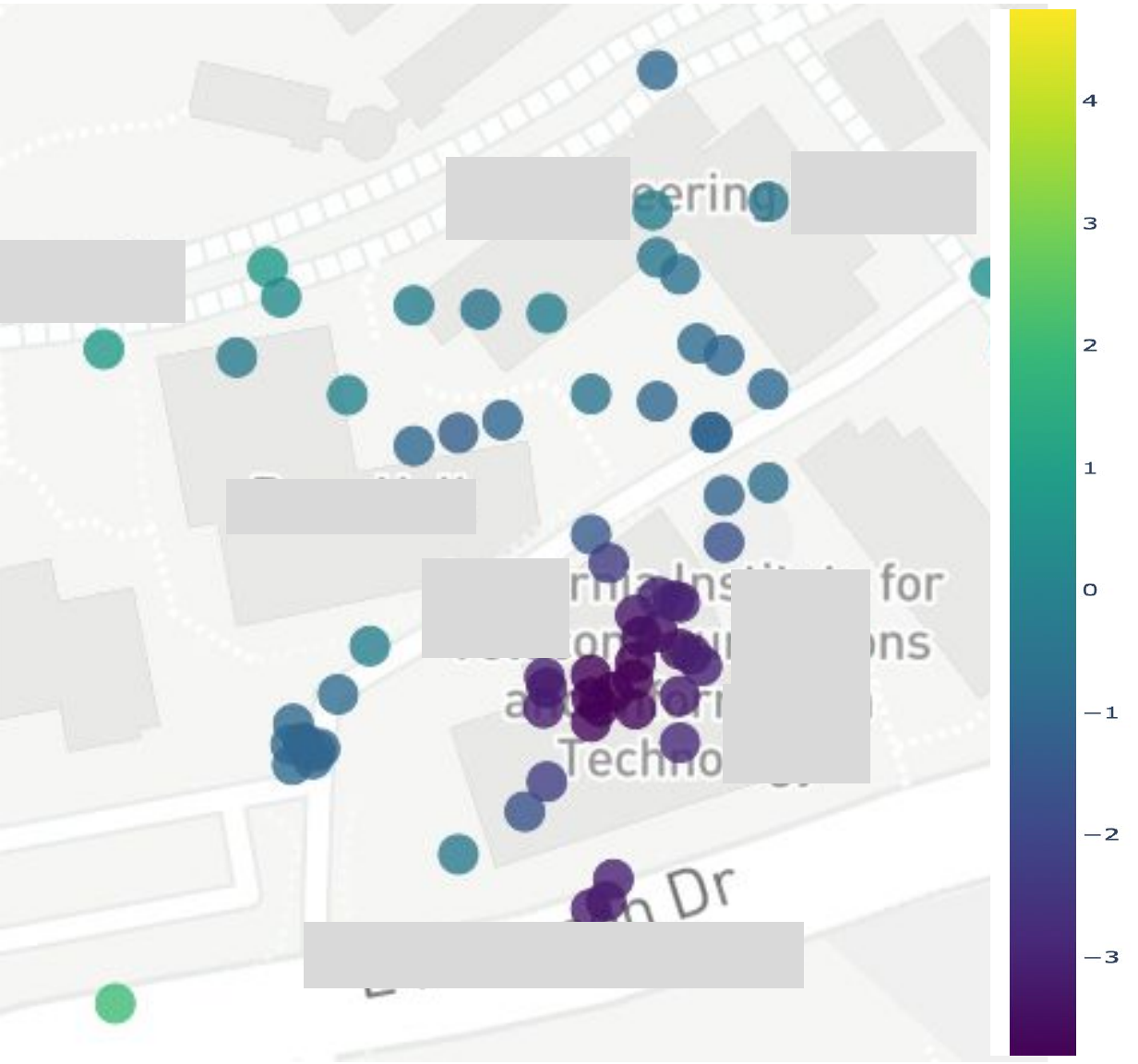}}
	\subfigure[Data Shapley $\phi_i$; $V \leftarrow$ MSE. \label{fig:campus_dshap_map_x034_mse}] {\includegraphics[width = 2in]{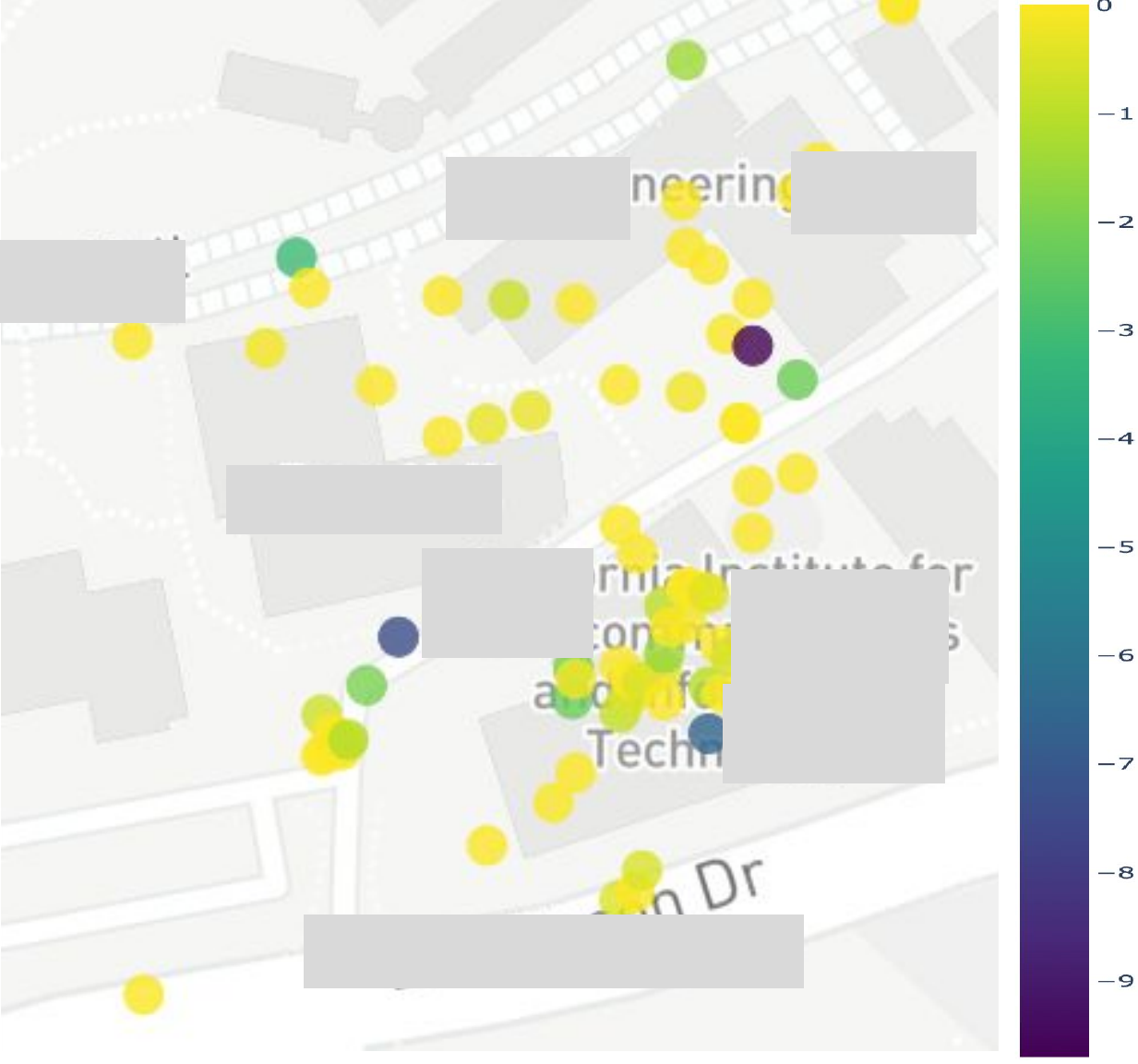}}
	\subfigure[Data Shapley $\phi_i$; $V \leftarrow \varepsilon_u$. \label{fig:campus_dshap_map_x034_rwgt}] {\includegraphics[width = 2in]{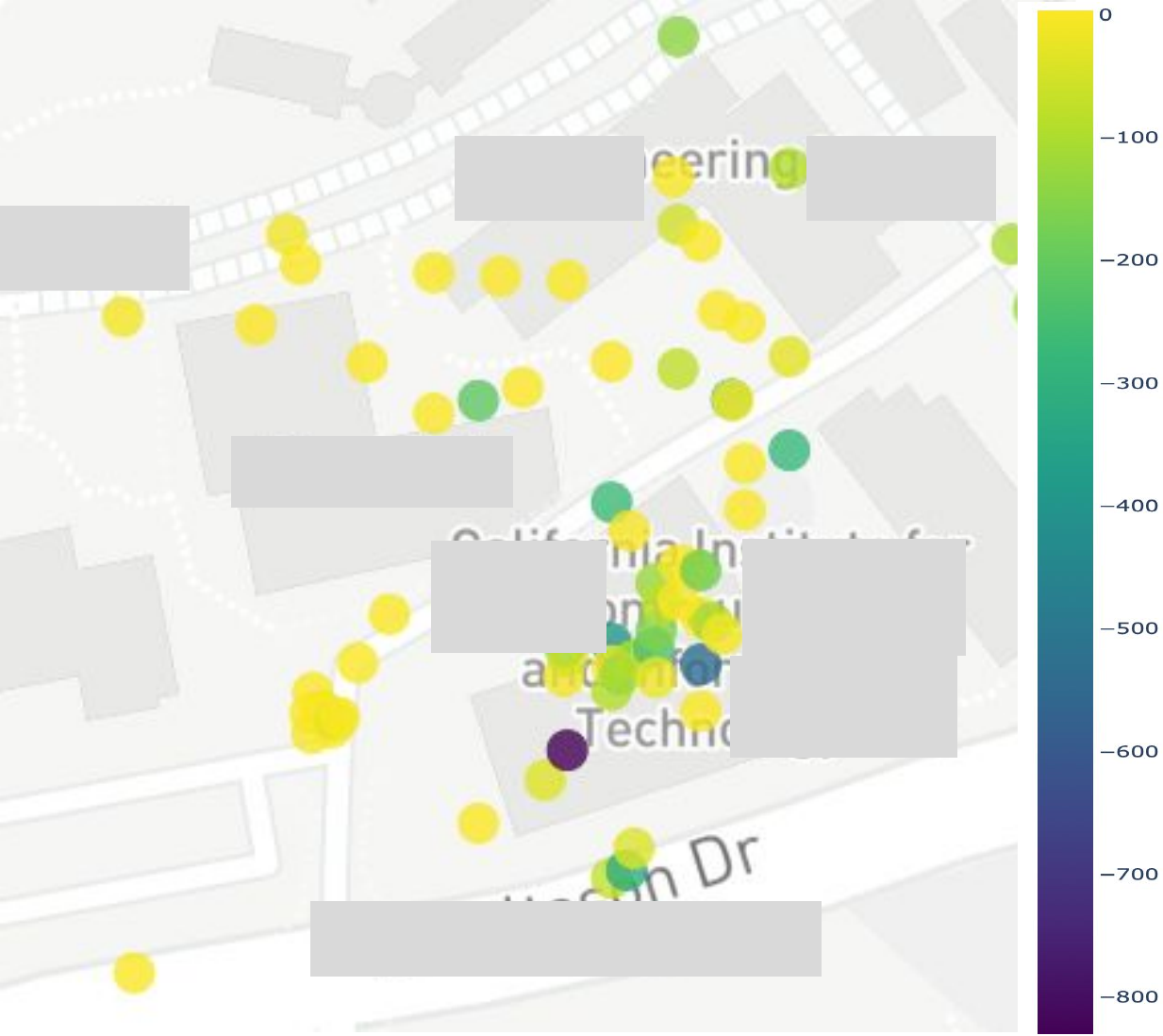}}
	\caption{\footnotesize  Example of how the reweighted error metric $\varepsilon_u$ affects the Data Shapley Values. (a) $w_u$ (\ie inversed data sampling) You can note the oversampled work location. (b) Data Shapley values for the Mean Squared Error valuation, \ie \problem{$I,k$}. (c) Data Shapley for the $\varepsilon_u$ performance score, \ie reweighted MSE or  \problem{$I,w_u$}. Please note how the oversampled areas (\ie with very low weights) at the home/work locations, have been assigned super small data shapley values. The performance score-evaluation function really matters.  \label{fig:campus_dshap_map_x034} }
\end{figure}

We further combined data Shapley and importance sampling, in order to provide data valuation for cellular operators objectives (\ie reweigthed error metrics). For this problem \problem{$I,w_u$}, we calculate the data Shapley values $\phi_i$ for the performance metric of $\varepsilon_u$ (uniform spatial error). Fig.~\ref{fig:campus_dshap_map_x034} shows a work location; in contrast to the \baseproblem~the over-sampled areas have a assigned a significant lower $\phi_i$ score because they do not contribute much to the maximization of the performance metric $\varepsilon_u$.  

\subsubsection{Summary}
We have proposed and demonstrated how to  compute  Shapley values of cellular measurements used for training in cellular performance predictors for any problem $P=(Q,W)$ in our framework. This enables us to identify  data points with low Shapley values, which should and can be removed for improving the prediction performance and data minimization, respectively.

\section{Conclusion and Discussion}\label{sec:conclusion}
We presented  a framework for predicting cellular performance from available measurements, which gives knobs to operators to express what they care most, \ie {\em what} performance metrics, in {\em what} regimes (via quality functions $Q$), and in {\em what locations} (via weights $W$). 
To that end, (1) we trained directly on the $Q$ instead of the RSRP domain; (2)  we used the  importance ratio re-weighting to address the mismatch between target and sampling distributions; and (3) we applied the data Shapley framework to assess the value $\phi$ of available measurements for the particular prediction task $P=(Q,W)$, which in turn enables data cleaning and minimization.
We evaluated these ideas on large, real-world LTE datasets and demonstrated their benefits.

Our contribution lies in the framework that deals with the measurements of unequal importance. It builds on top and is independent of the underlying prediction model; in this paper, we used our state-of-the-art Random Forests predictor\cite{alimpertis:19}, but this could be replaced by other regression (with square error loss) and classification predictors, if such become available.
To the best of our knowledge, this is the first signal maps prediction paper that goes beyond RSRP and MSE minimization, and the first to introduce the data Shapley framework idea in this context. Our focus has been exclusively on improving prediction, while adding privacy-enhancing mechanisms are directions for future work.

Another direction for future work is the applicability not only to LTE, but also to 5G deployment.
First, our framework can naturally handle prediction over small cells, \eg similarly to what we did in the \externaldataset, where prediction was not per cell, but across an area covered by multiple cells, (\ie \lteta) with \cid~used as a feature. 
Second, cellular operators have aggressively pushed the sub-6GHz deployments solution because of the mmWave practical limitations~\cite{5glimits} (\eg range of approx. 100s meters, only line-of-sight), and sub-6Ghz deployments share the same physical layer and network characteristics with the LTE networks. 
 Third, sampling biases will be amplified with small cells deployment and our re-weighting schema could help mitigate it. Last but not least, our weight functions $W(\mathbf{x})$  can express complex 5G operator objectives (see~\ref{sec:reweightedimportancesampling}) (\eg IoT, small cells \etcfree).

\section*{Acknowledgments}\label{sec:acknowledgements}
This work has been supported by NSF Awards 1956393,  1939237, 1900654, and 1526736. 
We are grateful to Tutela for sharing their city-wide datasets, to former students of our group that participated to the UCI study, and Justin Ley for discussions on the DNN models.

\newpage

\bibliographystyle{ACM-Reference-Format}
\bibliography{rssprediction.bib}

\vfill
\newpage
\appendix

\section*{Appendix A1. Data Shapley Formulation -  Details \label{appendix-shapley}}

This appendix extends Section \ref{sec:shapley-theory} with with details on the definition of Data Shapley.

Data Shapley is a framework developed in~\cite{dshapIcml:2019}, which attempts to provide an answer to the question: ``How do we quantify the value of the data in an algorithmic predictions and decisions?'' Data Shapley can provide us a valuation of the data (\ie assign an arithmetic value to each data point) in the setting of supervised ML.  What is an equitable measure of the value of each train data point (\aka datum) $(\text{\features}, y_i )$ to the training algorithm $\mathcal{A}$? In order to answer that, we have to take a closer look to the essential ingredients of a supervised ML algorithm: (a) training data, (b) learning algorithm, (c) performance metric. The prediction is a function that \emph{depends jointly} on all of them, therefore, each one of them affects the equitable measure assigned to our data. We follow the exposition and the organization of~\cite{dshapIcml:2019}; the notation of the aforementioned components is as follows:

{\em (a) Data: } The dataset of the ML setting follows the typical setup we have already seen in this paper: $\mathcal{D} = \{ (\text{\features}_i, y_i) \}^N_1$. We denote with $\mathcal{D}_{train}$ and $\mathcal{D}_{test}$ the training and the test data respectively; in the Data Shapley framework we also need $\mathcal{D}_{eval}$ for the final evaluation, \ie the heldout data.

{\em (b) Learning Algorithm $\mathcal{A}$: } A \emph{black box} for the data Shapley setting, which takes as input $\mathcal{D}_{train}$ and produces as ouput a predictor $\widehat{y} = \widehat{f}_y(\mathbf{x})$. For example, the algorithm could be a logistic regression or the \RFs~predictors we used in this paper.

{\em (C) Performance Score $V$: } We can treat it as a black box that takes an input $\widehat{f}$, the error metric - valuation we want to apply to each data point, the test data   $\mathcal{D}_{test}$  and outputs a performance score $V$. We denote with  $V(S,\mathcal{A}) = V(S)$ the performance score of a predictor trained on train data $S$ using the learning algorithm $\mathcal{A}$. Please note, that the performance score can be completely different than the loss function of the learning algorithm itself. For example, we can train \RFs~ regression with the typical $MSE$ minimization and then report also $RMSE$ as we did in Chapter 4, however, we could train a model on a different loss function than the evaluation functions as we did in some cases in Chapter~\ref{chapter_qosweights}. Data Shapley provides us a powerful framework for evaluating the different quality functions and weights functions (which essentially  are other forms of error metrics according to a target distribution of interest).

The {\em goal} is to compute the data Shapley value $\phi_i(\mathcal{D}, \mathcal{A}, V)  \equiv \phi_i(V) = \phi_i \in \mathbb{R}~\forall i, (\text{\features}_i,y_i) \in \mathcal{D}_{train}$, which follows the equitable valuation properties described next.

{\bf Equitable Valuation Conditions}\label{sec:dshap_equitableproperties}
It is desirable that the data Shapley valuation meets the following equitable valuation conditions:  ~\cite{dshapIcml:2019}:
\begin{enumerate}
	\item \emph{Null Property.} If a datum $(\text{\features}_i, y_i)$ does not change the performance it should be given $\phi_i = 0$. Thus, $\forall S \subseteq \mathcal{D} - \{i\}, V(S) = V(S \cup \{ i\})$ 
	\item \emph{Symmetry Property.} If two distinct datums $i,j$ contribute equally to the performance score $V$ then they should have equal data Shapley values. In other words, $\forall i,j \in \mathcal{D}$ such as $S\subseteq \mathcal{D} - \{ i,j\}$  and  $V(S \cup \{ i\}) = V(S \cup \{ j\}) \Rightarrow \phi_i = \phi_j$.
	\item \emph{Summation and linearity.} It is very typical in Machine Learning to have  an overall performance score which is the sum of seperate performance scores. A typical example is the mean squared error (MSE) that is the summation of the weighted equally (\ie the mean) squared losses of individual points. When $V$ consists of the summation of individual scores, then, the overall value of a datum should be the sum of its values for each score. In other words, for data Shapley we should have: $\phi_i (V+W) = \phi_i(V) + \phi_i(W) $ for performance scores $V,W$. It should be re-emphasized, that the data Shapley value is defined for the train data points according to the performance scores on the test data. Thus, $V = - \sum_{k \in test} l_k$ with $l_k$ to be the predictor's loss on the $k$th test point; the data Shapley value for quantifying the value of the $i$th source for predicting the $k$th test point is denoted with $\phi_i(V_k)$.  If datum $i$ contributes values $\phi_i(V1)$ and $\phi_i(V_2)$ to the predictors of the test points 1 and 2 respectively, then we expect the data Shapley value of $i$ in predicting both test points, \ie when $V= V_1 +V_2$, to be $\phi_i(V_1) + \phi_i(V_2)$.
\end{enumerate}

\textbf{Closed Formula and Approximation.} The data Shapley value which complies to the above three properties, must have the following closed form \cite{dshapIcml:2019}:
\begin{align}
\phi_i = C \sum_{\mathcal{D} - \{ i \}} \frac{ V(S \cup \{ i \})- V(S)}{ \binom{n-1}{|S|} } \label{eq:datashapleyform}
\end{align}
Intuitively, a data point interacts and influences the training procedure, which creates the predictor function, in conjunction with the other training points. Thus, conditions which formulate the interactions among the data points and how an holistic data valuation is generated should be taken into account. The data Shapley value in Eq. (\ref{eq:datashapleyform})  is essentially the average of the leave-one-out value (\aka marginal contributions) of all possible training subsets of data in $S$. It should be noted that an exhaustive computation of Eq.~(\ref{eq:datashapleyform}) is very expensive computationally. An approximation - truncated Monte Carlo algorithm (TMC-Shapley) is provided by~\cite{dshapIcml:2019} and is being used in this paper.

\textbf{Leave-one-out (LOO) Baseline.\label{sec:dshap_loodefinition}} A simpler method  is ``leave-one-out'' (or LOO), which calculates the datum value by leaving it out and calculating the performance score, \ie $\phi_i^{\text{LOO}} = V(D) - V(D - \{i\})$. However, LOO does not consider all possible subsets a data point may belong and does not satisfy the equitable conditions.

\section*{Appendix A2. Choice of Base Predictor }

This appendix extends Sections \ref{sec:rfsprediction} and \ref{sec:numerical_results_baseproblem} (\eg it elaborates on the comparison in Table \ref{tab:campus_allcells}), and provides additional evaluation results to justify the choice of \RFs~  as a good predictor for the base model $P=(I,k)$, underlying our framework. In our short paper \cite{alimpertis:19}, we had compared \RFs
~to model-based (\LDPL) and spatiotemporal predictors (OK, OKD). In this appendix, we added a comparison to state-of-the-art  DNN and CNN baselines, which have appeared after \cite{alimpertis:19}.   However, this is background information for completeness, and not the core contribution of this paper.

\subsection*{A2.1 Random Forests \label{appendix:RFS}}

We developed the Random Forests predictor in ~\cite{alimpertis:19, alimpertis:2020}, and we showed that it outperforms state-of-the-art predictors. In this paper, we use it as the base "workshorse" predictor.
We describe the design of the \RFs~predictor in Section \ref{sec:rfsprediction}, and we compare its performance to other candidates baseline predictors (including model-based, geospatial interpolation, and DNNs) in Section \ref{sec:numerical_results_baseproblem} and in this appendix. This is for completeness so that we justify that it is a good choice for base predictor. However, we emphasize that it could be replaced in our framework by any other predictor that minimizes MSE, without affecting the contributions of our framework, which build on top of any base predictor. \RFs~is {\em not} the contribution of this paper, thus its performance evaluation is mostly in this appendix (except for the summary in Table \ref{tab:campus_allcells}, which is in the main paper).

\subsection*{A2.2 Other Baseline Predictors}

In this section, we present representative state-of-the-art prediction methods, which are  used as baselines in this paper. Table~\ref{tab:table-compare-approaches} summarizes the family of predictors that can be used to generate \maps~as well as put our proposed work into perspective. There is a large literature on propagation models~\cite{winnerreport,raytracing:15,cost231:99} and they model the received signal strength given the location of receiver, transmitter and the propagation environment. As a representative baseline from the family of model-based predictors, we consider the  \textvtt{Log Distance Path Loss (LDPL)} propagation model, which is simple yet widely adopted in the literature. Additionally, it provides further understanding of the (i) low-level RSS statistical properties and (ii) the wireless network fundamentals that should be taken into account for the prediction task.

\subsubsection*{A2.2.1 Model Based Prediction (LDPL)\label{sec:LDPLmodel}}
The \textvtt{Log Distance Path Loss (LDPL)}~model predicts the power (in dBm) at location $\mathbf{l}_j$ at distance $||\mathbf{l}_\text{BS}-\mathbf{l}_j ||_2$ from the transmitting basestation (BS) or cell tower, as a log-normal random variable (\ie normal in dBm)~\cite{Rappaport:2001:WCP:559977, alimpertis:14}:
\begin{align}
y_{\text{\cid}}^{(t)}\left(  \mathbf{l}_j\right) = P_0^{(t)}- 10 n_{j} \log_{10} \left(\frac{||\mathbf{l}_\text{BS}-\mathbf{l}_j ||_2}{d_{0}}\right) + \omega_j^{(t)}.
\label{eq:LDPLmodel}
\end{align}
The most important parameter is $n_j$, \ie the  path loss exponent (PLE), which has typical values between 2 and 6. $P_0^{(t)}$ is the received power at reference distance $d_0$,  which is calculated by using the free-space path loss (Friis) transmission equation for the corresponding downlink frequency, gain and antenna directionality, and $\mathbf{l}_\text{BS}$ the location of the transmitting antenna. In its simplest form, the equation assumes antenna reception gain and base station antenna gain equal to $0$ dBi, but the application of an antenna directionality gain model as well as mobile gain model is also possible as shown in~\cite{alimpertis:14}. The log-normal shadowing is modeled by $\omega_j^{(t)} \sim \mathcal{N} (0,\sigma^2_{j}(t) )$ (in dB), with variance $\sigma^2_{j}(t)$ assumed independent across different locations. The cell  (identified by cell ID \cid) affects several parameters in Eq. \ref{eq:LDPLmodel}, including $P_0, \omega_j$, the locations of transmitting ($\mathbf{l}_\text{BS}$) and receiving ($\mathbf{l}_j$) antennas. The simplicity of this model lies in that it has only one  parameter (the path loss exponent $n_j$) to be estimated from the measurements. Prior work~\cite{alimpertis:14} has shown that the PLE values and the time variant RSS variance ($\sigma^2_{j} (t)$) can be estimated by a large number of collected measurements. It should be noted, that in real world setups, base stations' transmission power changes according to the network load and conditions~\cite{alimpertis:14}, contributing to the time varying component of the equation. We consider two cases. 

\textvtt{Homogeneous LDPL:} Much of the literature assumes that PLE $n_j$ is the same across all locations. We can estimate it from Eq.~(\ref{eq:LDPLmodel}) from all the training data points.

\textvtt{Heterogeneous \LDPLknn:} Recent work (\eg~\cite{specsense:17, alimpertis:14}) has considered that PLE changes across different locations. We considered various ways to partition the area into regions with different PLEs, and we present the one where we estimate $\widehat{n_j}$ via $knn$  regression, from the $k$ nearest neighbors, weighted according to their Euclidean distance (refer to as ``\LDPLknn'').

{\bf Baseline \LDPL~ Setup.}   The baseline \LDPL, whose performance is reported in Table \ref{tab:campus_allcells}, refers to the following parameters of Eq.~(\ref{eq:LDPLmodel}): we compute the distance from the base station using the online database from \url{opencellid.org};  breaking distance $d_0=1m$~\cite{alimpertis:14}; \freq~is obtained from the $EARFCN$ measurement readily available via the Android API. \footnote{For the \inhousedataset we got a limited number of $EARFCN$ measurements which indicated the most utilized frequencies in the area.}. In addition, for \LDPLknn: we select empirically $k=100$ neighbors for the \inhousedataset and $k=10\%$ of the training data points in each cell for the \externalcitiesdataset. 

\subsubsection*{A2.2.2 Geospatial Interpolation (OK-OKD)\label{sec:geostatisticsmodels}}
State-of-the-art approaches in data-driven RSS prediction ~\cite{specsense:17, molinari:15, phillips:12} have primarily relied on geospatial interpolation (\aka geostatistics), as presented earlier in Sec.~\ref{sec:relatedwork}. However, this approach is inherently limited to predicting RSS from spatial features $(l^x, l^y)$ and does not naturally extend to additional dimensions and contextual information. In this section, we present the best representatives of this family of predictors, namely ordinary kriging (OK)~\cite{molinari:15} and its variants OK detrending (OKD)~\cite{specsense:17}, which are used as baselines for comparison in this paper. 

\textvtt{Ordinary Kriging (OK):} It predicts RSS at the testing location $\mathbf{l}_j = (l_j^x, l_j^y)$ as a weighted average of the $K$ nearest measurements in the training set: $y_{j} = \sum ^{K}_{i=1}\lambda_{i}y_{i}$. The weights $\lambda_i$ are estimated by solving a system of linear equations that correlate the test with the training data via the semivariogram function  $\gamma(h)$, which defines the variance between two different data points. Semivariogram $\gamma(h)$  must be estimated by the training data for different values of the lag $h$ (\ie the distance between the data points) and each different environment requires different $\gamma(h)$; \ie distinct $\gamma(h)$ for \inhouse and \externalcitiesdataset. The solution of the system is given by a Lagrange multiplier; more details for the derivation can be found in~\cite{specsense:17}.

\textvtt{Ordinary Kriging Partitioning (OKP):} In~\cite{specsense:17}, Voro\-noi-based partitioning was used to identify regions with the same PLE and apply a different OK model in each region. This is comparable to the heterogeneous propagation model.  However, OKP solves separately a different optimization problem for each local subregion, which make them impractical for city wide signal maps.

\textvtt{Ordinary Kriging Detrending (OKD)~\cite{specsense:17}:} While OK  typically assumes the same mean value across locations\footnote{The first applications of geostatistics were interpolation for environmental measurements like humidity and temperature.}, this does not necessarily hold for RSS values. OKD incorporates a simplified version of \LDPL~in the prediction in order to address this issue \cite{specsense:17}. This can be thought as a hybrid approach of data-driven (geospatial) and model-driven (\LDPL). It is the best representative of the geospatial predictors and serves as our baseline for comparison.

The basic steps of OKD are as follows: (i) OKD needs a  transmitter location (we can use the strongest RSS algorirthm~\cite{yang:10} or the \textvtt{OpenCellID} online DB) and computes the perceived path-loss exponent $n_i$ at each training data point ($n_i = P_0 - \text{\rsrp}_i / \log_{10} d_i$ , where $d_i = ||\mathbf{l}_\text{BS}-\mathbf{l}_i ||_2$). (ii) After computing the mean PLE $\widehat{n}$, OKD computes $L(y_{i}) = y_{i} - 10 \widehat{n} \log_{10} (d_i) $ and the detrending component  $\delta_i = y_i - L(y^P_i) $ for the training data points. (iii) Finally, OKD applies OK to predict  $\delta_j$ by learning with the data $\delta_i$  and predicts $\delta_j$.

 {\bf Baseline Geostatistics Setup.} For the evaluation of geostatistics predictors reported in Table \ref{tab:campus_allcells}, we choose the following configuration.  The number of neighbors  was empirically set to $k=10$. For geospatial interpolation methods, a larger $k$ did not show any significant improvement, and it would result in much higher computational cost. An exponential fitting function of the semivariogram function $\gamma(h)$ was selected~\cite{specsense:17}; the maximum lag ($h$) was set to $200$m, as in~\cite{specsense:17}, for the \inhouse and \external environments, while it was set to $600$m for the \externalsuburban suburban environment. The approximated empirical semivariogram $\widehat{\gamma(h)}$ was calculated per $10m$~\cite{specsense:17}.

\subsubsection*{A2.2.3 Deep Neural Networks (DNNs)}

In our effort to pick a state-of-the-art baseline model, we considered and evaluated the following approaches.

Recent work in~\cite{raik:18} uses DNNs along with detailed 3D maps from Light and Range Detection Data (Li-DAR) and spatial coordinates $\mathbf{l}_j = (l_j^x, l_j^y)$ as features for predicting signal strength (LTE RSRP) values. In our paper, we compare  our \RFs~predictors with a DNN predictor. It should be noted that we only consider spatial coordinates since they are not Li-DAR readily available for our datasets. Most importantly, Li-DAR data are expensive and difficult to obtain. Our paper focuses on developing predictors with readily available data from the mobile devices provided by Android APIs. 

 We also did a straw man evaluation of a state-of-the-art CNN approach \cite{cnnUCLA}. We usied a simplified implementation based on the source code kindly provided by the authors of \cite{cnnUCLA}, and tried two datasets: (i) the same synthetically generated data as in \cite{cnnUCLA}  and (ii) on own real-world UCI campus data. Specifically, we used the synthetic data to construct ground truth masks that matched the UCI campus masks and then trained the CNNs with the original masks and our real-world sparse masks. The second option resulting to much lower performance due to data sparsity in our case. 
 
 {\bf DNN Baseline Setup.} The baseline ``DNN'', whose performance is reported in Table \ref{tab:campus_allcells}, refers to the following configuration. We tuned the multilayer perceptron deep neural network (DNN) architecture for each cell tower via the Keras Hyperband Tuner \cite{omalley2019kerastuner, JMLR:v18:16-558} which is an optimized version of random search. Specifically, we tuned the width (1024 nodes max.) and depth (4 layers max.) of each DNN, the activation functions, dropout percentage (0.25 max.) and used the Xavier normal weight initializer \cite{xavier-init}. We also added batch normalization which did not decrease the RMSE. After tuning the architecture, we re-train and test each cell DNN on five independent random splits of data and report the average performance.

\subsection*{A2.4 Baseline Models Comparison}
\label{sec:appendeix_baseproblem_results}

\begin{figure}[t!]
	\centering
	{\includegraphics[scale=0.21]{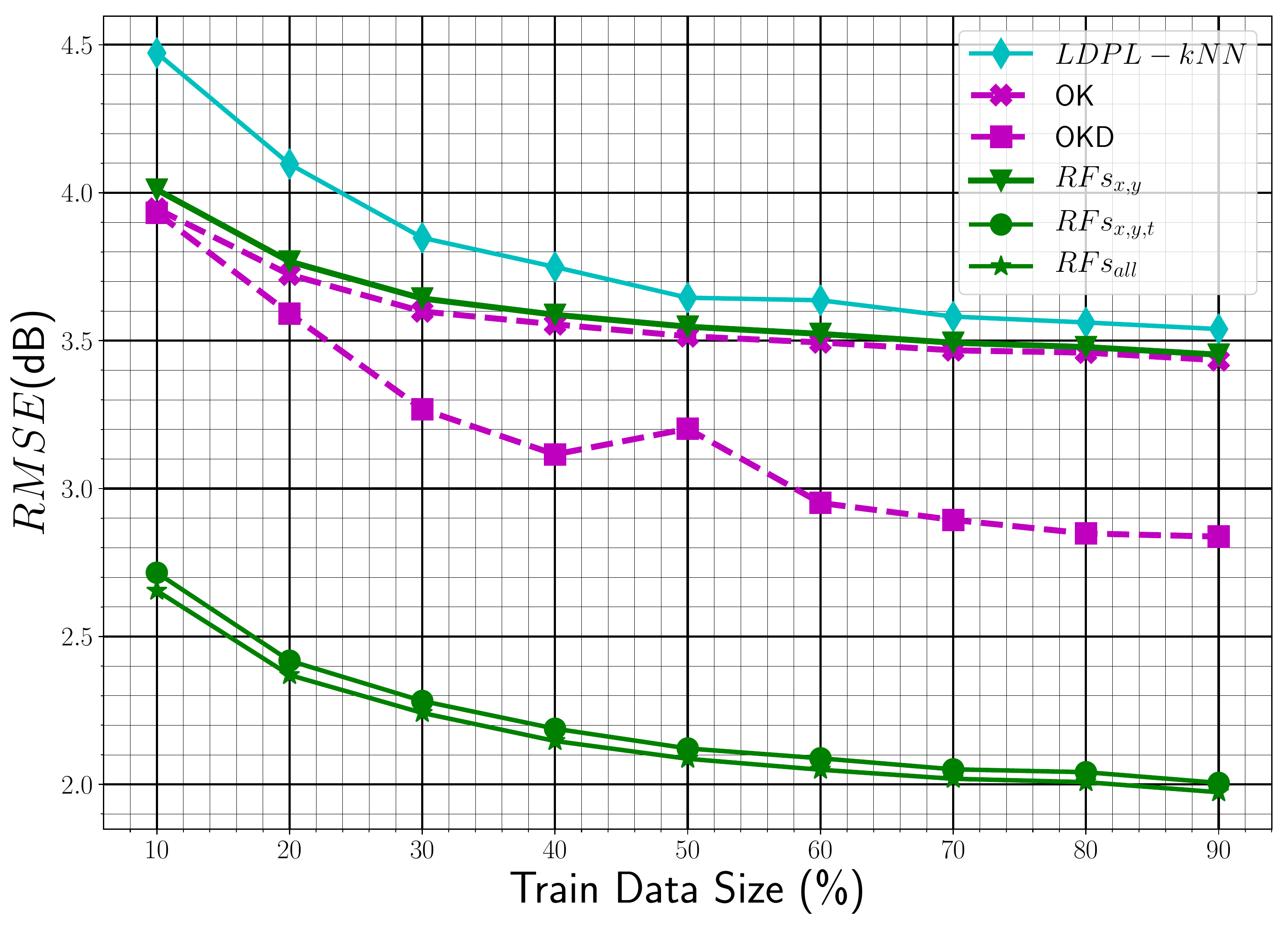}}
	\vspace{-10pt}
	\caption{\footnotesize \inhousedataset : $RMSE$ vs. Training Size. Our methodology (\RFs~with more than spatial features, \ie \RFsSpatiotemporal,~\RFsAll) significantly  improves the RMSE-cost tradeoff: it can reduce $RMSE$ by $17\%$ for the same number of measurements compared to state-of-the-art data-driven predictors(OKD);  \emph{or} it can achieve the lowest error possible by OKD ($\simeq 2.8$dB) with 10\% instead of 90\%  (and 80\% reduction) of the measurements. \label{fig:rmse_vs_trainsize_campus}}
	\vspace{-10pt} 
\end{figure}

\begin{figure}[t!]
	\centering
	\subfigure{\includegraphics[width=0.6\linewidth]{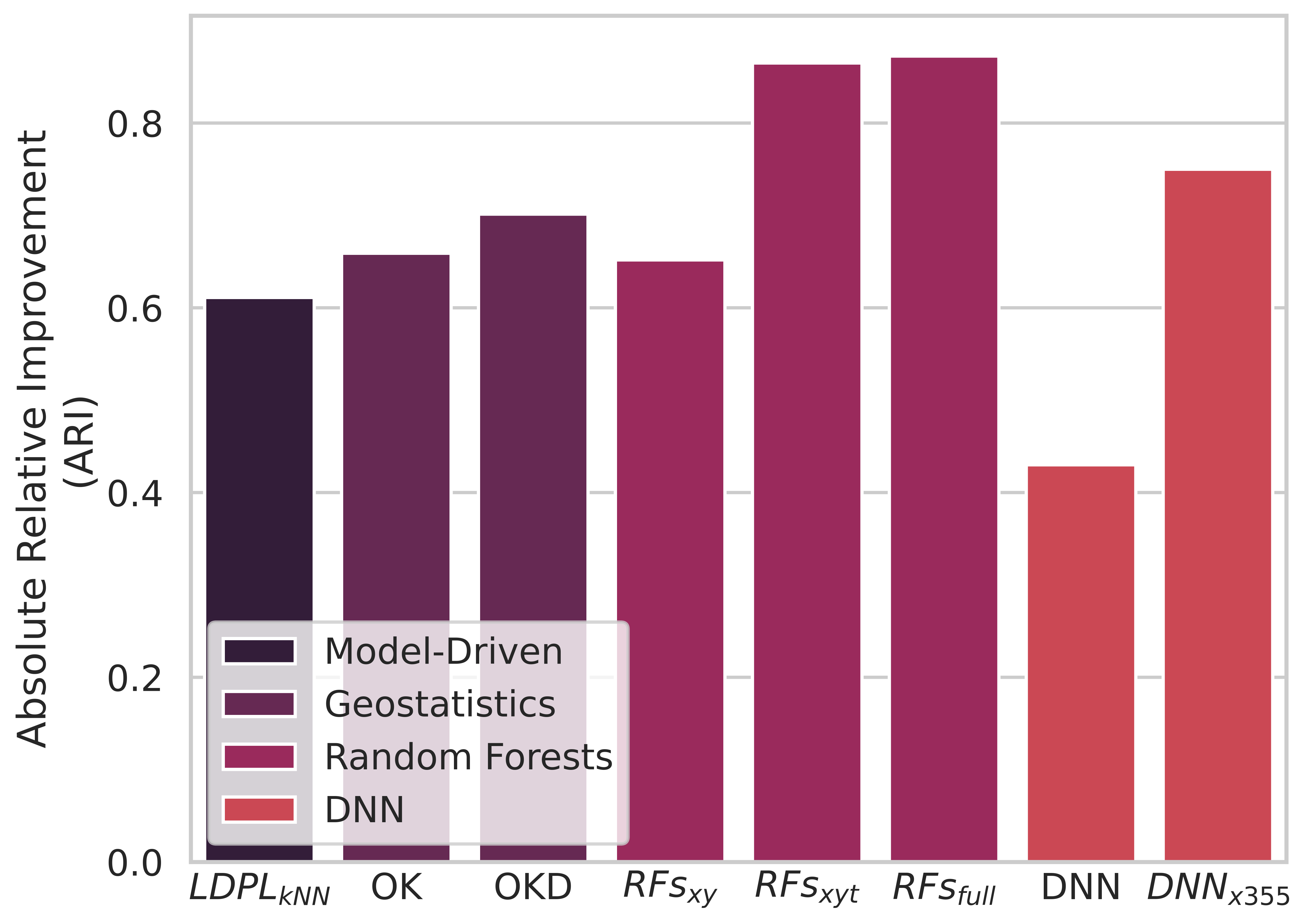}}
	\vspace{-15pt}
	\caption{\footnotesize Comparison of ARI for each predictor over the entire \inhousedataset. Our Approaches (\RFsSpatiotemporal, \RFsAll) outperform prior art in all scenarios.
	\label{fig:rmse_vs_ARI_campus_allmethods_v2}} 
	\vspace{-5pt}
\end{figure}

{\em a.  \inhousedataset:}  Table~\ref{tab:campus_allcells} reports the $RMSE$ for all predictors for each cell in the \inhouse dataset, for the default 70-30\% split. Fig. \ref{fig:rmse_vs_ARI_campus_allmethods_v2} compares all methods, calculating $RMSE$ over the entire \inhousedataset, instead of per cell. We can see that our \RFs-based predictors (\RFsSpatiotemporal,~\RFsAll) outperform model-based (\LDPL) and other data-driven (OK, OKD, DNN) predictors.

Fig.~\ref{fig:rmse_vs_trainsize_campus} shows the $RMSE$  as a function of the training size (as \% of all measurements in the dataset). First, the performance of OK and \RFsSpatial~is almost identical, as it can be seen for $RMSE$ over all measurements (Fig.~\ref{fig:rmse_vs_trainsize_campus} and Fig.~\ref{fig:rmse_vs_ARI_campus_allmethods_v2}) and $RMSE$ per cell (Table~\ref{tab:campus_allcells}). This result can be explained by the fact that both predictors are essentially a weighted average of their nearby measurements, although they achieve that in a different way: OK finds the weights by solving an optimization problem while \RFsSpatial~uses multiple decision trees and data splits.
Second and more importantly, considering additional features can significantly reduce the error. For the \inhousedataset, when time features $\mathbf{t}= (d,h)$ are added, \RFsSpatiotemporal~significantly outperforms OKD: it decreases $RMSE$ from $0.7 $ up to  $1.2$ dB. Alternatively, in terms of training size, \RFsSpatiotemporal~needs only $10\%$ of the data for training, in order to achieve OKD's lowest error ($\simeq2.8$dB) with $90\%$ of the measurement data for training. Our methodology achieves the lowest error of state-of-the-art geospatial predictors with $80\%$ less measurements. The absolute relative improvement of \RFsSpatiotemporal~compared to OKD is $17\%$, shown in Fig.~\ref{fig:rmse_vs_ARI_campus_allmethods_v2}(b).

\begin{figure}[t!]
	\centering
	\subfigure[\external Manhattan Midtown]{\includegraphics[scale=0.145]{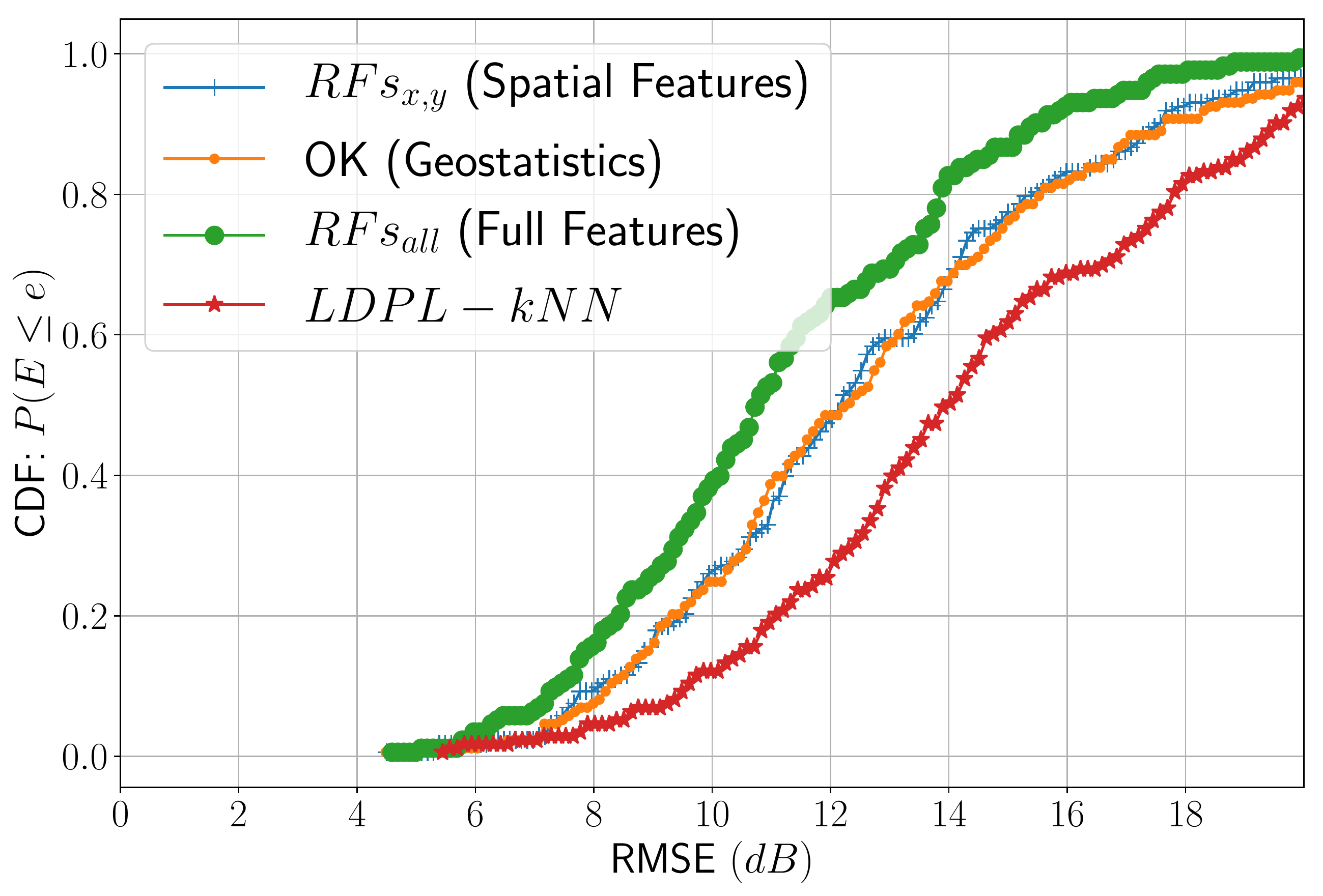}} 
	\subfigure[\externalsuburban Southern Suburb]{\includegraphics[scale=0.145]{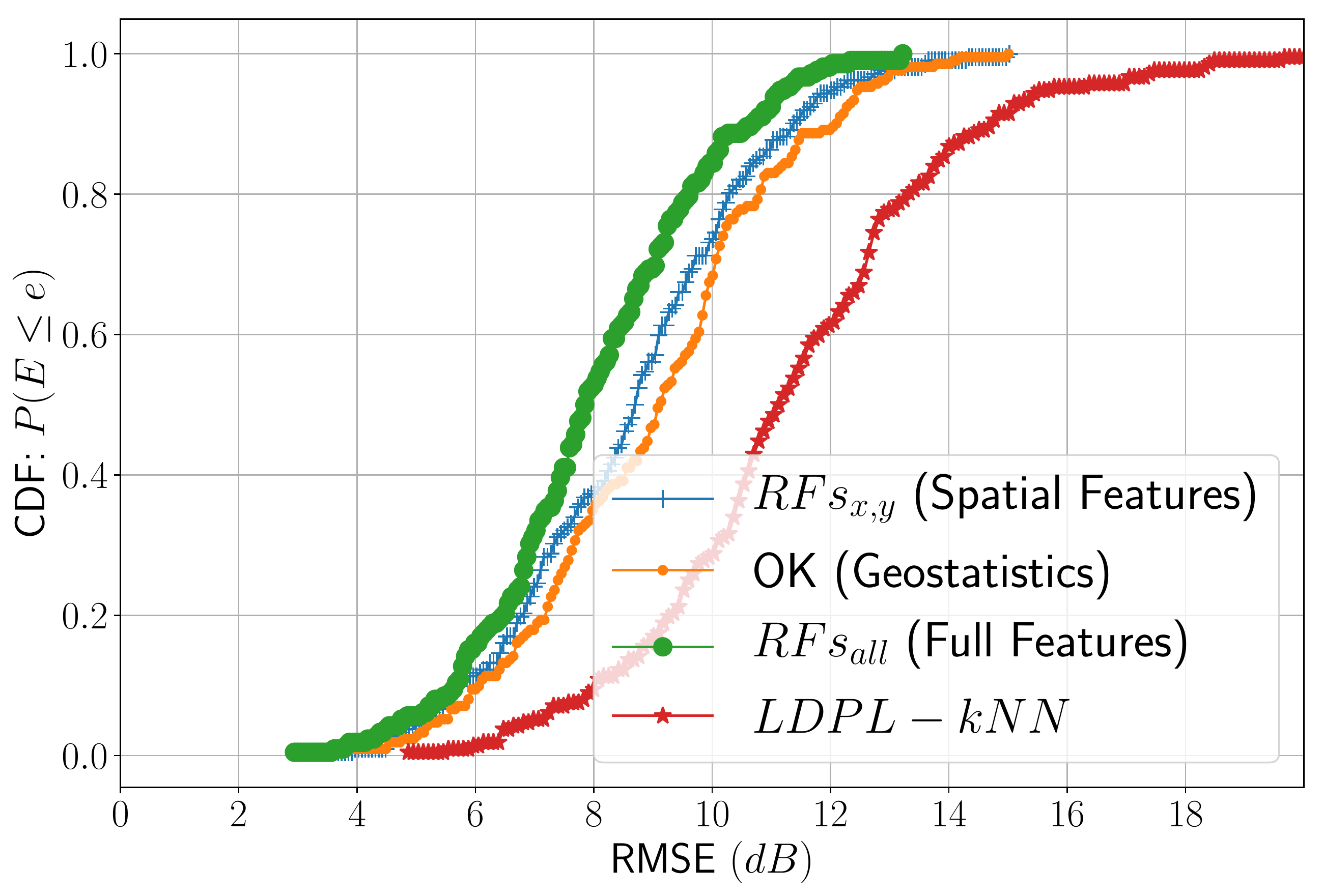}}
	\vspace{-10pt}
	\caption{\footnotesize \externalcitiesdataset: CDFs for $RMSE$ per \cid~ for two different LTE TAs, for the same major MNCarrier-1.  \RFsAll~  offer $2$dB gain  over the baselines for the 90th percentile.  
	\label{fig:cdf_rmse_citieswide}}
	\vspace{-5pt}
\end{figure}

{\em b.  \externalcitiesdataset:} Fig.~\ref{fig:cdf_rmse_citieswide} shows the error for two different LTE TAs, namely for \external Manhattan Midtown (urban) and for southern LA (suburban), where \RFs~have been trained per \cid. CDFs of the error per \cid~of the same \lteta~are plotted for different predictors. Again, OK performance is very close to \RFsSpatial, because they both exploit spatial features. However, \RFsAll~with the rich set of features improves by approx. $2$dB over the baselines for the 90th percentile, in both LTE TAs. Interestingly, the feature \dev~ is important, 
which is expected since this data, has heterogeneous devices reporting RSRP. Prior work has ignored  important device, radio frequency and hardware's receiver information. First, each device calculates its signal strength (\eg LTE RSRP)  differently (\ie proprietary algorithm in the device's cellular modem, since 3GPP just provides generic guidelines). Second, receiving sensitivity (the minimum RSRP for a feasible wireless communication) changes per device because each wireless receiver has different noise figure (NF)~\cite{alimpertis:14}. In sharp contrast to prior art, our \RFs-based predictor inherently considers this feature.

There are multiple reasons why \RFsAll~outperform geospatial interpolation predictors. The mean and variance of RSRP depend on time and location and the complex propagation environment. \RFs~can easily capture these dimensions instead of modeling a priori every single aspect. For example, \RFsSpatiotemporal~predicts a time-varying value for the measurements at the same location in Fig.~\ref{fig:campus_dense_x204}, while \RFsSpatial~or OK/OKD produce just a flat line over time. OK also relies on some assumptions (same mean over space, semivariogram  depending only on the distance between two locations), which  do not hold for RSRP. Even hybrid geospatial techniques (OKD) cannot naturally incorporate  additional features (\eg time,  device type, \etc). Finally, \RFs~significantly outperform propagation models.

Similarly, the \RFs-based predictors outperform the DNN-based predictor. First, DNNs cannot handle efficiently the disjoint regions and discontinuous values of the RSRP values as \RFs~inherently do (\eg in Fig.~\ref{fig:RFs_xy_examplesplits}). Second, DNNs need a huge amount of data in order to perform well with discontinuous and non smooth functions. Furthermore, more complex architectures of DNNs would be needed along with very specialized and expensive to obtain features such as Li-DAR data, 3D building models \etc ~\cite{raik:18}. Finally, DNNs need normalized input, which has significant effects on latitude/longitude values, and cannot support both categorical and continuous features.

We refer to our prior work in~\cite{alimpertis:19} for an extensive presentation of the MSE minimization, \problem{I,k}, where we considered: (i) The feature importance, (ii) the \RFs~model granularity, (iii) the differences between urban vs. suburban environments and (iv) the location density and we show how the \RFs~ do not overfit. For the rest of the paper, the focus is the prediction problem beyond MSE (\ie \problem{Q,k},  \problem{I,W} \problem{Q,W} and data Shapley valuation for data minimization.

\end{document}